%% file: main.tex
\documentclass[10pt,twocolumn,letterpaper]{article}
\usepackage{wacv}          

\input{preamble}

\definecolor{wacvblue}{rgb}{0.21,0.49,0.74}
\usepackage[pagebackref,breaklinks,colorlinks,allcolors=wacvblue]{hyperref}

\title{BigEarthNet.txt: A Large-Scale Multi-Sensor Image-Text Dataset and Benchmark for Earth Observation}

\author{Johann-Ludwig Herzog\textsuperscript{*1,2}, Mathis Jürgen Adler\textsuperscript{*1,2}, Leonard Hackel\textsuperscript{1,2}, Yan Shu\textsuperscript{3}, Angelos Zavras\textsuperscript{4},\\
Ioannis Papoutsis\textsuperscript{4}, Paolo Rota\textsuperscript{3}, Begüm Demir\textsuperscript{1,2}
\and
BIFOLD\textsuperscript{1}, TU Berlin\textsuperscript{2}, University of Trento\textsuperscript{3},\\
National Technical University and National Observatory of Athens\textsuperscript{4} \and
{\tt\small \{j.herzog,mathis.juergen.adler,l.hackel,demir\}@tu-berlin.de \{yan.shu,paolo.rota\}@unitn.it}\\
{\tt\small azabras@noa.gr ipapoutsis@mail.ntua.gr}
}

\begin{document}

\input{sec/_acronyms}

\maketitle
\begingroup
\renewcommand{\thefootnote}{*}
\footnotetext{These authors contributed equally to this work.}
\endgroup
\input{sec/0_abstract}
\acresetall

\section{Introduction}
\label{sec:intro}
\input{sec/1_intro}

\section{Related work}
\label{sec:related_works}
\input{sec/2_relatedWork}

\section{BigEarthNet.txt Dataset}
\label{sec:bigearth_txt}
\input{sec/3_bigearth_txt}

\section{Experiments}
\label{sec:experiments}
\input{sec/5_experiments}

\section{Conclusion}
\label{sec:conclusion}
\input{sec/6_conclusion}

\section*{Acknowledgements}

This work is supported by the European Research Council Project Agent-BigEarth (No. 101292498) and the European Union Horizon Project PROTEUS (No. 101296397). We thank Kai Norman Clasen for his support at the initial phase of this work.
\newpage
\FloatBarrier
{
    \small
    \bibliographystyle{ieeenat_fullname}
    \bibliography{main}
}

\FloatBarrier
\clearpage
\setcounter{section}{0}
\setcounter{table}{0}
\setcounter{figure}{0}
\setcounter{page}{1}
\renewcommand{\thesection}{\Alph{section}}
\label{sec:supplementary}
\input{sec/X_supplementary}

\end{document}

%% file: preamble.tex
\usepackage[T1]{fontenc}
\usepackage{siunitx}
\DeclareSIUnit{\nothing}{\relax}
\usepackage{acronym}
\usepackage{csquotes}
\usepackage{placeins}
\usepackage{makecell}
\usepackage{multirow}
\usepackage{xcolor}
\usepackage{colortbl}
\usepackage{tikz}
\usetikzlibrary{positioning}
\usepackage{xcolor}
\usepackage{colortbl}
\usepackage{array}
\usepackage{makecell}
\usepackage{pifont}
\usepackage{graphicx}
\usepackage{adjustbox}
\usepackage{ifthen}
\usepackage{xparse}
\usetikzlibrary{calc,fit,backgrounds,shapes.arrows,shapes.geometric,decorations.pathreplacing,decorations.text}
\usepackage{tikz-3dplot}
\usepackage{enumitem}
\usepackage{nicefrac}
\usepackage{subcaption}
\usepackage{enumitem}
\usepackage[most]{tcolorbox}
\definecolor{darkGreen}{RGB}{22, 119, 53}
\definecolor{lightGreen}{RGB}{198, 226, 208}
\definecolor{myGreen}{RGB}{22, 119, 53}
\usepackage{tabularx}
\usepackage{pifont}
\usepackage{tabularx}
\usepackage{array}
\usepackage{booktabs}
\usepackage{pifont}
\usepackage{ragged2e}
\usepackage{subcaption}
\usepackage{adjustbox}
\usepackage{longtable}

\newcommand{\missingref}[1]{\textcolor{red}{[REF?]}}
\newcommand{\rb}[1]{\rotatebox{90}{#1}}
\newcommand{\gap}[1]{\makebox[#1]{}}

\usepackage[commandnameprefix=always,final]{changes}

\usepackage{pifont}
\newcommand{\cmark}{\ding{51}}

\newcommand{\gcmark}{\textcolor{teal}{\ding{51}}}
\newcommand{\rcmark}{\textcolor{teal}{(\ding{51})}}
\newcommand{\rxmark}{\textcolor{red}{\ding{55}}}

\newcommand{\best}[1]{\textbf{#1}}
\newcommand{\second}[1]{\textit{#1}}

\makeatletter
\newcommand\notsotiny{\@setfontsize\notsotiny\@vipt\@viipt}
\makeatother

\newlength{\BenImgWidth}
\def\arrayStretchFactor{0.97}

\input{color_definitions}

\newcommand{\legendbox}[2]{%
  \textcolor{#1}{\rule{1.5ex}{1.5ex}}\,#2%
}
\def\centerarc[#1](#2)(#3:#4:#5)
    { \draw[#1] ($(#2)+({#5*cos(#3)},{#5*sin(#3)})$) arc (#3:#4:#5); }

\newcommand{\bentxt}{\texttt{BigEarthNet.txt}}
\newcommand{\bench}{\text{benchmark split}}
\newcommand{\genDet}{referring expression detection}
\newcommand{\pointDet}{referring point detection}
\newcommand{\refDet}{referring \ac{LULC} detection}

\newcommand{\suppref}[1]{supplementary material~\ref{#1}}

\newcommand{\clickablelabel}[1]{
    \stepcounter{enumi}
    \llap{\hyperref[#1]{\Alph{enumi}.}}
}

\input{tables/results/table_abbreviations}

\renewcommand{\arraystretch}{0.83}
\definecolor{boxblue}{RGB}{74,138,181}
\definecolor{lightbg}{RGB}{248,248,248}

\tikzset{
  captionbox/.style={
    draw=boxblue,
    rounded corners=3pt,
    line width=0.8pt,
    fill=lightbg,
    inner sep=2pt,
    align=left
  },
  row1captionbox/.style={
    captionbox,
    text width=0.29\linewidth
  },
  row2captionbox/.style={
    captionbox,
    text width=0.31\linewidth
  },
  lhrsbox/.style={
    captionbox,
    text width=0.36\linewidth
  },
  hugecaptionbox/.style={
    captionbox,
    text width=12.5cm
  },
  icon/.style={
    circle,
    draw=gray!60,
    fill=white,
    minimum size=7mm,
    inner sep=0pt
  },
}

%% file: color_definitions.tex
\definecolor{UrbanFabric111}{RGB}{230, 0, 77}                     
\definecolor{UrbanFabric112}{RGB}{255, 102, 102}                  
\definecolor{IndustrialCommercialUnits121}{RGB}{204, 77, 242}     
\definecolor{Unlabeled122}{RGB}{204, 204, 204}                    
\definecolor{Unlabeled123}{RGB}{0, 102, 204}                      
\definecolor{Unlabeled124}{RGB}{153, 153, 153}                    
\definecolor{Unlabeled131}{RGB}{166, 77, 0}                       
\definecolor{Unlabeled132}{RGB}{102, 51, 0}                       
\definecolor{Unlabeled133}{RGB}{255, 153, 51}                     
\definecolor{Unlabeled141}{RGB}{0, 204, 0}                        
\definecolor{Unlabeled142}{RGB}{102, 255, 102}                    

\definecolor{ArableLand211}{RGB}{255, 255, 102}                   
\definecolor{ArableLand212}{RGB}{204, 255, 102}                   
\definecolor{ArableLand213}{RGB}{153, 255, 153}                   
\definecolor{PermanentCrops221}{RGB}{204, 153, 0}                 
\definecolor{PermanentCrops222}{RGB}{255, 204, 102}               
\definecolor{PermanentCrops223}{RGB}{153, 204, 0}                 
\definecolor{Pastures231}{RGB}{204, 255, 153}                     
\definecolor{PermanentCrops241}{RGB}{255, 230, 153}               
\definecolor{ComplexCultivationPatterns242}{RGB}{255, 255, 179}   
\definecolor{LandNaturalVegetation243}{RGB}{204, 230, 179}        
\definecolor{AgroForestryAreas244}{RGB}{179, 230, 102}            

\definecolor{BroadLeavedForest311}{RGB}{0, 102, 0}                
\definecolor{ConiferousForest312}{RGB}{0, 153, 0}                 
\definecolor{MixedForest313}{RGB}{51, 204, 51}                    

\definecolor{NaturalGrassland321}{RGB}{204, 255, 204}             
\definecolor{MoorsHeathland322}{RGB}{153, 153, 102}               
\definecolor{MoorsHeathland323}{RGB}{102, 153, 51}                
\definecolor{TransitionalWoodlandShrub324}{RGB}{153, 204, 153}    

\definecolor{BeachesDunesSands331}{RGB}{255, 255, 204}            
\definecolor{Unlabeled332}{RGB}{192, 192, 192}                    
\definecolor{NaturalGrassland333}{RGB}{224, 224, 224}             
\definecolor{Unlabeled334}{RGB}{102, 0, 0}                        

\definecolor{InlandWetlands411}{RGB}{204, 153, 255}               
\definecolor{InlandWetlands412}{RGB}{153, 102, 204}               
\definecolor{CoastalWetlands421}{RGB}{204, 204, 255}              
\definecolor{CoastalWetlands422}{RGB}{153, 204, 255}              
\definecolor{Unlabeled423}{RGB}{102, 178, 255}                    

\definecolor{InlandWaters511}{RGB}{0, 128, 255}                   
\definecolor{InlandWaters512}{RGB}{0, 0, 204}                     
\definecolor{MarineWaters521}{RGB}{51, 153, 255}                  
\definecolor{MarineWaters522}{RGB}{0, 204, 255}                   
\definecolor{MarineWaters523}{RGB}{0, 51, 153}                    

\definecolor{Unlabeled999}{RGB}{0, 0, 0}                          

%% file: tables/results/table_abbreviations.tex
\newcommand{\geochatABBRns}{GC}
\newcommand{\lhrsbotABBRns}{LHRS}
\newcommand{\skyeyegptABBRns}{SE}
\newcommand{\earthdialrgbABBRns}{ED$_\text{rgb}$}
\newcommand{\earthdialmsABBRns}{ED$_\text{s2}$}
\newcommand{\earthmindrgbABBRns}{EM$_\text{rgb}$}

\newcommand{\earthmindsarmsABBRns}{EM$_\text{s1s2}$}
\newcommand{\gptABBRns}{GPT}
\newcommand{\qwenABBRns}{Qwen}
\newcommand{\glmABBRns}{GLM}
\newcommand{\llavaABBRns}{LLaVa}
\newcommand{\internvlABBRns}{IVL}
\newcommand{\internvlrsABBRns}{IVL$^*$}

\newcommand{\geochatABBR}{\geochatABBRns~\cite{kuckreja2024geochat}}
\newcommand{\lhrsbotABBR}{\lhrsbotABBRns~\cite{muhtar2024lhrs}}
\newcommand{\skyeyegptABBR}{\skyeyegptABBRns~\cite{zhan2025skyeyegpt}}
\newcommand{\earthdialrgbABBR}{\earthdialrgbABBRns~\cite{soni2025earthdial}}
\newcommand{\earthdialmsABBR}{\earthdialmsABBRns~\cite{soni2025earthdial}}
\newcommand{\earthmindrgbABBR}{\earthmindrgbABBRns~\cite{shu2025earthmind}}

\newcommand{\earthmindsarmsABBR}{\earthmindsarmsABBRns~\cite{shu2025earthmind}}
\newcommand{\gptABBR}{\gptABBRns~\cite{openai2025gpt5}}
\newcommand{\qwenABBR}{\qwenABBRns~\cite{bai2025qwen3}}
\newcommand{\glmABBR}{\glmABBRns~\cite{hong2025glm}}
\newcommand{\llavaABBR}{\llavaABBRns~\cite{li2024llavaonevision}}
\newcommand{\internvlABBR}{\internvlABBRns~\cite{zhu2025internvl3}}
\newcommand{\internvlrsABBR}{\internvlrsABBRns}

\newcommand{\presenceABBR}{Pr}
\newcommand{\areaABBR}{A}
\newcommand{\countABBR}{Cnt}
\newcommand{\adjacencyABBR}{Adj}
\newcommand{\relativepositionABBR}{RP}
\newcommand{\countryABBR}{Loc}
\newcommand{\seasonABBR}{S}
\newcommand{\climateABBR}{Clt}
\newcommand{\overallABBR}{OA}
\newcommand{\instructionfollowedABBR}{IF}

%% file: sec/_acronyms.tex
\acrodef{AA}{average accuracy}
\acrodef{RGB}{Red-Green-Blue}
\acrodef{AI}{artificial intelligence}
\acrodef{CBIR}{content-based image retrieval}
\acrodef{CL}{contrastive learning}
\acrodef{CLC}{CORINE Land Cover}
\acrodef{CV}{computer vision}
\acrodef{DL}{deep learning}
\acrodef{DEA}{Digital Earth Australia}
\acrodef{EO}{Earth observation}
\acrodef{FLOP}{floating-point operation}
\acrodef{FFN}{feed-forward network}
\acrodef{FM}{foundation model}
\acrodef{mIoU}[mIoU]{mean intersection-over-union}
\acrodef{LLM}{large language model}
\acrodef{LMM}{large multi-modal model}
\acrodef{LULC}{land-use/land-cover}
\acrodef{MAE}{masked autoencoder}
\acrodef{mAP}[mAP$_\mu$]{micro-mean average precision}
\acrodef{MC}{multiple-choice}
\acrodef{MCQ}{multiple-choice question}
\acrodef{MLP}{multi-layer perceptron}
\acrodef{MS}{multispectral}
\acrodef{MTLD}{measure of textual lexical diversity}
\acrodef{NLTK}{Natural Language Toolkit}
\acrodef{NTXent}[NT-Xent]{normalized temperature-scaled cross entropy}
\acrodef{PDF}{probability density function}
\acrodef{RS}{remote sensing}
\acrodef{RTC}{radiometric terrain-corrected}
\acrodef{S1}{Sentinel-1}
\acrodef{S2}{Sentinel-2}
\acrodef{SAR}{synthetic aperture radar}
\acrodef{SOTA}{state-of-the-art}
\acrodef{TCI}{true color image}
\acrodef{ViT}{vision transformer}
\acrodef{VLM}{vision-language model}
\acrodef{VQA}{visual question answering}

%% file: sec/0_abstract.tex
\begin{abstract}
\Acp{VLM} have shown strong performance in \ac{CV}, yet their performance on \ac{RS} data remains limited due to the lack of large-scale, multi-sensor \ac{RS} image-text datasets with diverse textual annotations. 
Existing datasets predominantly include aerial \acl{RGB} imagery, with short or weakly grounded captions, and provide limited diversity in annotation types. 
To address this limitation, we introduce \bentxt{}, a large-scale, multi-sensor image-text dataset designed to advance instruction-driven image-text learning in \acl{EO} across multiple tasks. 
It contains \num{464044} co-registered \acl{S1} \acl{SAR} and \acl{S2} \acl{MS} images with \SI{9.6}{\mega\nothing} text annotations, including:
i) geographically anchored captions describing \ac{LULC} classes, their spatial relations, and environmental context;
ii) \acl{VQA} pairs relevant for different tasks; and
iii) \genDet{} instructions for bounding box prediction.
A comparative statistical analysis shows that \bentxt{} surpasses existing \ac{RS} image-text datasets in textual richness and annotation type variety. 
We further establish a manually-verified \bench{} and extensively evaluate existing \acp{VLM} in \ac{RS} and \ac{CV}. The results show the limitations of these models on tasks that involve complex \ac{LULC} classes, whereas fine-tuning using \bentxt{} results in consistent performance gains across all considered tasks. 
Our code and data are publicly available at \href{https://txt.bigearth.net}{txt.bigearth.net}.
\vspace{-.5em}
\end{abstract}

%% file: sec/1_intro.tex
\input{figures/dataset_overview_v1}
Unprecedented advances in satellite technology have led to a significant increase in the volume of \ac{EO} data archives (\eg the Sentinel satellites of the Copernicus program alone acquire roughly 20~TB of satellite images per day).
In this data-rich landscape, natural language provides an intuitive interface for querying and interpreting these vast \ac{EO} archives, with \acp{VLM} emerging as an effective framework for jointly encoding visual observations and natural language for tasks such as image captioning~\cite{cheng2022nwpu, hoxha2020toward, sumbul2020sd}, \ac{VQA}~\cite{lobry2020rsvqa, chappuis2022prompt, hackel2023lit} and \genDet~\cite{kuckreja2024geochat, soni2025earthdial, zhan2025skyeyegpt}.

In general, \acp{VLM} are applied to \ac{EO} data either by using general-purpose \ac{CV} \acp{VLM} directly or by training \ac{RS}-specific \acp{VLM}.
The first approach is limited by the fundamentally different spectral and spatial characteristics between \ac{CV} and \ac{RS} images. For example, \ac{S2} \ac{MS} images comprise 13 spectral bands associated with varying spatial resolutions, while those in \ac{CV} include \ac{RGB} bands in general. The high spectral resolution data enable the discrimination of complex \ac{LULC} classes in RS. Although \ac{RGB} information may be sufficient to distinguish generic classes, \eg,\enquote{Water} from \enquote{Forests and seminatural areas}, additional spectral information beyond the visible spectrum is needed to differentiate complex \ac{LULC} classes, \eg, \enquote{Mixed forest} vs. \enquote{Coniferous forest}~\cite{sumbul2019bigearthnet}. Due to their pretraining phase using \ac{RGB}-only image-text data, the \acp{VLM} in \ac{CV} have limited capability to effectively query and interpret RS data. Although several \ac{RS}-specific \acp{VLM} trained on \ac{RS} image-text datasets~\cite{bazi2024rsllava, zhan2025skyeyegpt, kuckreja2024geochat, muhtar2024lhrs} have recently been developed, most of these models are still trained on \ac{RGB} data only and thus subject to the above-mentioned limitations. It is worth noting that there are only a few works that include multispectral or multi-sensor (e.g., multispectral, \ac{SAR})  image-text data during pre-training~\cite{shu2025earthmind, soni2025earthdial}. The use of multi-sensor data is particularly crucial, since different satellite sensors have the capability of measuring different physical properties of \ac{LULC} classes and their joint use can improve task performance ~\cite{sumbul2021bigearthnet, hong2020more, Sumbul_2025_ICCV}.
However, as shown in \cref{tab:existing_datasets_properties}, the existing \ac{RS} image-text datasets suffer from: i) limited availability of co-registered multi-sensor data that includes more than three bands, and ii) limited diversity in text annotation types.
These limitations directly hinder the ability of \acp{VLM} to characterize: 1) the complex spatial/spectral content of \ac{RS} images; and also 2) complementary information among multi-sensor data. 
 \input{tables/dataset/comp_existing_v2}

To overcome the above-mentioned limitations, in this paper we present \bentxt: a large-scale image-text dataset comprising \num{464044} co-registered \ac{S1} \ac{SAR} and \ac{S2} \acl{MS} images with \SI{9.6}{\mega\nothing} text annotations, including captions, \acp{VQA} with varying answer formats, and \genDet{} annotations (see \cref{fig:dataset_overview}). In addition, we introduce a \bench{} that contains \num{1082} image pairs with \num{15029} manually verified text annotations for systematic evaluation of \acp{VLM} for \ac{EO} tasks. Using our \bench{}, we demonstrate that current \acp{VLM} in both \ac{CV} and \ac{RS} show limited ability to characterize the complex content of multi-sensor \ac{RS} images. Fine-tuning a \ac{SOTA} \ac{VLM} (InternVL3-1B~\cite{zhu2025internvl3}) adapted for multi-sensor input (denoted as InternVL$^*$) on \bentxt{} leads to substantial performance gains across all considered tasks.
In summary, our contributions are as follows:
\begin{enumerate}
    \item We show that the existing \acp{VLM} from both \ac{CV} and \ac{RS} domains exhibit limited capability to characterize the complex spatial/spectral content of multi-sensor images in \ac{RS}. To this end, we conduct extensive experiments across 15 tasks of 4 categories.
    \item We introduce \bentxt{}, the first dataset supporting multiple task types, i.e., image captioning, binary VQA, MCQ, and \genDet{} with several different text annotation types for co-registered multi-modal images, with a manually checked subset (\bench{}).
    \item We adapt InternVL to accommodate multi-sensor RS inputs and perform domain-specific fine-tuning using \bentxt. Our results serve as experimental validation of the critical role of large-scale, multi-sensor image-text datasets for accurate interactions with EO data.
\end{enumerate}

%% file: figures/dataset_overview_v1.tex
\newcommand{\RingSegment}[6]{
\fill[#6]
  (#1) ++(#4:#2)
  arc[start angle=#4, delta angle=#5, radius=#2]
  -- ++(#4+#5:#3-#2)
  arc[start angle=#4+#5, delta angle=-#5, radius=#3]
  -- cycle;
}
\newcommand{\RingLabel}[5]{
  \pgfmathsetmacro{\angTmp}{#3 + 0.5*#4}
  \pgfmathsetmacro{\rotTmp}{\angTmp}
  \pgfmathsetmacro{\rotTmp}{%
      (\angTmp < 240) ? \angTmp + 180 : ((\angTmp < 300) ? \angTmp + 90 : \angTmp)
    }
  \node[
    font=\bfseries\scriptsize,
    align=center,
    rotate=\rotTmp
  ] at ($ (#1) + (\angTmp:#2) $) {#5};
}
\newcommand{\RingLabelCurved}[5]{
  \pgfmathsetmacro{\aMid}{#3 + 0.5*#4}
  \path[
    decorate,
    decoration={
      text along path,
      text={|\bfseries\footnotesize|#5},
      text align=center,
      raise=0.5ex
    }
  ]
  (#1) ++(#3:#2)
  arc[start angle=#3, delta angle=#4, radius=#2];
}
\NewDocumentCommand{\InputOutputBox}{m m m m m m m o}{%
    \node[align=justify, anchor=#1, text width=#3, minimum width=#3, minimum height=#2, fill=#4, rounded corners=2mm, font=\fontsize{6pt}{6pt}\selectfont] (#8) at (#5)
        {
            \textbf{Input:} #6\newline
            \textbf{Output:} #7
        };
}

\begin{figure*}[h!tb]
    \centering
    \begin{tikzpicture}
        \node[anchor=center] (refMap) at (0,0) {};

        \def\rInner{0.4*\BenImgWidth}
        \def\rMid{0.85*\BenImgWidth}
        \def\rOuter{1.25*\BenImgWidth}
        \def\lineSep{1pt}
        \pgfmathsetmacro{\rLabelInner}{1.9}
        \pgfmathsetmacro{\rLabelMid}{2.1}
        \pgfmathsetmacro{\rLabelOuter}{2.5}
        
        \def\aStart{90}
        \def\aBin{136.6}
        \def\aBB{83.1}
        \def\aCap{17.5}
        \def\aMCQ{122.8}
        
        \RingSegment{refMap}{\rInner}{\rMid}{\aStart}{\aBin}{Orchid!42.5, draw=white, line width=\lineSep}
        \RingSegment{refMap}{\rInner}{\rMid}{\aStart+\aBin}{\aBB}{red!37.5,draw=white,line width=\lineSep}
        \RingSegment{refMap}{\rInner}{\rOuter}{\aStart+\aBin+\aBB}{\aCap}{yellow!50,draw=white,line width=\lineSep}
        \RingSegment{refMap}{\rInner}{\rMid}{\aStart+\aBin+\aBB+\aCap}{\aMCQ}{YellowGreen!40,draw=white,line width=\lineSep}
        \RingLabelCurved{refMap}{\rLabelInner}{\aStart}{\aBin}{Binary VQA}
        \RingLabelCurved{refMap}{0.85*\rLabelInner}{\aStart+\aBin}{\aBB}{Ref. Exp.}
        \RingLabelCurved{refMap}{\rLabelInner}{\aStart+\aBin}{\aBB}{Detection}
        \RingLabel{refMap}{\rLabelMid}{\aStart+\aBin+\aBB}{\aCap}{Captioning}
        \RingLabelCurved{refMap}{\rLabelInner}{\aStart+\aBin+\aBB+\aCap}{\aMCQ}{Multiple-Choice VQA}

        \def\bBinAdj{31.8}
        \def\bBinArea{34.9}
        \def\bBinCount{34.9}
        \def\bBinPresence{34.9}
        \RingSegment{refMap}{\rMid}{\rOuter}{\aStart}{\bBinAdj}{Orchid!15, draw=white, line width=\lineSep}
        \RingSegment{refMap}{\rMid}{\rOuter}{\aStart+\bBinAdj}{\bBinArea}{Orchid!30, draw=white, line width=\lineSep}
        \RingSegment{refMap}{\rMid}{\rOuter}{\aStart+\bBinAdj+\bBinArea}{\bBinCount}{Orchid!45, draw=white, line width=\lineSep}
        \RingSegment{refMap}{\rMid}{\rOuter}{\aStart+\bBinAdj+\bBinArea+\bBinCount}{\bBinPresence}{Orchid!60, draw=white, line width=\lineSep}
        \RingLabel{refMap}{\rLabelOuter}{\aStart}{\bBinAdj}{\adjacencyABBR}
        \RingLabel{refMap}{\rLabelOuter}{\aStart+\bBinAdj}{\bBinArea}{\areaABBR}
        \RingLabel{refMap}{\rLabelOuter}{\aStart+\bBinAdj+\bBinArea}{\bBinCount}{\countABBR}
        \RingLabel{refMap}{\rLabelOuter}{\aStart+\bBinAdj+\bBinArea+\bBinCount}{\bBinPresence}{\presenceABBR}
        
        \def\bBBPoint{43.1}
        \def\bBBRef{40.0}
        \RingSegment{refMap}{\rMid}{\rOuter}{\aStart+\aBin}{\bBBPoint}{red!25,draw=white,line width=\lineSep}
        \RingSegment{refMap}{\rMid}{\rOuter}{\aStart+\aBin+\bBBPoint}{\bBBRef}{red!50,draw=white,line width=\lineSep}
        \RingLabel{refMap}{\rLabelOuter}{\aStart+\aBin}{\bBBPoint}{Ref. Point\\Detection}
        \RingLabel{refMap}{\rLabelOuter}{\aStart+\aBin+\bBBPoint}{\bBBRef}{Ref. LULC\\Detection}
        
        \def\bCap{17.5}
        
        \def\bMCQAdj{14.3}
        \def\bMCQArea{17.5}
        \def\bMCQClimate{17.5}
        \def\bMCQCount{17.5}
        \def\bMCQCountry{17.5}
        \def\bMCQPresence{17.5}
        \def\bMCQRelPos{3.7}
        \def\bMCQSeason{17.5}
        \RingSegment{refMap}{\rMid}{\rOuter}{\aStart+\aBin+\aBB+\aCap}{\bMCQAdj}{YellowGreen!10,draw=white,line width=\lineSep}
        \RingSegment{refMap}{\rMid}{\rOuter}{\aStart+\aBin+\aBB+\aCap+\bMCQAdj}{\bMCQArea}{YellowGreen!20,draw=white,line width=\lineSep}
        \RingSegment{refMap}{\rMid}{\rOuter}{\aStart+\aBin+\aBB+\aCap+\bMCQAdj+\bMCQArea}{\bMCQClimate}{YellowGreen!30,draw=white,line width=\lineSep}
        \RingSegment{refMap}{\rMid}{\rOuter}{\aStart+\aBin+\aBB+\aCap+\bMCQAdj+\bMCQArea+\bMCQClimate}{\bMCQCount}{YellowGreen!40,draw=white,line width=\lineSep}
        \RingSegment{refMap}{\rMid}{\rOuter}{\aStart+\aBin+\aBB+\aCap+\bMCQAdj+\bMCQArea+\bMCQClimate+\bMCQCount}{\bMCQCountry}{YellowGreen!50,draw=white,line width=\lineSep}
        \RingSegment{refMap}{\rMid}{\rOuter}{\aStart+\aBin+\aBB+\aCap+\bMCQAdj+\bMCQArea+\bMCQClimate+\bMCQCount+\bMCQCountry}{\bMCQPresence}{YellowGreen!60,draw=white,line width=\lineSep}
        \RingSegment{refMap}{\rMid}{\rOuter}{\aStart+\aBin+\aBB+\aCap+\bMCQAdj+\bMCQArea+\bMCQClimate+\bMCQCount+\bMCQCountry+\bMCQPresence}{\bMCQRelPos}{YellowGreen!70,draw=white,line width=\lineSep}
        \RingSegment{refMap}{\rMid}{\rOuter}{\aStart+\aBin+\aBB+\aCap+\bMCQAdj+\bMCQArea+\bMCQClimate+\bMCQCount+\bMCQCountry+\bMCQPresence+\bMCQRelPos}{\bMCQSeason}{YellowGreen!80,draw=white,line width=\lineSep}
        \RingLabel{refMap}{\rLabelOuter}{\aStart+\aBin+\aBB+\aCap}{\bMCQAdj}{\adjacencyABBR}
        \RingLabel{refMap}{\rLabelOuter}{\aStart+\aBin+\aBB+\aCap+\bMCQAdj}{\bMCQArea}{\areaABBR}
        \RingLabel{refMap}{\rLabelOuter}{\aStart+\aBin+\aBB+\aCap+\bMCQAdj+\bMCQArea}{\bMCQClimate}{\climateABBR}
        \RingLabel{refMap}{\rLabelOuter}{\aStart+\aBin+\aBB+\aCap+\bMCQAdj+\bMCQArea+\bMCQClimate}{\bMCQCount}{\countABBR}
        \RingLabel{refMap}{\rLabelOuter}{\aStart+\aBin+\aBB+\aCap+\bMCQAdj+\bMCQArea+\bMCQClimate+\bMCQCount}{\bMCQCountry}{\countryABBR}
        \RingLabel{refMap}{\rLabelOuter}{\aStart+\aBin+\aBB+\aCap+\bMCQAdj+\bMCQArea+\bMCQClimate+\bMCQCount+\bMCQCountry}{\bMCQPresence}{\presenceABBR}
        \RingLabel{refMap}{\rLabelOuter}{\aStart+\aBin+\aBB+\aCap+\bMCQAdj+\bMCQArea+\bMCQClimate+\bMCQCount+\bMCQCountry+\bMCQPresence}{\bMCQRelPos}{\relativepositionABBR}
        \RingLabel{refMap}{\rLabelOuter}{\aStart+\aBin+\aBB+\aCap+\bMCQAdj+\bMCQArea+\bMCQClimate+\bMCQCount+\bMCQCountry+\bMCQPresence+\bMCQRelPos}{\bMCQSeason}{\seasonABBR}

        \begin{scope}
          \clip[rounded corners=5mm]
            ($(refMap)+(-0.5*\BenImgWidth,-0.5*\BenImgWidth)$)
            rectangle
            ($(refMap)+( 0.5*\BenImgWidth, 0.5*\BenImgWidth)$);
        
          \node[anchor=center, inner sep=0pt] at (refMap)
            {\input{figures/ben_txt_logos/v1}};
        \end{scope}
        
        \def\boxHeight{1mm}
        \def\boxWidth{\BenImgWidth-6pt}
        \InputOutputBox{north east}{0.86*\boxHeight}{\boxWidth}{YellowGreen!40}{$(-\rOuter, \rOuter) + (-.3,-.3)$}{How many areas covered by urban fabric can be seen? a)~More than five, b)~3, c)~1, d)~0}{b}[count_box]
        \InputOutputBox{north west}{0.7*\boxHeight}{\boxWidth}{Orchid!15}{$(count_box.south west) + (0,-0.1)$}{Does any inland water border inland wetlands in this scene?}{Yes}[adjacency_box]
        \InputOutputBox{north west}{0.7*\boxHeight}{\boxWidth}{Orchid!60}{$(adjacency_box.south west) + (0,-.1)$}{Are there regions of coastal wetlands in the satellite image?}{No}[presence_box]
        \InputOutputBox{north west}{0.7*\boxHeight}{\boxWidth}{YellowGreen!60}{$(presence_box.south west) + (0,-.1)$}{Which season is shown in the satellite image? a)~Spring, b)~Summer, c)~Winter, d)~Autumn}{b}

        \InputOutputBox{north west}{\boxHeight}{\boxWidth}{YellowGreen!70}{$(\rOuter, \rOuter) + (.3,-.3)$}{What is the relative position of the arable land to the inland waters? a) to the left, b) to the bottom, c) to the top-right, d) to the top}{a}[rel_pos_box]
        \InputOutputBox{north west}{\boxHeight}{\boxWidth}{red!50}{$(rel_pos_box.south west) + (0,-.1)$}{Where can the <ref> largest area of urban fabric</ref> be found?}{[0.0 0.55, 0.2 1.0]}[ref_box]
        \InputOutputBox{north west}{\boxHeight}{\boxWidth}{red!25}{$(ref_box.south west) + (0,-.1)$}{Output a bounding box enclosing the land cover class instance positioned at <point>(0.83, 0.06)</point>.}{[0.49 0.0, 1.0 0.2]}[point_box]
        
        \def\boxWidth{20mm}
        \InputOutputBox{south}{0.1*\BenImgWidth}{4*\boxWidth-1.7cm}{yellow!50}{$(0, \rOuter) + (0,.1)$}{Describe the content of the image, including the region, climate zone, and land cover distribution.}{This satellite image, captured during the summer season in Switzerland, showcases a diverse landscape within the "temperate, no dry season, warm summer" climate zone. The dominant features are arable land ($\sim$526,000 sqm) [...] are adjacent to both inland waters ($\sim$305,000 sqm) and urban fabric. Notably, the urban fabric is distributed over three individual marginal areas. The varied landscape presents a mix of agricultural areas, wetlands, water bodies, and artificial surfaces.}[]

        \begin{scope}
          \node[anchor=south west, inner sep=0pt] at ($(count_box.north west) + (0,.1)$)
            {\includegraphics[width=\BenImgWidth]{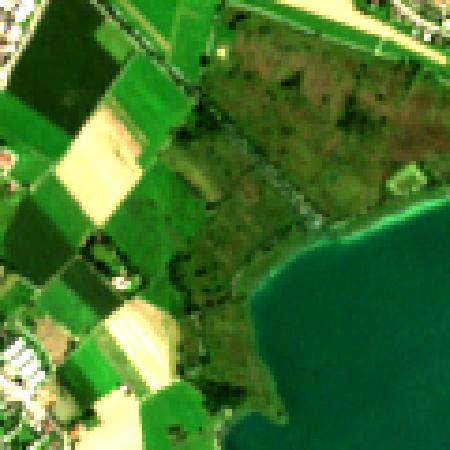}};
        \end{scope}
        \begin{scope}
          \node[anchor=south west, inner sep=0pt] at ($(rel_pos_box.north west) + (0,.1)$)
            {\includegraphics[width=\BenImgWidth]{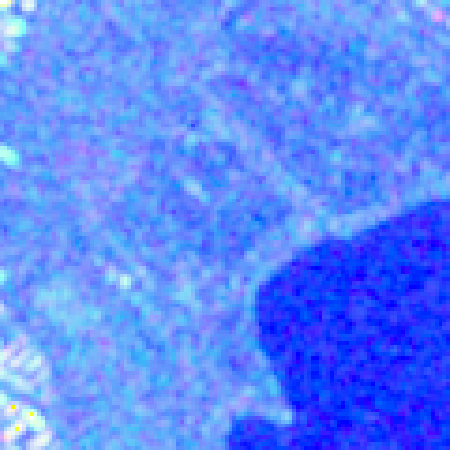}};
        \end{scope}
    \end{tikzpicture} 
    \caption{\bentxt{} comprises \num{464 044} co-registered \ac{S1} and \ac{S2} images with diverse text annotations, resulting in a total of $\sim~\num{9.6}$ million \ac{S1}-\ac{S2}-text triplets. The dataset supports 15 tasks (Presence, Area, Counting, Adjacency, Relative Position, Country, Season, and Climate Zone, denoted as \presenceABBR, \areaABBR, \countABBR, \adjacencyABBR, \relativepositionABBR, \countryABBR, \seasonABBR, and \climateABBR, respectively) across 4 broad categories.}
    \label{fig:dataset_overview}
\end{figure*}

%% file: figures/ben_txt_logos/v1.tex
\begin{tikzpicture}[scale=0.9]
  \fill[BlueGreen!20] (0,0) circle (1.2);
  \draw[white, line width=1pt] (0,0) circle (1.2);
  \draw[BlueGreen!50, line width=0.6pt] (0,0) circle (1);

  \draw[BlueGreen!60, line width=0.6pt] (-1.0,0) arc (180:360:1.0 and 0.35);
  \draw[BlueGreen!40, line width=0.3pt, dashed] (-1.0,0) arc (180:0:1.0 and 0.35);

  \draw[BlueGreen!40, line width=0.3pt, dashed] (0,-1.0) arc (-90:90:0.35 and 1.0);
  \draw[BlueGreen!60, line width=0.6pt] (0,-1.0) arc (-90:-270:0.35 and 1.0);

  \draw[BlueGreen!50, line width=0.6pt] (0,0) circle (1);

  \node at (-0.6, 0) {\includegraphics[scale=0.021,decodearray={0 0.65 0 0.65 0 0.65}]{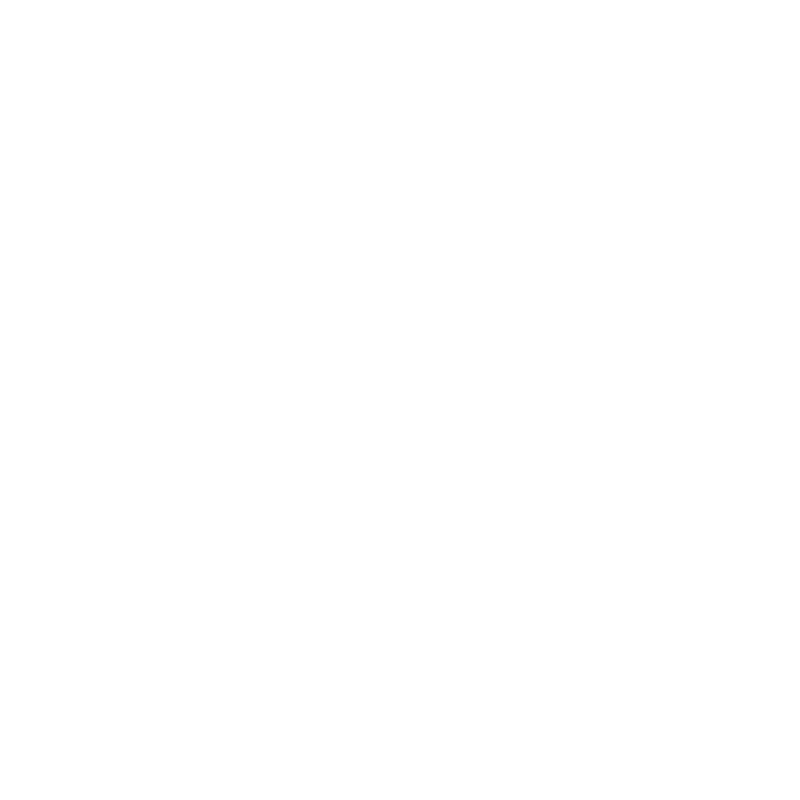}};
  \fill[gray!70] (-0.2,-0.2) rectangle (0.8,-0.12);
  \fill[gray!70] (-0.2,-0.05) rectangle (0.65,0.03);
  \fill[gray!70] (-0.2,0.10) rectangle (0.75,0.18);

  \path[decorate,decoration={text along path,text={|\bfseries\notsotiny|BigEarthNet},text align=center,raise=-1.6ex}] (-1.0,0) arc (180:360:1.0 and 0.35);
  \node[font=\bfseries\small] at (0,-0.8) {\texttt{.txt}};
\end{tikzpicture}

%% file: tables/dataset/comp_existing_v2.tex
\begin{table*}[h!tb]
    \centering
    \setlength\tabcolsep{4pt}
    \caption{
    Comparison of existing \ac{RS} image-text datasets. Datasets are compared in terms of: 1) the presence of co-registered multi-sensor imagery exceeding three spectral bands; 2) methods used for text annotation, including the presence of an extensive manual quality check (QC); 3) supported tasks; 4) availability of geolocation data and 5) the number of image-text (IT) samples. 
    \rcmark{} indicates that only parts of the dataset fulfill this property.
    }
    \begin{tabular}{l >{\centering\arraybackslash}p{0.5cm} >{\centering\arraybackslash}p{0.8cm} >{\centering\arraybackslash}p{0.5cm} c >{\centering\arraybackslash}p{0.5cm} >{\centering\arraybackslash}p{0.5cm} >{\centering\arraybackslash}p{0.5cm} >{\centering\arraybackslash}p{0.5cm} >{\centering\arraybackslash}p{0.5cm} c >{\centering\arraybackslash}p{0.5cm} >{\centering\arraybackslash}p{0.5cm} >{\centering\arraybackslash}p{0.8cm} c r}
    \toprule
     \multirow{2}{*}{\makecell{\\[14pt]Dataset}}     & \multicolumn{3}{c}{\makecell{Multi-Sensor\\[-4pt]Images}} && \multicolumn{5}{c}{Annotation Method} && \multicolumn{3}{c}{\makecell{Supported\\[-4pt]Tasks}} & \multirow{2}{*}{\rb{Geolocation data\gap{8pt}}} & \multirow{2}{*}{\rb{\#IT samples\gap{7pt}}}\\
     \cmidrule{2-4} \cmidrule{6-10} \cmidrule{12-14}
                                                     & \multicolumn{1}{c}{\rb{Available}} & \multicolumn{1}{c}{\rb{\makecell{>3 bands\\[-4pt]per image}}} & \multicolumn{1}{c}{\rb{Co-registered}} && \multicolumn{1}{c}{\rb{Manual}} & \multicolumn{1}{c}{\rb{Template}} & \multicolumn{1}{c}{\rb{Web scraped}} & \multicolumn{1}{c}{\rb{LLM-based}} & \multicolumn{1}{c}{\rb{Manual QC}} && \multicolumn{1}{c}{\rb{Captioning}} & \multicolumn{1}{c}{\rb{VQA}} & \multicolumn{1}{c}{\rb{\makecell{Ref. Exp.\\[-4pt]Detection}}}\\
\midrule
     UCM-Captions~\cite{qu2016deep}                  & \rxmark & \rxmark & \rxmark && \gcmark & \rxmark & \rxmark & \rxmark & \gcmark && \gcmark & \rxmark & \rxmark & \rxmark & 10.5k \\
     Sydney-Captions~\cite{qu2016deep}               & \rxmark & \rxmark & \rxmark && \gcmark & \rxmark & \rxmark & \rxmark & \gcmark && \gcmark & \rxmark & \rxmark & \rxmark & 3k \\
     RSICD~\cite{lu2017exploring}                    & \rxmark & \rxmark & \rxmark && \gcmark & \rxmark & \rxmark & \rxmark & \gcmark && \gcmark & \rxmark & \rxmark & \rxmark & 54.6k \\
     RSITMD~\cite{yuan2022exploring}                 & \rxmark & \rxmark & \rxmark && \gcmark & \rxmark & \rxmark & \rxmark & \gcmark && \gcmark & \rxmark & \rxmark & \rxmark & 23.7k \\
     NWPU-Captions~\cite{cheng2022nwpu}              & \gcmark & \rxmark & \rxmark && \gcmark & \rxmark & \rxmark & \rxmark & \gcmark && \gcmark & \rxmark & \rxmark & \rxmark & 157k \\
     GAIA~\cite{gaia_zavras}                         & \gcmark & \rxmark & \rxmark && \rxmark & \rxmark & \gcmark & \gcmark & \rxmark && \gcmark & \rxmark & \rxmark & \rcmark & 205k \\
     Git-10M~\cite{Text2Earth_git_10m}               & \gcmark & \rxmark & \rxmark && \rxmark & \rxmark & \rxmark & \gcmark & \rxmark && \gcmark & \rxmark & \rxmark & \gcmark & 10.5M \\
     RSVQAxBEN~\cite{lobryRSVQAMeetsBigearthnet2021} & \rxmark & \rxmark & \rxmark && \rxmark & \gcmark & \rxmark & \rxmark & \rxmark && \rxmark & \gcmark & \rxmark & \rxmark & 14.8M \\
     RS5M~\cite{zhang2024rs5m}                       & \gcmark & \rxmark & \rxmark && \rxmark & \rxmark & \rcmark & \rcmark & \rxmark && \gcmark & \rxmark & \rxmark & \rcmark & 5.0M \\
     RSTeller~\cite{ge2025rsteller}                  & \rxmark & \rxmark & \rxmark && \rxmark & \rxmark & \rxmark & \gcmark & \rxmark && \gcmark & \rxmark & \rxmark & \rxmark & 2.6M \\
     Landsat30-AU~\cite{ma2025landsat30}             & \rxmark & \rxmark & \rxmark && \rxmark & \rxmark & \rxmark & \gcmark & \rcmark && \gcmark & \gcmark & \rxmark & \rcmark & 216k \\
     ChatEarthNet~\cite{ChatEarthNet}                & \rxmark & \gcmark & \rxmark && \rxmark & \rxmark & \rxmark & \gcmark & \rcmark && \gcmark & \rxmark & \rxmark & \rxmark & 173k \\
     MS-Clip~\cite{marimo2025beyond}                 & \gcmark & \gcmark & \gcmark && \rxmark & \rxmark & \rxmark & \gcmark & \rxmark && \gcmark & \rxmark & \rxmark & \rxmark & 985k \\
     \bentxt\ (ours)                                  & \gcmark & \gcmark & \gcmark && \rxmark & \gcmark & \rxmark & \gcmark & \rcmark && \gcmark & \gcmark & \gcmark & \gcmark & 9.6M \\
    \bottomrule
    \end{tabular}
    \label{tab:existing_datasets_properties}
\end{table*}

%% file: sec/2_relatedWork.tex
\Acp{VLM} in \ac{RS} enable natural language interactions with images across multiple tasks, including \ac{VQA}, change captioning, and \genDet{} \cite{kuckreja2024geochat, irvin2025teochat, soni2025earthdial}. To facilitate the research and development of \acp{VLM}, several image-text datasets have been proposed in RS.
Early \ac{RS} image-text datasets, such as UCM-Captions, Sydney-Captions \cite{qu2016deep}, RSICD \cite{lu2017exploring}, RSITMD \cite{yuan2022exploring}, and NWPU-Captions \cite{cheng2022nwpu} contain high-resolution aerial imagery with human-written single-sentence descriptions with limited intra-class diversity. 
To overcome this bottleneck, recently proposed datasets shift toward large-scale, automatically generated texts: RS5M \cite{zhang2024rs5m} introduces the first million-scale corpus with 5 million image-caption pairs; GAIA \cite{gaia_zavras} provides \num{205150} pairs, with multi-sentence captions emphasizing environmental dynamics; and Landsat30-AU \cite{ma2025landsat30} contains \num{196262} image-caption pairs and \num{17725} \ac{MC} \acp{VQA}.
For \ac{VQA} specifically, RSVQA \cite{lobry2020rsvqa} comprises \num{77232} and \num{1066316} question-answer triplets for low- and high-resolution imagery, respectively, while RSVQAxBEN \cite{lobryRSVQAMeetsBigearthnet2021} scales to approximately \num{15} million triplets with \ac{S2} BigEarthNet~\cite{sumbul2019bigearthnet} \ac{RGB} composites and \acp{VQA} for \ac{CLC} class presence.
Recent studies also extend the spectral range of image-text datasets beyond \ac{RGB}. 
Yuan et al.~\cite{ChatEarthNet} generate captions for nine-band \ac{S2} images, yet descriptions rely on broad \ac{LULC} classes from ESA WorldCover~\cite{zanaga2022esaworldcover} (\eg water, tree, snow). Marimo et al.~\cite{marimo2025beyond} scale to \num{975000} co-registered \ac{S1}-\ac{S2} images from SSL4EO~\cite{wang2023ssl4eo} but generate captions using a \ac{VLM} that receives \ac{RGB} bands only, preventing texts from relying on information beyond the visible spectrum. 

%% file: sec/3_bigearth_txt.tex
To address the scarcity of large-scale multi-sensor image-text datasets in \ac{RS}, we introduce \bentxt{}, which comprises co-registered \ac{S1} and \ac{S2} images with semantically rich natural-language annotations relevant for 15 tasks across 4 broad categories: i)~captioning; ii)~binary and iii)~multiple-choice \ac{VQA}; and iv)~\genDet. The co-registered \ac{S1} and \ac{S2} images in \bentxt{} are taken from BigEarthNet~v2.0~\cite{clasen2025reben}, which comprises \num{549488} pairs of \ac{S1} and \ac{S2} images acquired over ten European countries. Each pair is accompanied by a pixel-level \ac{LULC} reference map based on the \ac{CLC} 2018 product  (\texttt{V2020\_20u1})~\cite{clcbook}. Some images in BigEarthNet~v2.0 are either partly covered by seasonal snow, clouds, and cloud shadows, or their respective reference map contains unclassified pixels (with no LULC label). To construct our dataset, we filter out such image pairs, resulting in \num{464044} co-registered \ac{S1} and \ac{S2} images, with an average acquisition time difference of \SI{20}{\hour} and a max. of \SI{134}{\hour} per image pair. In the following, we describe in detail our annotation generation pipeline, present the caption statistics and lastly introduce our \bench{}.

\subsection{Annotation Generation Pipeline}
The textual annotations for \bentxt{} are generated through a three-stage pipeline (template-based caption generation, \ac{LLM}-based linguistic augmentation, \ac{VQA} and \ac{LULC} annotation generation). 
An overview of the caption generation and linguistic augmentation stage is given in \cref{fig:caption_generation_process}.
\paragraph{Template-based caption generation.} We first construct captions by extracting four categories of spatial attributes directly from the reference maps: the \textit{presence} of \ac{LULC} classes, the \textit{count} of individual contiguous regions per class (instances), the \textit{size} of each class in total and per instance, and pairwise spatial \textit{adjacency} between classes. 
Area values are rounded to the nearest \SI{1000}{\meter\squared} to reduce label noise. 
Classes are grouped into three categories according to the proportion of the image they occupy:
i)~\textit{primary} (> \SI{25}{\percent}); 
ii)~\textit{secondary} (\SI{5}{\percent} - \SI{25}{\percent}); or
iii)~\textit{marginal} (<\SI{5}{\percent}).
Captions are composed from pre-defined templates that contain classes, sizes, and adjacency relations accordingly. 
Each caption is further grounded in additional context by appending the acquisition season, country, and Köppen-Geiger climate zone~\cite{beckHighresolution1Km2023}, derived from high-resolution climate maps.

\paragraph{LLM-based linguistic augmentation.} Although template-based captions guarantee factual correctness, they exhibit limited linguistic variety. 
Therefore, we apply a two-stage augmentation using the quantized Llama-4-Scout-17B model\chdeleted{\cite{touvron2023llamaopenefficientfoundation}}.
Initially, a \textit{paraphrasing} step diversifies lexical and syntactic structure while prohibiting the addition of unsupported information.
Then, a \textit{self-refinement} step \cite{self-refinement} revises the paraphrased output considering the original template to eliminate hallucinated content and restore missing information. 
To further increase valid lexical variation, the prompt includes the \ac{CLC} nomenclature, permitting semantically valid substitutions (\eg referring to \emph{Urban fabric} collectively as \emph{Artificial surfaces}) (see supplementary material for used prompts). 
Manual evaluation of \num{3209} randomly sampled augmented captions against four binary criteria (linguistic correctness, factual accuracy, completeness, and absence of generation artifacts) yields an average correctness of \SI{93.76}{\percent}, with \SI{77.50}{\percent} of captions satisfying all four criteria simultaneously.
\input{figures/generation_process_v6}
\paragraph{VQA and \genDet{} generation.} We further augment each image pair with question-answer annotations that are relevant for binary yes/no \ac{VQA}, \ac{MCQ} \ac{VQA}, and \genDet. 
Binary questions target the four spatial categories from the captioning stage (presence, count, size, adjacency), with one \enquote{yes} and one \enquote{no} answer generated per image pair. 
To prevent \enquote{no} answers from being solvable by class-absence detection alone, count and size questions are constructed such that the queried class is present but the stated quantity is incorrect; presence and adjacency \enquote{no} questions use semantically similar classes from the \ac{CLC} hierarchy. 
\Acp{MCQ} extend the four spatial categories from the captioning stage with questions about relative class position, acquisition country, season, and Köppen-Geiger climate zone.
Each set of answers for the \acp{MCQ} includes one correct answer and three incorrect ones, sampled by analogous principles to the binary \acp{VQA}. 
For \genDet, we generate \refDet{} instructions targeting instances covering between \SI{1}{\percent} and \SI{50}{\percent} of the image area, and also \pointDet{} instructions where the center point lies inside the targeted instance. 
Furthermore, we filter out bounding boxes where the targeted instance covers less than \SI{40}{\percent} of the bounding box area.
Examples are given in \cref{fig:dataset_overview}.
Each annotation type uses more than 20 linguistic templates to ensure variability without relying on \ac{LLM} augmentation (see \suppref{sec:supp_templates}). 
In total, each image pair is associated with up to 16 \ac{VQA} pairs, and approximately \SI{80}{\percent} of pairs include at least one \genDet{}.

\subsection{Caption Statistics\label{sec:statistics}}
\begin{figure}[h!tb]
\centering
    \begin{subfigure}[b]{\linewidth}
        \phantomsubcaption\label{fig:pdf_caption_length}
        \raisebox{14mm}{(a)}\hspace{2mm}
        \raisebox{0mm}{\rotatebox{90}{\hspace{4mm}\textsf{\notsotiny Density of Occurrences}}}
        \includegraphics[width=0.78\linewidth,trim=30 32 0 0, clip]{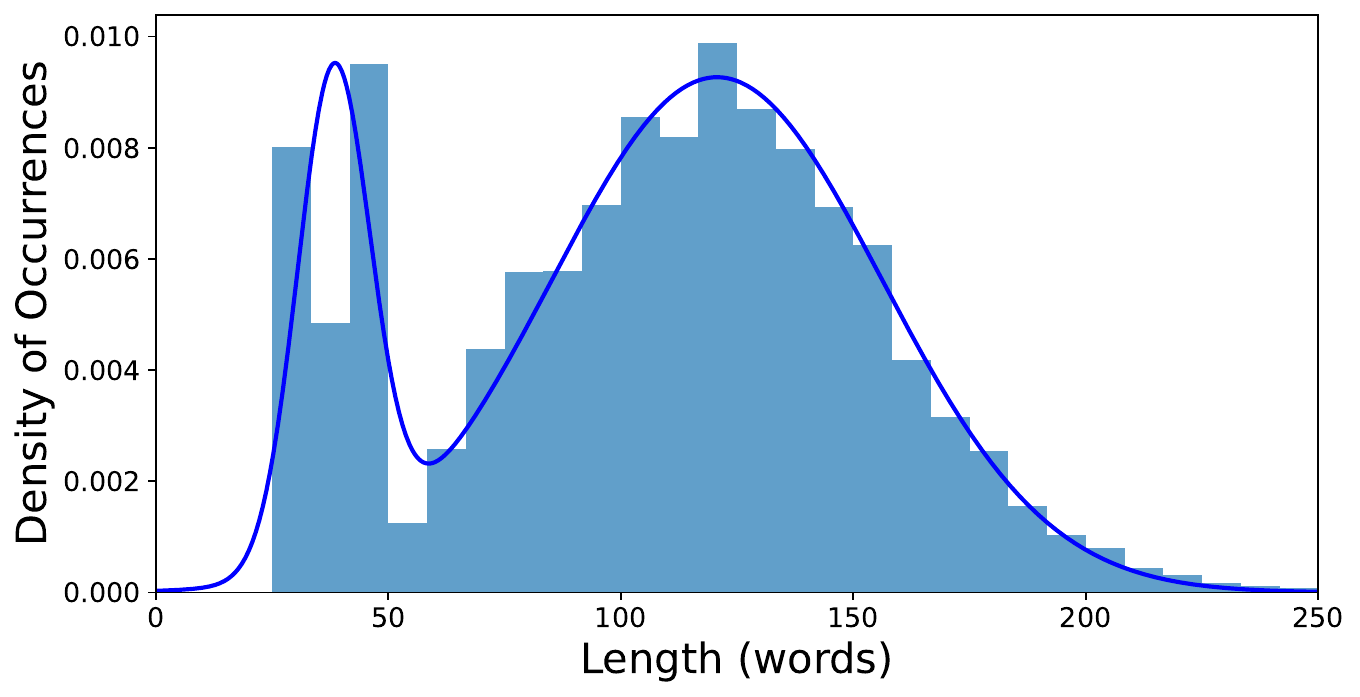}
    \end{subfigure}
    \raisebox{2mm}{\textsf{\notsotiny Length ($\nicefrac{\text{Words}}{\text{Caption}}$)}}
    \hfill
    \begin{subfigure}[b]{\linewidth}
        \phantomsubcaption\label{fig:pdf_sentences_per_caption}
        \raisebox{14mm}{(b)}\hspace{2mm}
        \raisebox{0mm}{\rotatebox{90}{\hspace{4mm}\textsf{\notsotiny Density of Occurrences}}}
        \includegraphics[width=0.75\linewidth,trim=30 32 0 0, clip]{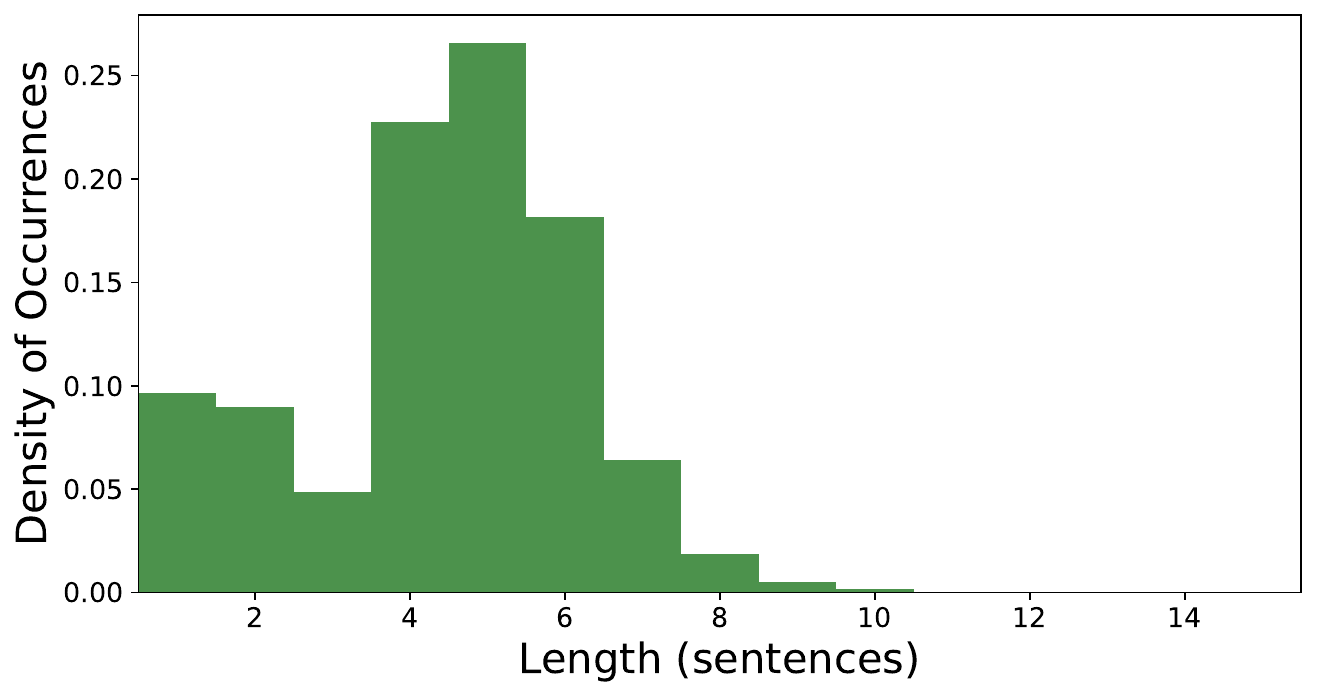}
    \end{subfigure}
    \raisebox{2mm}{\textsf{\notsotiny Length ($\nicefrac{\text{Sentences}}{\text{Caption}}$)}}
    \hfill
    \begin{subfigure}[b]{\linewidth}
        \phantomsubcaption\label{fig:mtld_score}
        \raisebox{22mm}{(c)}\hspace{2mm}
        \raisebox{0mm}{\rotatebox{90}{\hspace{15mm}\textsf{\notsotiny MTLD Score}}}
        \includegraphics[width=0.83\linewidth,trim=30 30 30 0, clip]{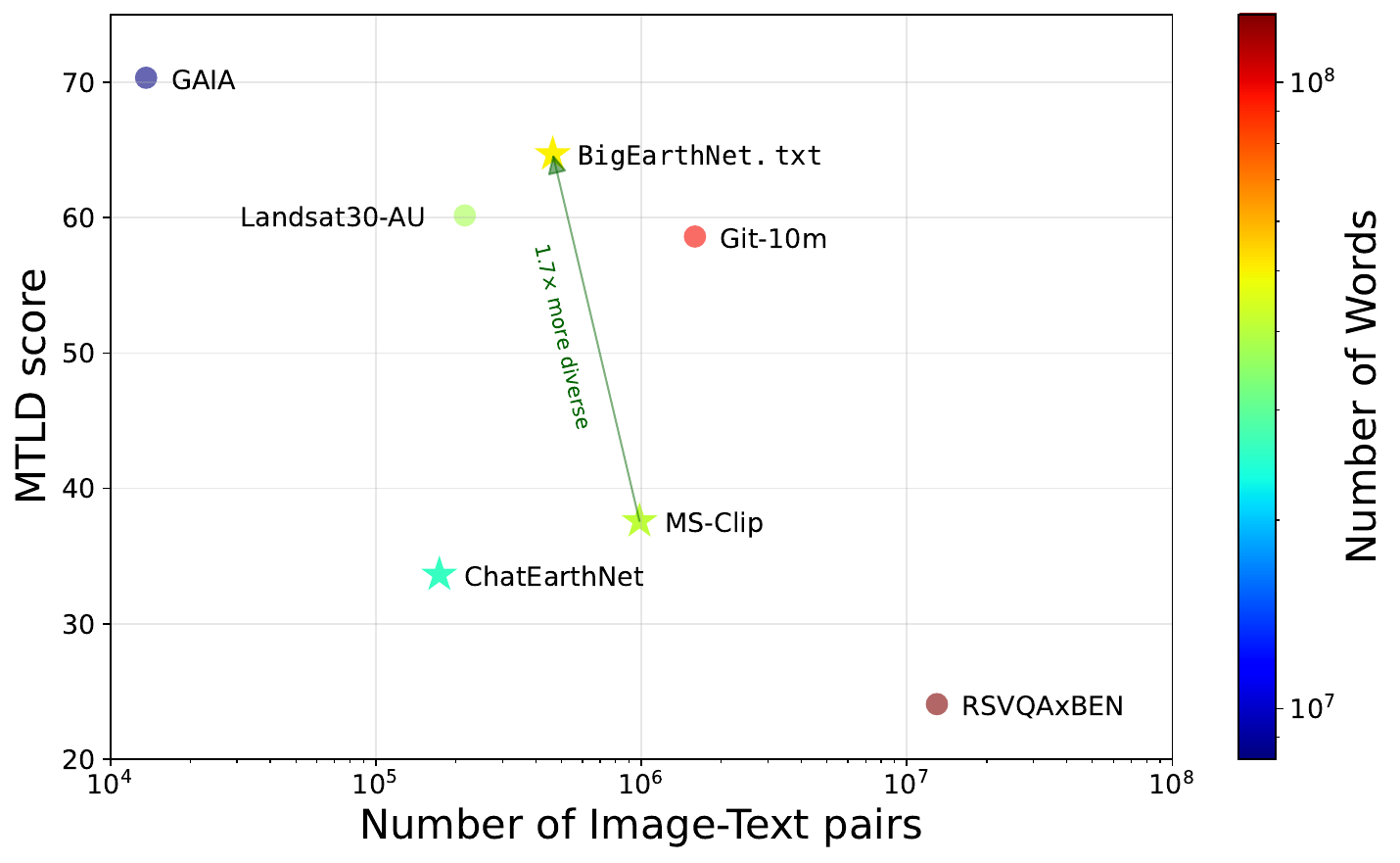}
        \raisebox{0mm}{\rotatebox{90}{\hspace{13mm}\textsf{\notsotiny Number of Words}}}
    \end{subfigure}
    \raisebox{2mm}{\textsf{\notsotiny Number of Image-Text Pairs}}
    \caption{\Ac{PDF} of (a) caption lengths and (b) number of sentences in a caption in the \bentxt{} dataset.
     (c) Comparison of existing image-text \ac{RS} datasets in terms of size and semantic richness of the text data. The \ac{NLTK} word tokenizer is used to divide the captions and questions into tokens. The color of each point indicates the total number of tokens in the dataset. Datasets that encompass images with more than three spectral bands are denoted with a star. The \bentxt{} dataset is $1.7\times$ more diverse compared to the largest existing \ac{RS} dataset encompassing more than three bands.}
    \label{fig}
\end{figure}

The caption annotations of \bentxt{} contain approximately \num{50} million words and \num{2.1} million sentences, averaging 107 words and \num{4.5} sentences per caption. 
The vocabulary comprises \num{12394} unique terms. 
Caption lengths follow a bimodal distribution (\cref{fig:pdf_caption_length}) corresponding to single-class and multi-class scenes: single-class captions are shorter due to the absence of adjacency descriptions, while multi-class captions vary with scene complexity. 
Sentence counts per caption (\cref{fig:pdf_sentences_per_caption}) peak between four and six, reflecting the predominance of complex scenes. 
As shown in \cref{fig:mtld_score}, the caption annotations of \bentxt{} achieve a \ac{MTLD} score~\cite{mccarthy2005assessment} of \num{64.69}, surpassing the largest existing image-text \ac{RS} dataset containing images with more than three bands (MS-CLIP~\cite{marimo2025beyond}) by more than $1.7\times$. 
Additionally, caption annotations of \bentxt{} encompass $\sim$\SI{25}{\percent} more words in half as many samples compared to MS-CLIP. This shows that \bentxt{} is not only lexically more diverse, but also richer in the semantic content per caption.

\subsection{The BigEarthNet.txt \bench{}}

We divide \bentxt{} into train, validation and test splits as in BigEarthNet~v2.0, resulting in \num{229114}, \num{118095}, and \num{116835} image pairs and \num{4674281}, \num{2454690}, and \num{2424991} text annotations, respectively. However, because LLM-augmented captions may contain hallucinated content, we construct a curated \bench{} to enable a reliable model evaluation on the annotations introduced in \bentxt{}. The \bench{} contains \num{15029} text annotations for \num{1082} image pairs from the test split whose augmented captions passed manual verification on all four quality dimensions.
To expose answer-bias in evaluated models, binary and \ac{MCQ} annotations are balanced across answer options.
Annotations are further balanced across \ac{LULC} classes to the extent permitted by their natural distribution. 
Overall, the \bench{} contains \num{6927} binary and \num{5550} multiple-choice annotations, \num{970} captions, and \num{1582} \genDet{} annotations\chdeleted{ across \num{1082} image pairs}.

%% file: figures/generation_process_v6.tex
\begin{figure}[!ht]
    \centering
    \resizebox{\linewidth}{!}{
    \begin{tikzpicture}        
        \node[inner sep=0pt,minimum size=0pt] (CaptionAnchor) at (-4.7, 0) {};
        \node[single arrow,
              minimum height=65mm, minimum width=30mm,
              single arrow head extend=2mm,
              anchor=west, rotate=-90, fill=gray!15] at (1.33,-4) {};
        \node[anchor=west] (refMap) at (0,0)
            {\includegraphics[width=\BenImgWidth]{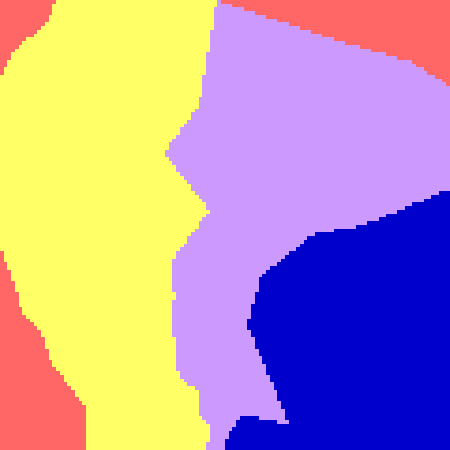}};
        \node[anchor=south] (refMapCap) at ($(refMap.north) + (0,-.2)$) {\small\textcolor{gray}{Reference map}};

        \node[align=justify, anchor=south east, text width=30mm, minimum width=30mm, minimum height=20mm, fill=cyan!20, rounded corners=2mm, font=\normalsize] (PresenceSizeTemplate) at ($(refMap.north west) + (-.2,.7)$) {This image primarily shows \textbf{[class]} (\textbf{[size]})... Smaller areas are occupied by \textbf{[class]} (\textbf{[size]}) and \textbf{[class]} (\textbf{[size]}).};
        
        \node[align=justify, anchor=south west, text width=30mm, minimum width=30mm, minimum height=20mm, fill=WildStrawberry!20, rounded corners=2mm, font=\normalsize] (SizeCountTemplate) at ($(refMap.north east) + (.2,.2)$) {\textbf{[Class]} is distributed over \textbf{[count]} individual areas, where \textbf{[count]} are smaller (\textbf{[size]} and \textbf{[size]}) and ... and \textbf{[count]} are marginal.};
        
        \node[align=justify, anchor=north west, text width=30mm, minimum width=30mm, minimum height=15mm, fill=Dandelion!20, rounded corners=2mm, font=\normalsize] (AdjacencyTemplate) at ($(refMap.south east) + (.2,-.2)$) {\textbf{[Class]} is adjacent to \textbf{[class]}. Also, ... and \textbf{[class]} borders \textbf{[class]}.};
        
        \node[align=justify, anchor=north east, text width=30mm, minimum width=30mm, minimum height=15mm, fill=LimeGreen!20, rounded corners=2mm, font=\normalsize] (ClimateLocationTemplate) at ($(refMap.south west) + (-.2,-.2)$) {The image was taken during \textbf{[season]} in \textbf{[location]} in the \textbf{[climate zone]}.};

        \node[anchor=center, align=left, font=\normalsize] (Size) at ($(PresenceSizeTemplate)!0.5!(SizeCountTemplate) + (0, -.1)$) {
            \legendbox{ArableLand211}{\SI{526000}{\meter\squared}}\\
            \legendbox{InlandWetlands411}{\SI{460000}{\meter\squared}}\\
            \legendbox{InlandWaters512}{\SI{305000}{\meter\squared}}\\
            \legendbox{UrbanFabric112}{marginal}\\
            \legendbox{UrbanFabric112}{marginal}\\
            \legendbox{UrbanFabric112}{marginal}
        };
        \node[anchor=south, align=center, font=\normalsize] (SizeCap) at ($(Size.north) + (0,-.15)$) {\textcolor{gray}{Size}};

        \node[anchor=center, align=left, font=\normalsize] (Presence) at ($(PresenceSizeTemplate)!0.5!(ClimateLocationTemplate) + (0, 0.3)$) {
            \legendbox{UrbanFabric112}{Urban fabric}\\
            \legendbox{ArableLand211}{Arable land}\\
            \legendbox{InlandWetlands411}{Inl. Wetlands}\\
            \legendbox{InlandWaters512}{Inl. water}
        };
        \node[anchor=south, align=left, rotate=90, font=\normalsize] (PresenceCap) at ($(Presence.west) + (0,0)$) {\textcolor{gray}{Presence}};

        \node[anchor=center, align=left, font=\normalsize] (Count) at ($(SizeCountTemplate)!0.5!(AdjacencyTemplate) + (0, 0.3)$) {
            \legendbox{ArableLand211}{\small $\times$ 1}\\
            \legendbox{InlandWetlands411}{\small $\times$ 1}\\
            \legendbox{InlandWaters512}{\small $\times$ 1}\\
            \legendbox{UrbanFabric112}{\small $\times$ 3}
        };
        \node[anchor=south, align=center,rotate=-90, font=\normalsize] (CountCap) at ($(Count.east) + (0,0)$) {\textcolor{gray}{Count}};

        \node[anchor=center] (Adjacency) at ($(ClimateLocationTemplate)!0.5!(AdjacencyTemplate) + (0, -.1)$) {};
        \node[anchor=center, align=left, circle, minimum size=5pt, inner sep=0pt, fill=UrbanFabric112] (UF1) at ($(Adjacency) + (-.7,.3)$) {};
        \node[anchor=center, align=left, circle, minimum size=5pt, inner sep=0pt, fill=UrbanFabric112] (UF2) at ($(Adjacency) + (-.7,-.3)$) {};
        \node[anchor=center, align=left, circle, minimum size=5pt, inner sep=0pt, fill=UrbanFabric112] (UF3) at ($(Adjacency) + (.7,.3)$) {};
        \node[anchor=center, align=left, circle, minimum size=5pt, inner sep=0pt, fill=ArableLand211] (AL) at ($(Adjacency) + (-.25,0)$) {};
        \node[anchor=center, align=left, circle, minimum size=5pt, inner sep=0pt, fill=InlandWetlands411] (IWl) at ($(Adjacency) + (.25,0)$) {};
        \node[anchor=center, align=left, circle, minimum size=5pt, inner sep=0pt, fill=InlandWaters512] (IWa) at ($(Adjacency) + (.7,-.3)$) {};
        \node[anchor=north, align=center, font=\normalsize] (AdjacencyLabel) at ($(Adjacency.south) + (0,-.2)$) {\textcolor{gray}{Adjacency}};
        \draw[thick] ($(UF1)!0.3!(AL)$) -- ($(UF1)!0.7!(AL)$);
        \draw[thick] ($(UF2)!0.3!(AL)$) -- ($(UF2)!0.7!(AL)$);
        \draw[thick] ($(AL)!0.3!(IWl)$) -- ($(AL)!0.7!(IWl)$);
        \draw[thick] ($(IWl)!0.3!(IWa)$) -- ($(IWl)!0.7!(IWa)$);
        \draw[thick] ($(IWl)!0.3!(UF3)$) -- ($(IWl)!0.7!(UF3)$);

        \node[anchor=south east] (loc) at ($(ClimateLocationTemplate.north) + (-.3,-.1)$) {\includegraphics[height=0.3\BenImgWidth]{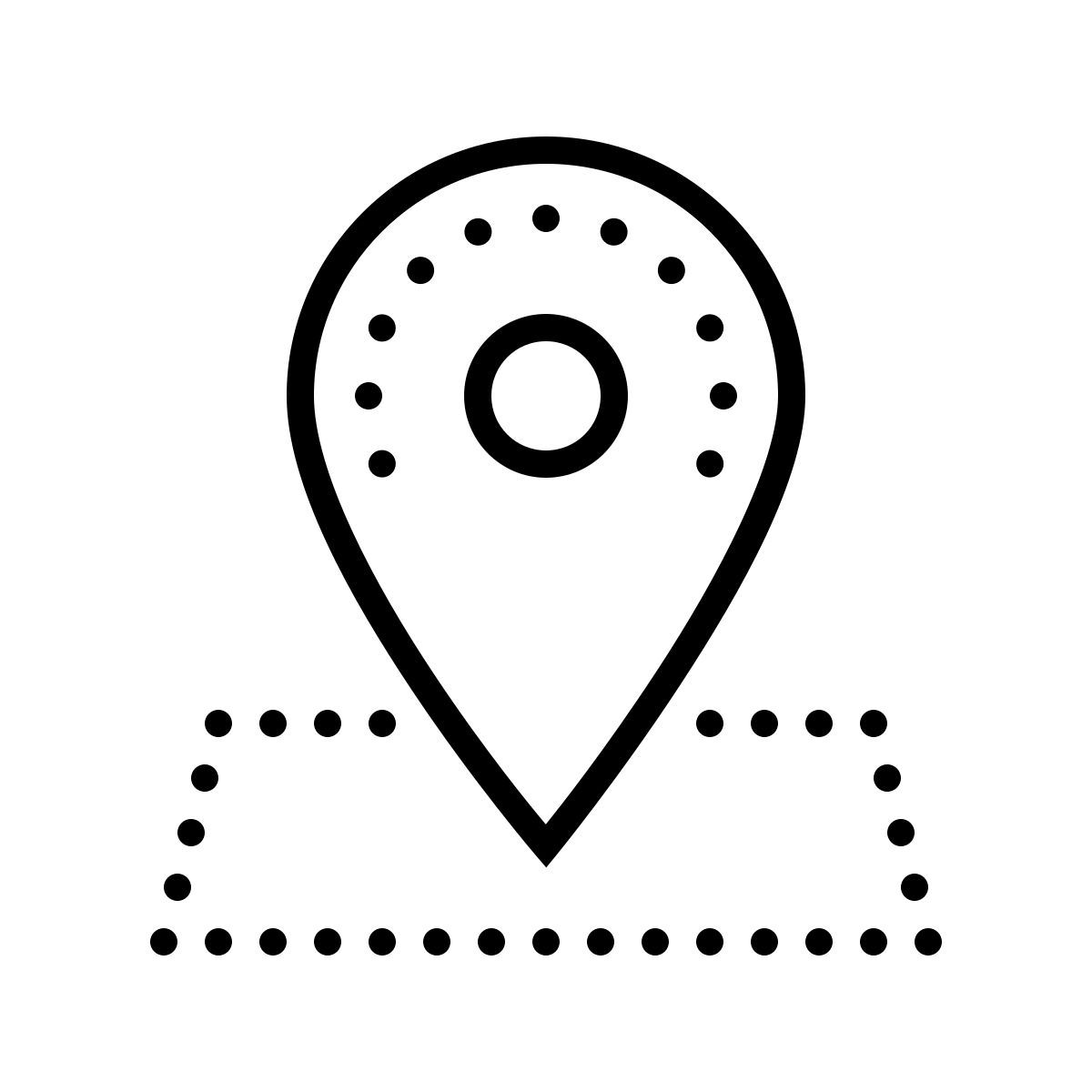}};
        \node[anchor=south, font=\normalsize] (LocCap) at ($(loc.north) + (0,-.2)$) {\textcolor{gray}{Location}};
        \node[anchor=south west] (KGMap) at ($(ClimateLocationTemplate.north) + (-.1,-.1)$) {\includegraphics[height=0.3\BenImgWidth]{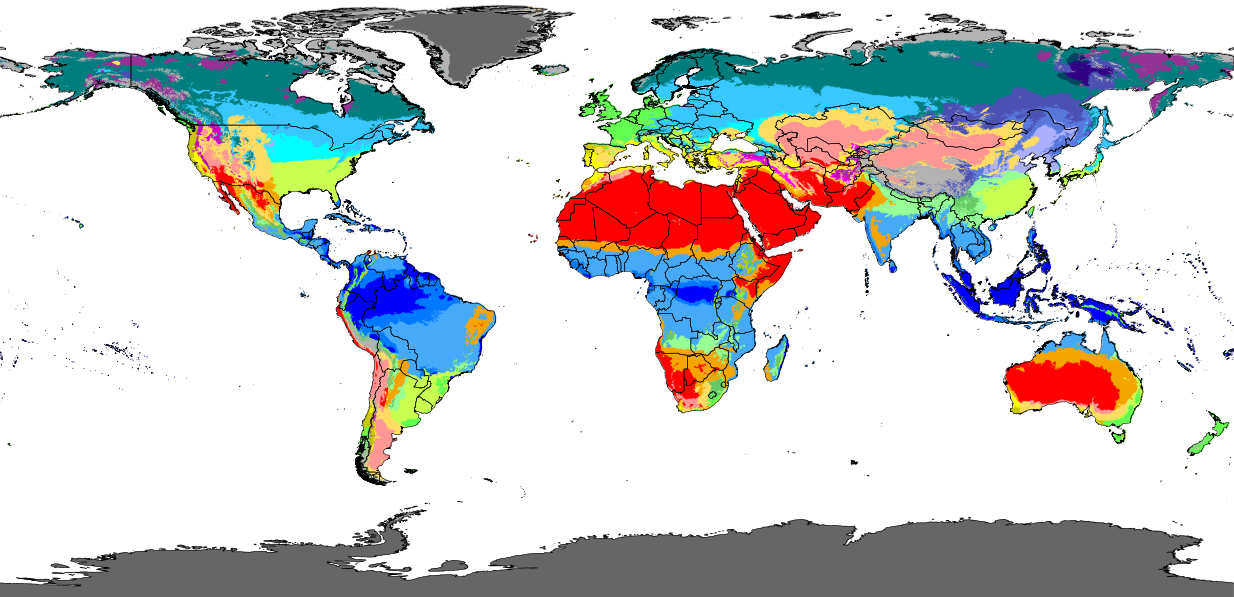}};
        \node[anchor=center, font=\normalsize] (KGMapCap) at ($(KGMap |- LocCap) + (0,0)$) {\textcolor{gray}{Climate}};
        \draw[thin,dotted] ($(PresenceCap.north west) + (0, -0.3)$) -- ($(KGMap.north east |- PresenceCap.west) + (0.05, -0.3)$) -- ($(ClimateLocationTemplate.south east) + (0.05, -.0)$);

        \node[single arrow,minimum height=12mm, minimum width=8mm, single arrow head extend=2mm,
            anchor=west, rotate=90, fill=cyan!90, opacity=0.5] at ($(refMap.north) + (0,-.3)$) {};
        \node[single arrow,minimum height=12mm, minimum width=8mm, single arrow head extend=2mm,
            anchor=west, rotate=0, fill=cyan!90, opacity=0.5] at ($(refMap.east) + (-.3,.4)$) {};
        \node[single arrow,minimum height=12mm, minimum width=8mm, single arrow head extend=2mm,
            anchor=west, rotate=-90, fill=cyan!90, opacity=0.5] at ($(refMap.south) + (0,.3)$) {};
        \node[single arrow,minimum height=10mm, minimum width=8mm, single arrow head extend=2mm,
            anchor=west, rotate=180, fill=cyan!90, opacity=0.5] at ($(refMap.west) + (.3,.4)$) {};

        \node[draw, rounded corners=2mm, thick, dotted, inner sep=4pt, fit=(ClimateLocationTemplate)(SizeCountTemplate), gray!70] (box1) {};

        \node[align=center, anchor=north, text width=99mm, font=\normalsize] (CombinedTemplates)
            at ($(box1.south) + (0,-0.05)$)
            {
                \colorbox{cyan!20}{This image primary ...  (\textbf{[size]}).}
                \colorbox{WildStrawberry!20}{\textbf{[Class]} is distributed ...}
                \colorbox{Dandelion!20}{\textbf{[Class]} is adjacent ... \textbf{[class]}.}
                \colorbox{LimeGreen!20}{The image was ... \textbf{[climate zone]}.}
            };
        \node[draw, rounded corners=2mm, thick, dotted, inner sep=-1pt, fit=(CombinedTemplates), gray!70] (CombinedTemplatesBox) {};
        \node[draw, rounded corners=2mm, thin, dashed, inner sep=2pt, fit=(CombinedTemplatesBox)(box1), gray!90] (TemplatesBox) {};
        \node[inner sep=0pt,minimum size=0pt] (CombineAnchor) at ($(box1.north)!0.5!(CombinedTemplatesBox.south)$) {};
        \node[anchor=center,gray!90, rotate=90, font=\normalsize] at ($(CaptionAnchor |- TemplatesBox)$){\textcolor{gray!90}{\makecell{Extract Attributes,\\Fill \& Concatenate Templates}}};

        \def\scaleLLMRing{0.4}
        \def\LLMRingPointSize{0.05cm}
        \node (baseline_step2_1) at ($(TemplatesBox.south) + (0,-0.9)$) {};
        \node[anchor=center, draw, circle, minimum size=0.5cm, inner sep=1.5pt, fill=gray!25, font=\tiny] (LLM_paraphrase) at ($(baseline_step2_1) + (-4.15, 0)$) {\scalebox{0.8}{LLM}};
        \node[draw, circle, minimum size=\LLMRingPointSize, inner sep=0pt] (LLM_paraphrase_west) at ($(LLM_paraphrase) + (-\scaleLLMRing,0)$) {};
        \node[draw, circle, minimum size=\LLMRingPointSize, inner sep=0pt] (LLM_paraphrase_east) at ($(LLM_paraphrase) + ( \scaleLLMRing,0)$) {};
        \node[draw, circle, minimum size=\LLMRingPointSize, inner sep=0pt] (LLM_paraphrase_south) at ($(LLM_paraphrase) + (0,-\scaleLLMRing)$) {};
        \node[draw, circle, minimum size=\LLMRingPointSize, inner sep=0pt] (LLM_paraphrase_north) at ($(LLM_paraphrase) + (0, \scaleLLMRing)$) {};
        \node[draw, circle, minimum size=\LLMRingPointSize, inner sep=0pt, fill=black] at ($(LLM_paraphrase) + (-.7070*\scaleLLMRing, .7070*\scaleLLMRing)$) {};
        \node[draw, circle, minimum size=\LLMRingPointSize, inner sep=0pt, fill=black] at ($(LLM_paraphrase) + (-.7070*\scaleLLMRing, -.7070*\scaleLLMRing)$) {};
        \node[draw, circle, minimum size=\LLMRingPointSize, inner sep=0pt, fill=black] at ($(LLM_paraphrase) + (.7070*\scaleLLMRing, .7070*\scaleLLMRing)$) {};
        \node[draw, circle, minimum size=\LLMRingPointSize, inner sep=0pt, fill=black] at ($(LLM_paraphrase) + (.7070*\scaleLLMRing, -.7070*\scaleLLMRing)$) {};
        \centerarc[draw](LLM_paraphrase)(15:30:\scaleLLMRing)
        \centerarc[draw](LLM_paraphrase)(60:75:\scaleLLMRing)
        \centerarc[draw](LLM_paraphrase)(105:120:\scaleLLMRing)
        \centerarc[draw](LLM_paraphrase)(150:165:\scaleLLMRing)
        \centerarc[draw](LLM_paraphrase)(-15:-30:\scaleLLMRing)
        \centerarc[draw](LLM_paraphrase)(-60:-75:\scaleLLMRing)
        \centerarc[draw](LLM_paraphrase)(-105:-120:\scaleLLMRing)
        \centerarc[draw](LLM_paraphrase)(-150:-165:\scaleLLMRing)
        \node[align=justify, anchor=west, text width=80mm, font=\normalsize] (ParaphrasePrompt)
            at ($(LLM_paraphrase_east.east) + (.1,0)$)
            {\textit{You are a Remote Sensing expert. Use the following definitions to increase variance in the output: ... \\
            Do not violate these rules: ...}};
        \node[draw, rounded corners=2mm, thin, dashed, inner sep=2pt, inner ysep=2pt, fit=(LLM_paraphrase_west)(LLM_paraphrase_north)(LLM_paraphrase_south)(ParaphrasePrompt), gray!90] (box2_1) {};
        \node[anchor=center, gray!90, rotate=90, font=\normalsize] at ($(CaptionAnchor |- box2_1)$) {\textcolor{gray!90}{\makecell{Para-\\phrase}}};

        \node[anchor=center, draw, circle, minimum size=0.5cm, inner sep=1.5pt, fill=gray!25, font=\tiny] (LLM_refine) at ($(LLM_paraphrase) + (0, -1.45)$) {\scalebox{0.8}{LLM}};
        \node[draw, circle, minimum size=\LLMRingPointSize, inner sep=0pt] (LLM_refine_west) at ($(LLM_refine) + (-\scaleLLMRing,0)$) {};
        \node[draw, circle, minimum size=\LLMRingPointSize, inner sep=0pt] (LLM_refine_east)at ($(LLM_refine) + ( \scaleLLMRing,0)$) {};
        \node[draw, circle, minimum size=\LLMRingPointSize, inner sep=0pt] (LLM_refine_south) at ($(LLM_refine) + (0,-\scaleLLMRing)$) {};
        \node[draw, circle, minimum size=\LLMRingPointSize, inner sep=0pt] (LLM_refine_north) at ($(LLM_refine) + (0, \scaleLLMRing)$) {};
        \node[draw, circle, minimum size=\LLMRingPointSize, inner sep=0pt, fill=black] at ($(LLM_refine) + (-.7070*\scaleLLMRing, .7070*\scaleLLMRing)$) {};
        \node[draw, circle, minimum size=\LLMRingPointSize, inner sep=0pt, fill=black] at ($(LLM_refine) + (-.7070*\scaleLLMRing, -.7070*\scaleLLMRing)$) {};
        \node[draw, circle, minimum size=\LLMRingPointSize, inner sep=0pt, fill=black] at ($(LLM_refine) + (.7070*\scaleLLMRing, .7070*\scaleLLMRing)$) {};
        \node[draw, circle, minimum size=\LLMRingPointSize, inner sep=0pt, fill=black] at ($(LLM_refine) + (.7070*\scaleLLMRing, -.7070*\scaleLLMRing)$) {};
        \centerarc[draw](LLM_refine)(15:30:\scaleLLMRing)
        \centerarc[draw](LLM_refine)(60:75:\scaleLLMRing)
        \centerarc[draw](LLM_refine)(105:120:\scaleLLMRing)
        \centerarc[draw](LLM_refine)(150:165:\scaleLLMRing)
        \centerarc[draw](LLM_refine)(-15:-30:\scaleLLMRing)
        \centerarc[draw](LLM_refine)(-60:-75:\scaleLLMRing)
        \centerarc[draw](LLM_refine)(-105:-120:\scaleLLMRing)
        \centerarc[draw](LLM_refine)(-150:-165:\scaleLLMRing)
        \node[align=justify, anchor=west, text width=80mm, font=\normalsize] (RefinePrompt)
            at ($(LLM_refine_east.east) + (.1,0)$)
            {\textit{Control your caption. Ensure that you followed the rules. Update the caption by ... Do NOT explain the mistakes.}};
        \node[draw, rounded corners=2mm, thin, dashed, inner sep=4pt, inner ysep=2pt, fit=(LLM_refine_west)(LLM_refine_north)(LLM_refine_south)(RefinePrompt), gray!90] (box2_2) {};
        \node[anchor=center,gray!90, rotate=90, font=\normalsize] at ($(CaptionAnchor |- box2_2)$) {\textcolor{gray!90}{\makecell{Refine}}};

        \node[align=justify, anchor=north, text width=108mm, inner sep=2pt, font=\normalsize] (FinalCaption)
            at ($(box2_2.south) + (0,-0.1)$)
            {This satellite image, captured during the summer season in Switzerland, showcases a diverse landscape within the "temperate, no dry season, warm summer" climate zone. The dominant features are arable land ($\sim$526,000 sqm) and inland wetlands ($\sim$460,000 sqm), which are adjacent to each other. The arable land borders both inland wetlands and urban fabric ($\sim$149,000 sqm). Moreover, the inland wetlands are adjacent to both inland waters ($\sim$305,000 sqm) and urban fabric. Notably, the urban fabric is distributed over three individual marginal areas. The varied landscape presents a mix of agricultural areas, wetlands, water bodies, and artificial surfaces.};
        \node[draw, rounded corners=2mm, thin, dashed, inner sep=0pt, fit=(FinalCaption)($(box2_2.east |- FinalCaption)$)($(box2_2.west |- FinalCaption)$), gray!90] (box3) {};
        \node[anchor=center,gray!90, rotate=90,font=\normalsize] at ($(CaptionAnchor |- box3)$) {\textcolor{gray!90}{\makecell{Final\\Caption}}};

    \end{tikzpicture} 
    }
    \caption{Caption generation process for \bentxt{}: i) Attribute extraction from the reference map and template filling based on the attributes and metadata such as location, season and climate zone information, followed by concatenation of the templates; ii) increasing linguistic variance and ensuring correctness via self-refinement.}
    \label{fig:caption_generation_process}
\end{figure}

%% file: sec/5_experiments.tex
\subsection{Performance Evaluation of VLMs for RS Tasks}
To benchmark \bentxt{} we have considered a wide range of general-purpose \ac{CV} and specialized \ac{RS} \acp{VLM}. In detail, we select five \ac{SOTA} models each from the \ac{CV} and \ac{RS} domain. For the \ac{CV} domain, we select GPT-5.2\footnote{OpenAI API, endpoint \texttt{/v1/chat/completions}, model \enquote{gpt-5.2-2025-12-11}, temperature=0, and top\_p=0}~\cite{openai2025gpt5}, Qwen3-VL~\cite{bai2025qwen3}, GLM-4.6v-Flash~\cite{hong2025glm}, LLaVa-OneVision~\cite{li2024llavaonevision}, and InternVL3-1B~\cite{zhu2025internvl3}, denoted as \gptABBRns, \qwenABBRns, \glmABBRns, \llavaABBRns, and \internvlABBRns, respectively. For the \ac{RS} domain, we select GeoChat~\cite{kuckreja2024geochat}, LHRS-Bot~\cite{muhtar2024lhrs}, SkyEyeGPT~\cite{zhan2025skyeyegpt}, EarthDial~\cite{soni2025earthdial}, and EarthMind~\cite{shu2025earthmind}, denoted as \geochatABBRns, \lhrsbotABBRns, \skyeyegptABBRns, \earthdialrgbABBRns/\earthdialmsABBRns, and \earthmindrgbABBRns/\earthmindsarmsABBRns, respectively.
The majority of these models accept only \ac{RGB} inputs; for these, we provide only the \ac{RGB} bands of \ac{S2} during inference.
As EarthDial and EarthMind also accept multispectral and multi-sensor inputs, we evaluate each under two configurations:
i) \earthdialrgbABBRns{} and \earthmindrgbABBRns{} using \ac{RGB} inputs; and
ii) \earthdialmsABBRns{} and \earthmindsarmsABBRns{} using twelve \ac{S2} bands and two \ac{S1} and twelve \ac{S2} bands as input, respectively.
We provide the \bentxt{} instructions by default, with additional instructions to follow the output format and model specific tokens where necessary. 
We extract answers even when they do not adhere exactly to the specified format. 
The final instructions used for each model are provided in \suppref{sec:supp_evalprompts}.
To reduce computational resources during evaluation, we disable thinking for \glmABBRns.
Throughout the paper, we report the results on the \bench{} only, whereas additional results on the complete test split of \bentxt{} are provided in \suppref{sec:supp_testresults}.

\cref{tab:main} shows the results obtained for each of our general categories: captioning, binary \ac{VQA}, \ac{MCQ}, and \genDet{}, grouped by \ac{RS} specialized \acp{VLM} (top) and general-purpose \ac{CV} \acp{VLM} (bottom).
From the table, one can observe that none of the \acp{VLM} in general provides good performance, while \ac{CV} \acp{VLM} perform better than \ac{RS} \acp{VLM} in general despite processing only \ac{RGB} input.
This is because \ac{CV} \acp{VLM} exhibit stronger generalizability and instruction-following ability (Tab.~\ref{tab:binary_results}--\ref{tab:bb_point_results}), whereas most \ac{RS} \acp{VLM} are mainly suitable for tasks achievable using only \ac{RGB} data.
\input{tables/results/main_table}
The considered \ac{CV} \acp{VLM} may also have been pretrained using some \ac{RS} data, which is a growing trend in recent works~\cite{simeoni2025dinov3}.
Additionally, models accepting multispectral (\earthdialmsABBRns) or multi-sensor (\earthmindsarmsABBRns) inputs show no consistent benefit (and in some cases decreased performance) compared to their \ac{RGB} counterparts.
This is because their pre-training and fine-tuning data largely consist of \ac{RGB} images, and the models were therefore not trained to exploit the additional spectral information provided at inference.
Even \gptABBR{}, which is the largest model, scores only 60.39\% on binary \ac{VQA} and 34.93\% on \ac{MCQ} due to its inherent \ac{RGB} data limitation.
\cref{tab:binary_results} shows detailed per-task results for binary \ac{VQA}.
From the table, one can clearly observe that for some tasks, models perform noticeably better (\eg, \SI{69.34}{\percent} accuracy for EarthMind on the binary \emph{presence} task).
\input{tables/results/binary}
This good result on presence \ac{VQA} is expected, as determining whether a \ac{LULC} class appears in an image is closely analogous to scene classification, which is a dominant objective in \ac{RS} \ac{VLM} pre-training datasets~\cite{kuckreja2024geochat, muhtar2024lhrs, zhan2025skyeyegpt, soni2025earthdial, shu2025earthmind, hu2025rsgpt}.
Nevertheless, none of these models, including the largest, \gptABBRns, achieve consistently strong performance across all binary \ac{VQA} tasks.
In \cref{tab:mcq_results}, one can observe the performance on the \ac{MCQ} tasks.
\input{tables/results/mcq}
These results closely mirror the binary \ac{VQA} findings: the best performance is again observed on tasks that most closely resemble typical pre-training objectives.
Beyond understanding the image content, instruction-following is an additional challenge: most \ac{RS} models struggle to consistently adhere to the \ac{MCQ} format, as this task is less common in \ac{RS} \ac{VLM} pre-training data and thus less reliably learned.
The results on \refDet{} and \pointDet{} are presented in \cref{tab:bb_reference_results} and \cref{tab:bb_point_results}.
Since EarthMind was trained on segmentation rather than bounding box prediction, we derive bounding boxes from their segmentation outputs.
In \refDet, models must identify and localize a target instance from a textual description alone, without any spatial prior.
In \pointDet, models are additionally provided with a point within the target instance and must predict the enclosing bounding box.
As one can see from the table, several models from the \ac{CV} domain perform substantially better on \pointDet{} than on \refDet.
We think that the provided point allows models with strong general spatial reasoning to constrain the predicted box to the vicinity of the given location, reducing the task to local boundary estimation rather than open-vocabulary instance search.
GPT achieves the best performance on both task types, mainly owing to its scale and the breadth of its pre-training data.
\input{tables/results/bbox_reference}
\input{tables/results/bbox_point}
Finally, \cref{tab:captioning} shows the performance obtained in the captioning task in several metrics.
\input{tables/results/captioning}
We employ n-gram-based (BLEU, ROUGE, METEOR, CIDEr), embedding-based (BERTScore, SBERT-Cosine \cite{zhang2019bertscore, reimers-gurevych-2019-sentence}), and \ac{LLM}-based (CLAIR~\cite{chan-etal-2023-clair}) metrics.
For CLAIR, an \ac{LLM} is prompted to output a score between 0 and 100 for a candidate caption based on the probability that it describes the same image as the reference caption. We use the distilled reasoning \textit{DeepSeek-R1-Distill-Qwen-32B} with greedy decoding as judge.
We observe that all evaluated \ac{SOTA} \acp{VLM} have limited capabilities in describing existing \ac{LULC} classes and their spatial composition in detail.

In general, these results demonstrate that \ac{SOTA} \acp{VLM} fall short on tasks requiring multi-sensor spectral information and complex \ac{LULC} understanding.
Exceptions arise when a task closely resembles the model's pre-training objectives, \eg, EarthMind's strong performance on binary presence \ac{VQA}.
This suggests that the poor generalization to other \bentxt{} tasks is primarily a consequence of insufficient pre-training data, rather than an inherent limitation of the model architectures themselves.
Qualitative examples are provided in \suppref{sec:supp_qualitatives}.

\subsection{Results of Fine-tuned InternVL (InternVL$^*$)}

Using the training and validation sets of \bentxt{}, we fine-tune a \ac{SOTA} \ac{VLM} adapted to accommodate multi-sensor input and evaluate it on the \bench{}. To this end, we build on InternVL-3-1B~\cite{zhu2025internvl3} as a backbone and introduce modality-specific branches for \ac{S1} and \ac{S2} inputs, while we retain the original InternVL components. The model is denoted as InternVL$^*$ and its overview is shown in \suppref{sec:supp_rsinternvl}. 
For each sensor modality, a pretrained \ac{ViT} produces patch embeddings, which are aligned to the InternVL \ac{LLM} embedding space via linear projection layers.
The projected \ac{S1} and \ac{S2} tokens are concatenated with the \ac{RGB} tokens and the tokenized instruction before being passed to the \ac{LLM}.
To preserve pretrained representations and reduce computational complexity, all \ac{ViT} backbones are frozen. 
Only the modality-specific projections and LoRA adapters for the \ac{LLM} (rank 8, $\alpha=32$, dropout \num{0.1}) are trained, resulting in \SI{5.8}{\mega\nothing} trainable parameters out of \SI{1.1}{\bel\nothing} in total.
We initialize the \ac{S1} and \ac{S2} encoders with BigEarthNet-pretrained \acp{ViT}~\cite{clasen2025reben}, removing their classification heads. 
We process only the \SI{10}{\meter} and \SI{20}{\meter} bands of \ac{S2}, as the \SI{60}{\meter} bands are primarily used for cloud screening and atmospheric correction~\cite{sumbul2021bigearthnet}.
Training uses batch size 8 with a linear warm-up cosine annealing schedule, increasing the learning rate from $10^{-6}$ to $10^{-4}$ over the first \SI{1}{\percent} of steps, followed by cosine decay.
We fine-tune separately for each task to provide per-task baselines.
The model is fine-tuned on the combined training and validation sets for one epoch and evaluated on the \bentxt{} \bench{}.
Fine-tuning takes approximately two days on four NVIDIA H200 GPUs in total.

\cref{tab:main_ours} shows the results obtained by InternVL$^*$ on the general categories.
From the table, one can see that, once a model is fine-tuned on pairs of \ac{S1} and \ac{S2} data with diverse text annotations, performance on the \bench{} improves substantially across all tasks, even at a small model scale.
Further detailed results obtained by InternVL$^*$ on all 15 tasks are provided in \suppref{sec:supp_rsinternvlresults}.
\input{tables/results/main_table_ours}

%% file: tables/results/main_table.tex
\begin{table}[h!bt]
    \centering
    \small
    \caption{Results of \ac{VLM}s in \ac{RS} and \ac{CV} on the main categories of the \bentxt{} benchmark split. Reported metrics for captioning: BLEU-4, binary \ac{VQA}: accuracy, \ac{MCQ}: accuracy, and \genDet: \ac{mIoU}. All results in percent (\%). *: Reported parameter count.}
    \begin{tabular}{l c *6{S[table-format=2.2]}}
        \toprule{Model}
        & {\# Parameters}
        & {\makecell{Captioning}}
        & {\makecell{Binary\\VQA}}
        & {\makecell{MCQ}}
        & {\makecell{Ref. Exp.\\Detection}}
        \\
        \midrule
        \geochatABBR & 7B & 0.75 & 50.82 & 28.36 & 4.85 \\ 
        \lhrsbotABBR  & 7B & 0.72 & 48.23 & 22.99 & 4.79 \\
        \skyeyegptABBR & 7B & 0.00 & 48.87 & 27.60 & 6.08 \\
        \earthdialrgbABBR & 4B & 0.48 & 58.38 & 32.94 & 7.13 \\ 
        \earthdialmsABBR & 4B & 0.03 & 44.06 & 8.43 & 0.49 \\
        \earthmindrgbABBR & 4B & \best{1.66} & 57.90 & 34.25 & 12.12 \\
        \earthmindsarmsABBR & 4B & \second{1.46} & 57.79 & 35.26 & 16.18 \\
        \midrule
        \gptABBR & 2T* & 0.30 & \second{60.39} & 34.93 & \best{31.73} \\
        \qwenABBR & 8B & 0.70 & \best{61.96} & \best{37.55} & 18.00 \\
        \glmABBR & 10B & 0.79 & 56.59 & 34.45 & \second{27.24} \\
        \llavaABBR & 7B & 0.96 & 58.60 & \second{36.27} & 20.17 \\
        \internvlABBR & 1B & 0.34 & 54.11 & 26.76 & 5.76 \\
        \bottomrule
    \end{tabular}
    \label{tab:main}
\end{table}

%% file: tables/results/binary.tex
\begin{table}[tb]
    \centering
    \caption{Results on the binary VQA tasks for \ac{VLM}s in \ac{RS} and \ac{CV}: Presence, Area, Counting, Adjacency, and Overall, denoted as \presenceABBR, \areaABBR, \countABBR, \adjacencyABBR, and \overallABBR, in terms of accuracy. All results are in percent (\%). \instructionfollowedABBR\ shows whether an unambiguous answer could be extracted all the time.}
    \begin{tabular}{l *5{S[table-format=2.2]} *1{S[table-format=3.2]}}
        \toprule
        {Model}
        & {\presenceABBR}
        & {\areaABBR}
        & {\countABBR}
        & {\adjacencyABBR}
        & {\overallABBR}
        & {\instructionfollowedABBR}
        \\
        \midrule
        \geochatABBR & 51.65 & 46.86 & 43.57 & 55.51 & 50.82 & \gcmark \\
        \lhrsbotABBR & 48.83 & 51.23 & 49.23 & 46.10 & 48.23 & \rxmark \\
        \skyeyegptABBR & 52.19 & 51.61 & 46.63 & 46.94 & 48.87 & \gcmark \\
        \earthdialrgbABBR & 64.47 & 53.06 & 51.99 & \best{60.62} & 58.38 & \rxmark \\
        \earthdialmsABBR & 37.45 & 40.20 & 47.17 & 47.78 & 44.06 & \rxmark \\
        \earthmindrgbABBR & \best{69.34} & 56.20 & 51.07 & 55.97 & 57.90 & \gcmark \\
        \earthmindsarmsABBR & \second{69.07} & 54.59 & 50.15 & 56.98 & 57.79 & \gcmark \\
        \midrule
        \gptABBR & 61.59 & \best{67.38} & \second{61.94} & 55.86 & \second{60.39} & \gcmark \\
        \qwenABBR & 64.33 & \second{66.62} & \best{62.33} & 58.45 & \best{61.96} & \gcmark \\
        \glmABBR & 60.01 & 62.40 & 54.75 & 53.03 & 56.59 & \gcmark \\
        \llavaABBR & 62.07 & 58.19 & 54.36 & \second{58.94} & 58.60 & \gcmark \\ 
        \internvlABBR & 56.24 & 48.77 & 51.61 & 56.60 & 54.11 & \rxmark \\
        \bottomrule
        \end{tabular}
    \label{tab:binary_results}
\end{table}

%% file: tables/results/mcq.tex
\begin{table*}[h!bt]
    \centering
    \caption{Results on the \ac{MCQ} tasks for \ac{VLM}s in \ac{RS} and \ac{CV}: Presence, Area, Counting, Adjacency, Relative Position, Country, Season, Climate Zone, and Overall, denoted as \presenceABBR, \areaABBR, \countABBR, \adjacencyABBR, \relativepositionABBR, \countryABBR, \seasonABBR, \climateABBR, and \overallABBR, respectively in terms of accuracy. All results are in percent (\%). \instructionfollowedABBR\ shows whether an unambiguous answer could be extracted all the time.}
    \begin{tabular}{l *9{S[table-format=2.2]} *1{S[table-format=3.2]}}
    \toprule
    Model
    & {\presenceABBR}
    & {\areaABBR}
    & {\countABBR}
    & {\adjacencyABBR}
    & {\relativepositionABBR}
    & {\countryABBR}
    & {\seasonABBR}
    & {\climateABBR}
    & {\overallABBR}
    & {\instructionfollowedABBR} \\
    \midrule
        \geochatABBR & 32.58 & 26.80 & 24.72 & 29.75 & 25.46 & 34.62 & 22.57 & 30.52 & 28.23 & \rxmark \\ 
        \lhrsbotABBR & 23.84 & 16.54 & 21.19 & 27.23 & 17.48 & 28.53 & 23.71 & 23.56 & 22.99 & \rxmark \\ 
        \skyeyegptABBR & 30.89 & 21.75 & 24.91 & 30.51 & 22.09 & 29.81 & 24.71 & 33.63 & 27.60 & \gcmark \\
        \earthdialrgbABBR & \best{46.40} & 27.41 & \best{36.43} & 30.66 & 29.14 & \best{54.17} & 25.57 & 27.26 & 32.94 & \rxmark \\ 
        \earthdialmsABBR & 21.16 & 5.51 & 4.65 & 8.62 & 0.00 & 15.06 & 6.86 & 7.26 & 8.43 & \rxmark \\ 
        \earthmindrgbABBR & 41.75 & 24.81 & 27.32 & 42.87 & 29.91 & 40.06 & 23.43 & 37.04 & 34.25 & \rxmark \\ 
        \earthmindsarmsABBR & \second{43.58} & 25.57 & 25.28 & \best{44.70} & 31.60 & 40.06 & 25.71 & 36.74 & 35.26 & \rxmark \\ 
        \midrule
        \gptABBR & 38.08 & \best{44.41} & 22.68 & 39.16 & 27.91 & 41.35 & 28.43 & 34.52 & 34.93 & \gcmark \\
        \qwenABBR & 34.41 & \second{37.67} & 29.00 & 42.87 & \second{34.66} & \second{47.76} & \second{27.29} & \best{45.93} & \best{37.55} & \gcmark \\
        \glmABBR & 34.27 & 32.31 & \second{34.01} & 39.44 & 31.90 & 38.78 & 24.71 & 37.93 & 34.45 & \gcmark \\
        \llavaABBR & 34.56 & 32.47 & 32.71 & \second{43.33} & \best{34.82} & 40.38 & \best{27.43} & \second{39.56} & \second{36.27} & \gcmark \\
        \internvlABBR & 28.77 & 26.03 & 23.05 & 30.59 & 23.01 & 26.92 & 24.29 & 26.96 & 26.76 & \rxmark \\
    \bottomrule
    \end{tabular}
    \label{tab:mcq_results}
\end{table*}

%% file: tables/results/bbox_reference.tex
\begin{table}[h!bt]
    \centering
    \caption{Results on \refDet{} for \ac{VLM}s in \ac{RS} and \ac{CV} in terms of \ac{mIoU}, and accuracy@$k$ (where a prediction is counted as correct if the overlap with the reference is greater than $k\%$). All results are in percent (\%). \instructionfollowedABBR\ shows whether an unambiguous answer could be extracted all the time.}
    \begin{tabular}{l *5{S[table-format=2.2]} *1{S[table-format=3.2]}}
        \toprule
        {Model} 
        & {mIoU}
        & \multicolumn{4}{c}{Accuracy}
        & {\instructionfollowedABBR} \\
        \cmidrule(lr){3-6}
        & 
        & {@25} 
        & {@50} 
        & {@75} 
        & {@90}
        & \\
        \midrule
        \geochatABBR   & 7.51 & 11.34 & 1.51 & 0.00 & 0.00 & \gcmark \\ 
        \lhrsbotABBR   & 7.47 & 12.59 & 1.89 & 0.50 & 0.00 & \rxmark \\
        \skyeyegptABBR & 12.12 & 20.03 & 6.55 & 2.27 & 0.13 & \gcmark \\
        \earthdialrgbABBR   & 8.99 & 12.85 & 2.02 & 0.00 & 0.00 & \gcmark \\
        \earthdialmsABBR   & 0.98 & 0.63 & 0.00 & 0.00 & 0.00 & \rxmark \\ 
        \earthmindrgbABBR   & 21.90 & 35.52 & \second{17.88} & \best{5.67} & \best{1.39} & \rxmark \\ 
        \earthmindsarmsABBR & 22.65 & \second{38.41} & 16.62 & \second{4.53} & 1.01 & \rxmark \\ 
        \midrule
        \gptABBR & \best{25.16} & \best{40.81} & \best{19.52} & 4.28 & \second{1.26} & \gcmark \\
        \qwenABBR & 15.60 & 25.31 & 9.95 & 2.39 & 0.38 & \gcmark \\
        \glmABBR & 22.35 & 37.03 & 16.75 & 3.53 & \best{1.39} & \gcmark \\
        \llavaABBR & \second{23.04} & \second{38.41} & 15.74 & 3.90 & 0.63 & \gcmark \\
        \internvlABBR & 5.01 & 5.29 & 0.88 & 0.00 & 0.00 & \rxmark \\
        \bottomrule
    \end{tabular}
    \label{tab:bb_reference_results}
\end{table}

%% file: tables/results/bbox_point.tex
\begin{table}[h!bt]
    \centering
    \caption{Results on \pointDet{} for \ac{VLM}s in \ac{RS} and \ac{CV} in terms of \ac{mIoU}, and accuracy@$k$ (where a prediction is counted as correct if the overlap with the reference is greater than $k\%$). \instructionfollowedABBR\ shows whether an unambiguous answer could be extracted all the time.}
    \begin{tabular}{l *5{S[table-format=2.2]} *1{S[table-format=3.2]}}
        \toprule
        {Model} 
        & {mIoU}
        & \multicolumn{4}{c}{Accuracy}
        & {\instructionfollowedABBR} \\
        \cmidrule(lr){3-6}
        & 
        & {@25} 
        & {@50} 
        & {@75} 
        & {@90}
        & \\
        \midrule
        \geochatABBR & 2.16 & 3.43 & 0.63 & 0.13 & 0.00 & \rxmark \\
        \lhrsbotABBR & 3.09 & 0.25 & 0.00 & 0.00 & 0.00 & \rxmark \\
        \skyeyegptABBR & 0.00 & 0.00 & 0.00 & 0.00 & 0.00 & \rxmark \\
        \earthdialrgbABBR & 5.26 & 3.05 &  0.13 &  0.00 &  0.00 & \gcmark \\
        \earthdialmsABBR & 0.00 & 0.00 & 0.00 & 0.00 & 0.00 & \rxmark \\
        \earthmindrgbABBR & 9.35 & 14.21 & 4.57 & 1.27 & 0.13 & \rxmark \\
        \earthmindsarmsABBR & 12.65 & 17.51 & 5.08 & 0.89 & 0.13 & \rxmark \\
        \midrule
        \gptABBR & \best{38.95} & \best{64.21} & \best{35.53} & \best{8.63} & \best{2.79} & \gcmark \\
        \qwenABBR & 20.42 & 28.81 & 8.25 & 0.89 & 0.00 & \gcmark \\
        \glmABBR & \second{32.16} & \second{56.09} & \second{22.46} & \second{4.82} & \second{1.40} & \gcmark \\
        \llavaABBR & 17.27 & 19.67 & 3.17 & 0.51 & 0.13 & \gcmark \\
        \internvlABBR & 6.52 & 3.05 & 0.00 & 0.00 & 0.00 & \rxmark \\
        \bottomrule
    \end{tabular}
    \label{tab:bb_point_results}
\end{table}

%% file: tables/results/captioning.tex
\begin{table}[h!bt]
    \centering
    \small
    \caption{Results on the captioning task for \ac{VLM}s in \ac{RS} and \ac{CV} in terms of n-gram-based (BLEU-4, ROUGE, METEOR, CIDEr), sentence-embedding-based (BERTScore, SBert-Cosine) and \ac{LLM}-based (CLAIR) metrics, denoted as B4, R, M, C, B, SB, and CL respectively. All results are in percent (\%).}
    \begin{tabular}{l *7{S[table-format=2.2]}}
        \toprule
         Model
         & {B4}
         & {R}
         & {M}
         & {C}
         & {B}
         & {SB}
         & {CL}
         \\
         \midrule
         \geochatABBR & 0.75 & 14.95 & 14.24 & 0.65 & 83.42 & 43.05 & 19.91 \\
         \lhrsbotABBR & 0.72 & 12.44 & 16.93 & 0.01 & 82.52 & \second{54.61} & 28.62 \\
         \skyeyegptABBR & 0.00 & 5.91 & 2.21 & 0.00 & 82.79 & 38.30 & 43.86 \\
         \earthdialrgbABBR & 0.48 & 13.65 & 11.36 & 0.51 & 83.45 & 50.49 & 46.51 \\
         \earthdialmsABBR & 0.03 & 8.64 & 3.51 & 0.00 & 83.34 & 35.00 & \best{59.70} \\
         \earthmindrgbABBR & \best{1.66} & \best{16.54} & 14.93 & \best{0.96} & \best{84.34} & 50.55 & 56.73 \\
         \earthmindsarmsABBR & \second{1.46} & \second{16.43} & 14.62 & 0.76 & \second{84.15} & 51.97 & 56.45 \\
         \midrule
         \gptABBR & 0.30 & 13.39 & 12.69 & 0.63 & 83.58 & 52.52 & 57.68 \\
         \qwenABBR & 0.70 & 11.09 & \best{17.83} & 0.01 & 81.87 & 52.62 & 28.38 \\
         \glmABBR & 0.79 & 14.94 & \second{17.05} & 0.43 & 83.51 & \best{55.90} & 50.91 \\
         \llavaABBR & 0.96 & 14.84 & 15.41 & \second{0.81} & 83.89 & 52.39 & \second{58.91} \\
         \internvlABBR & 0.34 & 11.61 & 16.71 & 0.06 & 81.75 & 51.22 & 35.68 \\
         \bottomrule
    \end{tabular}
    \label{tab:captioning}
\end{table}

%% file: tables/results/main_table_ours.tex
\begin{table}[!tb]
    \centering
    \caption{Results on the main tasks of the \bentxt{} benchmark split for the fine-tuned adapted InternVL$^*$ model as well as best results from \ac{VLM}s in \ac{RS} and \ac{CV}, which are \earthmindrgbABBR{} and \llavaABBR{} for captioning, \earthdialrgbABBR{} and \qwenABBR{} for binary \ac{VQA}, \earthmindsarmsABBR{} and \qwenABBR{} for \ac{MCQ}, and \earthmindsarmsABBR{} and \gptABBR{} for \genDet. 
    Reported metric for captioning is BLEU-4, while that for binary \ac{VQA}, and \ac{MCQ} is accuracy, and that for \genDet{} is mIoU. All results are in percent (\%).}
    \begin{tabular}{l *5{S[table-format=2.2]}}
        \toprule
        {Model}
        & {\makecell{Captioning}}
        & {\makecell{Binary VQA}}
        & {\makecell{MCQ}} 
        & {\makecell{Ref. Exp.\\Detection}}
        \\
        \midrule
        \ac{SOTA} \ac{RS} & 1.66 & 58.38 & 35.26 & 16.18 \\
        \ac{SOTA} \ac{CV} & 0.96 & 61.96 & 37.55 & 31.73  \\
        \midrule
        \textbf{InternVL$^*$} & \best{34.04} & \best{73.29} & \best{65.84} & \best{52.41}  \\
        \bottomrule
    \end{tabular}
    \label{tab:main_ours}
\end{table}

%% file: sec/6_conclusion.tex
In this work, we present \bentxt{}, a large-scale multi-sensor \ac{RS} image-text dataset that encompasses 15 downstream tasks across four distinct categories: image captioning, binary \ac{VQA}, \ac{MCQ}, and \genDet.
Our dataset addresses the critical scarcity of large-scale \ac{RS} image-text resources spanning diverse textual annotations associated with co-registered multi-sensor data.
With this dataset, we also provide a \bench{}, which contains manually verified image-text samples, enabling reliable evaluation of \ac{VLM} capabilities to model the complex spatial and spectral characteristics of multi-sensor \ac{RS} data.
Through systematic evaluation on the \bench{}, we show that both general-purpose \ac{CV} and \ac{RS}-specialized \acp{VLM} are subject to limited capabilities. In particular, \ac{RS}-specific models lead to almost no advantage over their general-purpose counterparts.
To demonstrate that these limitations stem from data scarcity rather than architectural constraints, we adapt InternVL-3-1B~\cite{zhu2025internvl3} for multi-sensor data and fine-tune on \bentxt{}.
The adapted model achieves an improvement of \SI{11.33}{\percent} to \SI{32.38}{\percent} over \ac{SOTA}, demonstrating that even a small-scale \ac{VLM} with simple adaptations can effectively model the complex content of multi-sensor data when an adequate training dataset is considered. 
In this context, the proposed dataset is very promising for effective and interactive \ac{EO}, enabling natural language interactions with the complex content of \ac{RS} images, thus democratizing the use of EO data for non-experts as well. 

%% file: sec/x_supplementary.tex
\onecolumn
\section*{Overview of Supplementary Material}
\renewcommand{\theenumi}{\Alph{enumi}}

\begin{enumerate}[leftmargin=3em]
    \setcounter{enumi}{0}
    \item [\clickablelabel{sec:supp_discussion}] Discussions and Broader Impact
    \item [\clickablelabel{sec:supp_prompts}]  Prompts for Caption Generation Pipeline
    \item [\clickablelabel{sec:supp_templates}] Templates for \ac{VQA} and \genDet{} Generation
    \item [\clickablelabel{sec:supp_testresults}] Additional Results on Test Split
    \item [\clickablelabel{sec:supp_evalprompts}] Full Instructions for Model Evaluation on \bentxt
    \item [\clickablelabel{sec:supp_qualitatives}] Qualitative Examples of Model Answers
    \item [\clickablelabel{sec:supp_rsinternvl}] InternVL$^*$ Architecture Overview
    \item [\clickablelabel{sec:supp_rsinternvlresults}] InternVL$^*$ Results for all Tasks
    \item [\clickablelabel{sec:supp_rsinternvlablations}] InternVL$^*$ Ablations
\end{enumerate}
\FloatBarrier

\section{Discussions and Broader Impact}\label{sec:supp_discussion}

In \cref{sec:supp_impact} we discuss the broader impact of our work and in \cref{sec:supp_future} limitations and future work.

\FloatBarrier
\subsection{Broader Impact}\label{sec:supp_impact}

\bentxt{} supports the development and testing of \acp{VLM} for \ac{RS} images with complex content. 
By providing co-registered multi-sensor data with diverse text annotations describing complex \ac{LULC} patterns \bentxt{} can serve as a valuable fine-tuning resource, as well as part of pre-training a large scale foundation \ac{VLM} in conjunction with other datasets.
The manually-checked \bentxt{} \bench{} on the other hand enables the faithful evaluation of pre-trained \acp{VLM}.
Furthermore, through \bentxt{} natural language interaction with \ac{RS} image archives can be made possible, reducing the barrier for non-experts.
This allows the broader public to assess and analyze \ac{RS} archives and therefore accelerate and democratize several \ac{EO} applications.
Finally, as we publicly released the complete dataset generation pipeline the overall process can easily be adapted and applied to different domains even beyond \ac{EO}.
The generation of almost \num{500000} \ac{LLM}-augmented and self-refined captions as well as the fine-tuning of InternVL$^*$ and model inference for \ac{SOTA} analysis incurs non-trivial computational cost and energy consumption. 
As the dataset, creation pipeline, and fine-tuned InternVL$^*$ checkpoints are publicly released this is a one-time cost.

\FloatBarrier
\subsection{Limitations and Future Work}\label{sec:supp_future}

While \bentxt{} is a valuable resource for advancements in image-text learning for \ac{RS}, it has some limitations.
\bentxt{} is limited in geographical coverage to ten European countries.
Consequently, \ac{LULC} classes that are underrepresented or absent in Europe are not covered.
At the moment the co-registered multi-sensor data is limited to two sensors (\ac{S1} and \ac{S2}).
In addition, we employ \acp{LLM} for rephrasing and CLAIR~\cite{chan-etal-2023-clair} evaluation for image captioning that could cause hallucinations or nonfactual responses.
Our \bench{} counteracts this as it is manually-verified and completely correct.
For image captioning, we also report performance under several standard captioning metrics.
Finally, while the text annotations of \bentxt{} are already diverse in semantic content and type, there exists an inherent limit due to the generation from \ac{CLC} maps and the 19-class nomenclature.
As future work we plan to increase geographical coverage of \bentxt{}, consider additional \ac{LULC} sources, add additional sensing modalities, and increase annotation type variety even further.

\FloatBarrier
\section{Prompts for Caption Generation Pipeline}\label{sec:supp_prompts}
\Cref{fig:prompt_rephrase} shows the full prompt used for \ac{LLM}-based paraphrasing of template-based captions.
The prompt comprises a general task-description, set of rules to adhere to, in-context examples (shortened for readability), and the actual template-based caption input.
Finally we provide the rephrased output from the model.

\input{figures/supplementary/prompt_caption_paraphrase}

In \cref{fig:prompt_refinement} we show the prompt for \ac{LLM}-based caption refinement.
This includes the paraphrase prompt and output from \cref{fig:prompt_rephrase}, followed by a refinement instruction asking the model to improve the paraphrased caption based on the initial template-based version and rephrasing instruction.
\input{figures/supplementary/prompt_caption_refinement}

\FloatBarrier
\section{Instruction Templates}\label{sec:supp_templates}
We present examples of the instruction templates used for each task from; i) binary \ac{VQA} in \cref{tab:templates_vqa}, ii) multiple-choice \ac{VQA} in \cref{tab:templates_mcq}, iii) \genDet{} in \cref{tab:templates_bb}, and iv) captioning in \cref{tab:templates_cap}.
Overall 469 templates are used for \bentxt.
\input{tables/supplementary/templates_vqa}
\input{tables/supplementary/templates_mcq}
\input{tables/supplementary/templates_bb}
\input{tables/supplementary/templates_cap}

\FloatBarrier
\section{Additional Results on Test Set}\label{sec:supp_testresults}
Additional results for image captioning, binary \ac{VQA}, \ac{MCQ}, and \genDet{} on the full \bentxt{} test set are detailed in \cref{tab:main_test}.
Due to computational constraints, we evaluate the strongest models from \ac{RS} (EarthMind) and \ac{CV} (Qwen3-VL), the backbone model for InternVL$^*$ (InternVL-3 (1B)), and the fine-tuned InternVL$^*$.
\input{tables/supplementary/test_split_results}
From the table, one can observe that performance on the full test set is mostly similar (within a few \%) compared to performance on the \bentxt{} \bench.
The deviation in performance is mainly due to the sampling strategy employed for the \bench{} to allow unbiased and fair evaluation.
Accordingly, InternVL$^*$, fine-tuned on the train set shows the largest performance difference.
For fair, unbiased, and manually-checked evaluation, we recommend evaluating on the \bench{} that also drastically reduces computational cost.

\FloatBarrier
\section{Full Instructions for Model Evaluation on \bentxt}\label{sec:supp_evalprompts}
\Cref{tab:inferenceprompts} lists the precise prompts per category used for model evaluation.
\input{tables/supplementary/inference_prompts}
In general, we only add additional prompts when required.
This includes model-specific special tokens (\eg \texttt{[refer]} for GeoChat), or instructions to follow the expected output format (\eg \texttt{"Answer with a single word:"yes" or "no"."} when the model alone is unable to reliably do so.
Additionally, for area \acp{VQA} and for the captioning task we add \texttt{"The whole image area is 1440000 square meters."} before the instruction, to enable the relation between pixels and areas for each model.

\FloatBarrier
\section{Qualitative Examples of Model Answers}\label{sec:supp_qualitatives}
\Cref{fig:supp_qualitatives1,fig:supp_qualitatives2,fig:supp_qualitatives3} show qualitative per model answers for the binary \ac{VQA}, \ac{MCQ}, and \genDet{} tasks.
\Cref{fig:qual_caption_skeleton} gives an example for the captioning task with \ac{RS} model answers and \cref{fig:qual_caption_skeleton_cv} shows the answers of \ac{CV} models.
One can clearly see that given adequate pre-training data a model is capable of answering correctly.
Especially in \cref{fig:qual_caption_skeleton,fig:qual_caption_skeleton_cv} it is evident that multi-sensor and diverse pre-training data leads to a more specific description of complex \ac{LULC} classes (\eg "mixed forest" instead of just "forest").
The considered \ac{RS} models are GeoChat~\cite{kuckreja2024geochat} (denoted as \geochatABBRns), LHRS-Bot~\cite{muhtar2024lhrs} (denoted as \lhrsbotABBRns), SkyEyeGPT~\cite{zhan2025skyeyegpt} (denoted as \skyeyegptABBRns), EarthDial~\cite{soni2025earthdial} (denoted as \earthdialrgbABBRns/ \earthdialmsABBRns), and EarthMind~\cite{shu2025earthmind} (denoted as \earthmindrgbABBRns/ \earthmindsarmsABBRns).
The considered \ac{CV} models are GPT-5.2~\cite{openai2025gpt5} (denoted as \gptABBRns), Qwen3-VL~\cite{bai2025qwen3} (denoted as \qwenABBRns), GLM-4.6v-Flash~\cite{hong2025glm} (denoted as \glmABBRns), LLaVa-OneVision~\cite{li2024llavaonevision} (denoted as \llavaABBRns), and InternVL3-1B~\cite{zhu2025internvl3} (denoted as \internvlABBRns).
Further we provide model responses for the fine-tuned InternVL$^*$ (denoted as \internvlrsABBR).

\input{figures/supplementary/qualitatives}
\input{figures/supplementary/examples_caption}
\input{figures/supplementary/examples_captions_cv}

\FloatBarrier
\section{InternVL$^*$ Architecture Overview}\label{sec:supp_rsinternvl}
The detailed architecture of InternVL$^*$ is provided in \cref{fig:model_architecture}.
We retain the original InternVL components (green), while adding additional paths for \ac{S1} and \ac{S2} (orange).
To leverage InternVL's pre-trained capabilities we provide the \ac{RGB} bands of \ac{S2} to the InternViT.
Only the projection layers and LoRA adapters for InternLM are fine-tuned.
\input{figures/model_v1}

\FloatBarrier
\section{InternVL$^*$ Results for all Tasks}\label{sec:supp_rsinternvlresults}
\Cref{tab:binary_results_ours,tab:mcq_results_ours,tab:bb_point_results_ours,tab:bb_reference_results_ours,tab:captioning_ours} show results on the \bench{} for the fine-tuned InternVL$^*$ on all 15 tasks across the four categories binary \ac{VQA} (\cref{tab:binary_results_ours}), \ac{MCQ} (\cref{tab:mcq_results_ours}), \genDet{} (\cref{tab:bb_point_results_ours,tab:bb_reference_results_ours}), and captioning (\cref{tab:captioning_ours}).
We include the top result obtained by an \ac{RS} model as well as a \ac{CV} model for each task.
One can clearly see that performance is increased drastically when adequate pre-training data is used.
Only for the \ac{MCQ} area task GPT performs slightly better, due to its much larger parameter count.
In general, on \ac{MCQ} GPT performs less favorably.

\input{tables/results/binary_ours}
\input{tables/results/mcq_ours}
\input{tables/results/bbox_reference_ours}
\input{tables/results/bbox_point_ours}
\input{tables/results/captioning_ours}
\FloatBarrier

\section{InternVL$^*$ Ablations}
\label{sec:supp_rsinternvlablations}

To evaluate the generalizability of models trained on \bentxt{}, we re-train InternVL$^*$ with the two largest country splits Serbia and Finland ($\sim47\%$ of train and validation samples) held out and evaluate on them. 
\Cref{tab:supp_countryabl} shows the results.
When location questions are not evaluated, the accuracy for MCQ reaches 51.58\%.
This is a reasonable strategy to assess the generalizability, since \bentxt{} already covers a large geographical area of 10 countries associated with 19 LULC classes and 10 of 31 global climate zones (see \cite{clasen2025reben} for a detailed discussion of geographic coverage).
Additionally, the generalizability of BigEarthNet itself is widely studied, and existing domain adaptation approaches are also applicable for models trained on \bentxt.
\input{tables/results/country_ablation}
To evaluate the role of multispectral over RGB data we ablate InternVL$^*$ by fine-tuning only on the RGB bands of the S2 imagery of \bentxt{}.
\Cref{tab:supp_rgbabl} shows the results.
As already shown in ~\cite{clasen2025reben}, jointly using co-registered S1 and S2 images of BigEarthNet improves performance.
Similarly, we observe a performance drop for all annotation categories when using only the RGB bands.

\input{tables/results/rgb_ablation}

%% file: figures/supplementary/prompt_caption_paraphrase.tex
\begin{tcolorbox}[colback=lightGreen,
colframe=darkGreen,
coltitle=white,
title=Prompt for the LLM-based Augmentation - Paraphrase,
fonttitle=\bfseries,
coltext=black,
boxrule=1pt,
width=\textwidth,
enhanced,
breakable,
boxsep=1mm,
overlay={
        \node[anchor=north east, xshift=-3.5mm, yshift=-145mm] at (interior.north east)
        {\includegraphics[width=0.05\textwidth]{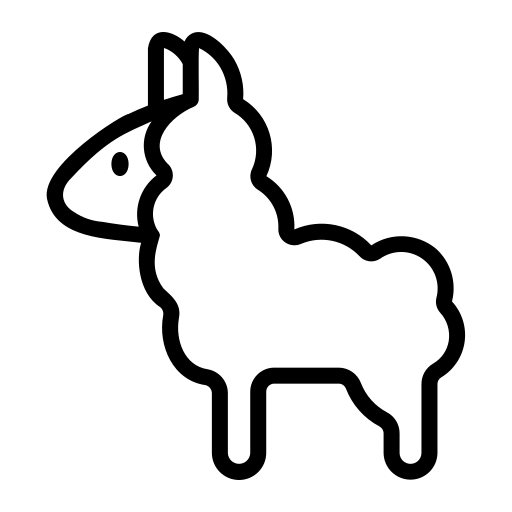}};
    }
]

\scriptsize

\noindent
\begin{minipage}[t]{0.14\textwidth}
\raggedright\bfseries
Task Description
\end{minipage}\hfill
\begin{minipage}[t]{0.83\textwidth}
You are a remote-sensing expert. You will be given a static description of a satellite image with well-defined classes. The well-defined classes are enclosed in quotation marks, followed by their area in square meters in brackets. The description also includes information about which classes are bordering or adjacent to one another, ordered by their relative size without repetition. Use varying natural language and sentence structure to rephrase the description into a single, cohesive, well-written remote sensing caption that contains all of the provided information.

\smallskip

Use the following hierarchy definition to further increase the variance in your output:

\begin{itemize}
\item \enquote{Urban fabric} and \enquote{Industrial or commercial units} are part of the more general class \enquote{Artificial surfaces}.
\item \enquote{Arable land}, \enquote{Permanent crops}, \enquote{Pastures}, \enquote{Complex cultivation patterns}, \enquote{Land principally occupied by agriculture with significant areas of natural vegetation}, and \enquote{Agro-forestry areas} are part of the more general class \enquote{Agricultural areas}.
\item \enquote{Broad-leaved forest}, \enquote{Coniferous forest}, \enquote{Mixed forest}, \enquote{Natural grasslands or sparsely vegetated areas}, \enquote{Moors, heathland or sclerophyllous vegetation}, \enquote{Transitional woodlands or shrubs}, and \enquote{Beaches, dunes, or sands} are part of the more general class \enquote{Forest and seminatural areas}.
\item \enquote{Coastal wetlands} and \enquote{Inland wetlands} are part of the more general class \enquote{Wetlands}.
\item \enquote{Inland waters} and \enquote{Marine waters} are part of the more general class \enquote{Water bodies}.
\end{itemize}    

\end{minipage}

\smallskip
\hrule
\smallskip

\noindent
\begin{minipage}[t]{0.14\textwidth}
\raggedright\bfseries
Set of Rules
\end{minipage}\hfill
\begin{minipage}[t]{0.83\textwidth}

While writing the caption, do not violate the following rules:
\begin{itemize}
    \item Use precise language to describe observable features.
    \item Avoid assumptions or interpretations beyond the given data.
    \item Do not make assumptions towards the land use and avoid the phrases \enquote{land-cover} and \enquote{land-use}.
    \item Include the associated area information if you do not change the class names.
    \item If you group classes, ensure to either remove the associated areas or sum them together and re-evaluate the border information.
    \item Either choose \SI{5}{\percent} or 72'000 sqm when talking about the smaller regions, but do not use both.
    \item Remove the quotation marks for the predefined classes in your output.
    \item Do not prefix your output with "Output:".
    \item If the input mentions the associated climate zone, keep the climate zone definition as is.
    \item Include the seasonal and country information in the output.
\end{itemize}

\end{minipage}

\smallskip
\hrule
\smallskip

\noindent
\begin{minipage}[t]{0.14\textwidth}
\raggedright\bfseries
In-context Examples
\end{minipage}\hfill
\begin{minipage}[t]{0.83\textwidth}
The following provides a few examples.

\smallskip

Input: The image primarily shows \enquote{Arable land} ($\sim$708'000 sqm), followed by \enquote{Broad-leaved forest} ($\sim$413'000 sqm). Smaller areas are covered by \enquote{Pastures} ($\sim$319'000 sqm) and \enquote{Mixed forest} ($\sim$204'000 sqm). \enquote{Arable land} borders \enquote{Broad-leaved forest} and \enquote{Pastures}. Moreover, \enquote{Broad-leaved forest} is adjacent to \enquote{Mixed forest}. The \enquote{Arable land} is distributed over three individual areas, where one is a larger area ($\sim$525'000 sqm) and two are smaller areas (110'000 sqm and 73'000 sqm). The \enquote{Mixed forest} is distributed over two individual smaller areas (116'000 sqm and 88'000 sqm). The image was taken during Summer in Austria. The associated climate zone is \enquote{cold, no dry season, warm summer}.

\smallskip

Output: This satellite image, captured in Austria during summer, depicts a diverse landscape dominated by agricultural and forested areas. Arable land (708,000 sqm) is most prominent, covering a large area and two smaller areas, and is next to broad-leaved forest ($\sim$413,000 sqm) and pastures ($\sim$319,000 sqm). The expansive broad-leaved forest is adjacent to a mixed forest ($\sim$204,000 sqm), which is spread across two smaller areas, suggesting a varied forest ecosystem. The landscape is characterized by a mix of agricultural, forest and seminatural areas, set within the \enquote{cold, no dry season, warm summer} climate zone.

\smallskip

Input: ...

\smallskip

Output: ...

\smallskip

Input: ...

\smallskip

Output ...
\end{minipage}

\smallskip
\hrule
\smallskip

\noindent
\begin{minipage}[t]{0.14\textwidth}
\raggedright\bfseries
Input Template Caption
\end{minipage}\hfill
\begin{minipage}[t]{0.83\textwidth}
The image primarily shows \enquote{Arable land} ($\sim$526'000 sqm), followed by \enquote{Inland wetlands} ($\sim$460'000 sqm). Smaller areas are covered by \enquote{Inland waters} ($\sim$305'000 sqm) and \enquote{Urban fabric} ($\sim$149'000 sqm). \enquote{Arable land} borders \enquote{Inland wetlands} and \enquote{Urban fabric}. Moreover, \enquote{Inland wetlands} is adjacent to \enquote{Inland waters} and \enquote{Urban fabric}. The \enquote{Urban fabric} is distributed over three individual marginal areas. The image was taken during Summer in Switzerland. The associated climate zone is \enquote{temperate, no dry season, warm summer}.
\end{minipage}

\smallskip
\hrule
\smallskip

\begin{minipage}[t]{0.93\textwidth}
This satellite image, captured during the summer season in Switzerland, showcases a diverse landscape within the \enquote{temperate, no dry season, warm summer} climate zone. The dominant features are arable land ($\sim$526,000 sqm) and inland wetlands ($\sim$460,000 sqm), which are adjacent to each other. The arable land also borders urban fabric ($\sim$149,000 sqm), which is scattered across three small, marginal areas. Furthermore, the inland wetlands are next to inland waters ($\sim$305,000 sqm) and urban fabric. The varied landscape presents a mix of agricultural areas, wetlands, water bodies, and artificial surfaces.
\end{minipage}
\end{tcolorbox}
\noindent
\begin{minipage}{\textwidth}
\captionof{figure}{Promp for \ac{LLM}-based caption rephrasing based on template-based version, including general task description, set of rules for the model to follow, three in-context examples (shortened for brevity), actual template-based caption input, and model output. As model we employ LLama-4-Scout-17B\chdeleted{ \cite{touvron2023llamaopenefficientfoundation}}.}
\label{fig:prompt_rephrase}
\end{minipage}

%% file: figures/supplementary/prompt_caption_refinement.tex
\begin{tcolorbox}[
    colback=lightGreen,
    colframe=darkGreen,
    coltitle=white,
    title=LLM-based Augmentation - Refinement,
    fonttitle=\bfseries,
    coltext=black,
    boxrule=1pt,
    width=\textwidth,
    enhanced,
    breakable,
    boxsep=1mm,
    overlay={
        \node[anchor=north east, xshift=-3.5mm, yshift=-20mm] at (interior.north east)
        {\includegraphics[width=0.05\textwidth]{figures/supplementary/llama.png}};
        \node[anchor=north east, xshift=-3.5mm, yshift=-43mm] at (interior.north east)
        {\includegraphics[width=0.05\textwidth]{figures/supplementary/llama.png}};
    }
]

\scriptsize

\noindent
\begin{minipage}[t]{0.14\textwidth}
\raggedright\bfseries
Paraphrase Prompt
\end{minipage}\hfill
\begin{minipage}[t]{0.83\textwidth}
You are a remote-sensing expert. You will be given a static description of a satellite image with well-defined classes. The well-defined classes are enclosed in quotation marks, followed by their area in square meters in brackets. The description also includes information about which classes are bordering or adjacent to one another, ordered by their relative size without repetition. Use varying natural language and sentence structure to rephrase the description into a single, cohesive, well-written remote sensing caption that contains all of the provided information.

...
\end{minipage}

\smallskip
\hrule
\smallskip

\noindent
\begin{minipage}[t]{0.93\textwidth}
This satellite image, captured during the summer season in Switzerland, showcases a diverse landscape within the \enquote{temperate, no dry season, warm summer} climate zone. The dominant features are arable land ($\sim$526,000 sqm) and inland wetlands ($\sim$460,000 sqm), which are adjacent to each other. The arable land also borders urban fabric ($\sim$149,000 sqm), which is scattered across three small, marginal areas. Furthermore, the inland wetlands are next to inland waters ($\sim$305,000 sqm) and urban fabric. The varied landscape presents a mix of agricultural areas, wetlands, water bodies, and artificial surfaces.
\end{minipage}

\smallskip
\hrule
\smallskip

\noindent
\begin{minipage}[t]{0.14\textwidth}
\raggedright\bfseries
Refinement Instruction
\end{minipage}\hfill
\begin{minipage}[t]{0.83\textwidth}
Compare your rephrased caption with the initially provided static description.
Ensure that your output fulfills the previously defined rules and does not have less information than the original description.
Update the caption accordingly and provide the full caption in your output.
Do NOT explain what mistake you have found or what you are correcting. Only include the updated caption in your response.
\end{minipage}

\smallskip
\hrule
\smallskip

\noindent
\begin{minipage}[t]{0.93\textwidth}
This satellite image, captured during the summer season in Switzerland, showcases a diverse landscape within the \enquote{temperate, no dry season, warm summer} climate zone. The dominant features are arable land ($\sim$526,000 sqm) and inland wetlands ($\sim$460,000 sqm), which are adjacent to each other. The arable land borders both inland wetlands and urban fabric ($\sim$149,000 sqm). Moreover, the inland wetlands are adjacent to both inland waters ($\sim$305,000 sqm) and urban fabric. Notably, the urban fabric is distributed over three individual marginal areas. The varied landscape presents a mix of agricultural areas, wetlands, water bodies, and artificial surfaces.
\end{minipage}
\end{tcolorbox}
\noindent
\begin{minipage}{\textwidth}
\captionof{figure}{Prompt used for \ac{LLM}-based caption self-refinement, including paraphrasing prompt with output from \cref{fig:prompt_rephrase}, the refinement instruction and the refined caption output from the model.}
\label{fig:prompt_refinement}
\end{minipage}

%% file: tables/supplementary/templates_vqa.tex
\begin{table}[h!t]
\centering
\caption{Examples of question templates for each task of the binary \ac{VQA} category. \{...\} denotes a placeholder for the actual value depending on the image content.}
\scriptsize
\begin{tabular}{p{0.11\textwidth}p{0.86\textwidth}}
\toprule
\textbf{Task} & \textbf{Templates} \\
\midrule
Presence & 
Is/Are any \{class\} present in the satellite image?\newline
Is/Are any \{class\} shown in the image?\newline
Do you see any area(s) of \{class\} in the satellite image?\newline
Are there regions of \{class\} in the satellite image?\newline
Is any portion of the image covered by \{class\}?\\
\midrule
Area &
Is the image fully covered by \{class\}?\newline
Do \{class\} cover \{bound\} of the image?\newline
Are \{class\} occupying \{lower bound\} or more of the image?\newline
Is the total area of \{class\} less than \{upper bound\}?\newline
Do \{class\} cover between \{lower bound\} and \{lower bound\} of the image?\\
\midrule
Count & 
Does the image contain exactly one continuous region of \{class\}?\newline
Are \{class\} present in multiple distinct continuous areas within the image?\newline
Does the image show exactly two distinct continuous areas of \{class\}?\newline
Do \{class\} occupy two or less continuous areas in the image?\newline
Would you say that \{class\} occupy four or more continuous areas in the image?\\
\midrule
Adjacency &
Does any (instance of) \{class\} border a/\_ \{class\} in this scene?\newline
Is there any (instance of) \{class\} and a/\_ \{class\} directly connected in the scene?\newline
Is any (instance of) \{class\} directly adjacent to a/\_ \{class\} in this image?\newline
Is there direct adjacency between any (instance of) \{class\} and a/\_ \{class\} in the scene?\newline
Does any (instance of) \{class\} directly border a/\_ \{class\} in this scene?\\
\bottomrule
\end{tabular}\label{tab:templates_vqa}
\end{table}

%% file: tables/supplementary/templates_mcq.tex
\begin{table}[h!t]
\centering
\caption{Examples of question templates for each task of the \ac{MCQ} category. \{...\} denotes a placeholder for the actual value depending on the image content.}
\scriptsize
\begin{tabular}{p{0.11\textwidth}p{0.86\textwidth}}
\toprule
\textbf{Task} & \textbf{Templates} \\
\midrule
Presence & 
Which of these classes is present in the image? a) \{class\}, b) \{class\}, c) \{class\}, d) \{class\}\newline
Which land cover class appears in this satellite image? a) \{class\}, b) \{class\}, c) \{class\}, d) \{class\}\newline
From these options, which class can be observed in the image? a) \{class\}, b) \{class\}, c) \{class\}, d) \{class\}\\
\midrule
Area &
How much of the image do \{class\} cover? a) \{area range\}, b) \{area range\}, c) \{area range\}, d) \{area range\}\newline
What is the total area of \{class\}? a) \{area range\}, b) \{area range\}, c) \{area range\}, d) \{area range\}\newline
Which range corresponds to the area covered by \{class\}? a) \{area range\}, b) \{area range\}, c) \{area range\}, d) \{area range\}\newline
How large is the area covered by \{class\}? a) \{area range\}, b) \{area range\}, c) \{area range\}, d) \{area range\}\\
\midrule
Count & 
How many continuous areas are covered by \{class\}? a) \{number\}, b) \{number\}, c) \{number\}, d) \{number\}\newline
Count the separate regions of \{class\}: a) \{number\}, b) \{number\}, c) \{number\}, d) \{number\}\newline
How many separate areas of \{class\} can be seen? a) \{number\}, b) \{number\}, c) \{number\}, d) \{number\}\\
\midrule
Adjacency &
Identify the adjacent classes: a) \{class pair\}, b) \{class pair\}, c) \{class pair\}, d) \{class pair\}\newline
Which classes are directly connected? a) \{class pair\}, b) \{class pair\}, c) \{class pair\}, d) \{class pair\}\newline
Which pair of classes share a boundary? a) \{class pair\}, b) \{class pair\}, c) \{class pair\}, d) \{class pair\}\\
\midrule
Relative Position &
How is/are the \{class\} relatively positioned to the \{class\} in the image? a) \{rel. pos.\}, b) \{rel. pos.\}, c) \{rel. pos.\}, d) \{rel. pos.\}\newline
In relation to the \{class\}, where is/are the \{class\} located in the image? a) \{rel. pos.\}, b) \{rel. pos.\}, c) \{rel. pos.\}, d) \{rel. pos.\}\newline
What is the relative position of the \{class\} to the \{class\}? a) \{rel. pos.\}, b) \{rel. pos.\}, c) \{rel. pos.\}, d) \{rel. pos.\}\\
\midrule
Country &
Which country is shown in the satellite image? a) \{country\}, b) \{country\}, c) \{country\}, d) \{country\}\newline
From the options below, select the country shown in the image: a) \{country\}, b) \{country\}, c) \{country\}, d) \{country\}\newline
Choose the country depicted in the image: a) \{country\}, b) \{country\}, c) \{country\}, d) \{country\}\\
\midrule
Season &
In which of the following seasons was the satellite image taken? a) \{season\}, b) \{season\}, c) \{season\}, d) \{season\}\newline
Select the season shown in this satellite image: a) \{season\}, b) \{season\}, c) \{season\}, d) \{season\}\newline
In which season was the satellite image captured? a) \{season\}, b) \{season\}, c) \{season\}, d) \{season\}\\
\midrule
Climate Zone &
From the following options, pick the one that corresponds to the climate zone shown in the satellite image: a) \{climate zone\}, b) \{climate zone\}, c) \{climate zone\}, d) \{climate zone\}\newline
Identify which of the following climate zones is shown in the satellite image: a) \{climate zone\}, b) \{climate zone\}, c) \{climate zone\}, d) \{climate zone\}\newline
Which climate zone does the satellite image capture? a) \{climate zone\}, b) \{climate zone\}, c) \{climate zone\}, d) \{climate zone\}\\
\bottomrule
\end{tabular}\label{tab:templates_mcq}
\end{table}

%% file: tables/supplementary/templates_bb.tex
\begin{table}[h!t]
\centering
\caption{Examples of question templates for \refDet{} and \pointDet{}. \{...\} denotes a placeholder for the actual value depending on the image content.}
\scriptsize
\begin{tabular}{p{0.11\textwidth}p{0.86\textwidth}}
\toprule
\textbf{Task} & \textbf{Templates} \\
\midrule
Referring\newline
LULC\newline
Detection& 
Where can the smallest/largest continuous region of \{class\} be found?\newline
Identify the location of the smallest/largest contiguous area of \{class\}.\newline
Where is the smallest/largest continuous area of \{class\} in the image?\newline
Locate a continuous area of \{class\}.\newline
Indicate the position of a continuous region of \{class\}.\\
\midrule
Referring\newline
Point\newline
Detection&
Output a bounding box surrounding the land cover class instance positioned at \{(x, y)\} in the image.\newline
Generate a bounding box enclosing the land cover class instance at \{(x, y)\} in the satellite image.\newline
Create a bounding box around the land cover class instance found at \{(x, y)\} in the satellite image.\newline
Provide a bounding box enclosing the land cover class instance located at \{(x, y)\} in the image.\newline
Output a bounding box surrounding the land cover class instance at \{(x, y)\} in the satellite image.\\
\bottomrule
\end{tabular}\label{tab:templates_bb}
\end{table}

%% file: tables/supplementary/templates_cap.tex
\begin{table}[h!t]
\centering
\caption{Examples of question templates for captioning.}
\scriptsize
\begin{tabular}{p{0.11\textwidth}p{0.86\textwidth}}
\toprule
\textbf{Task} & \textbf{Templates} \\
\midrule
Captioning & 
Describe this satellite image, noting the region, seasonal context, and distribution of land cover types.\newline
Provide a comprehensive description of the scene, including the location, seasonal context and land cover distribution.\newline
Describe the observed landscape, specifying the geographic location, season, and land cover composition.\newline
Explain what this image shows, including the region, climate zone, and land cover characteristics.\newline
Provide an description of the image content, highlighting the geographic location, season, and spatial patterns.\\
\bottomrule
\end{tabular}\label{tab:templates_cap}
\end{table}

%% file: tables/supplementary/test_split_results.tex
\begin{table}[h!t]
    \centering
    \caption{Inference performance of \ac{RS} and \ac{CV} \ac{VLM}s on the main tasks on the \bentxt{} test split. Reported metrics for captioning: BLEU-4, binary \ac{VQA}: accuracy, \ac{MCQ}: accuracy, and \genDet: mIoU. All results are in percent (\%).}
    \begin{tabular}{l c *5{S[table-format=2.2]}}
        \toprule
        {Model}
        & {\# Parameters}
        & {\makecell{Captioning}}
        & {\makecell{Binary VQA}}
        & {\makecell{MCQ}}
        & {\makecell{Ref. Exp. Det.}}
        \\
        \midrule
        EarthMind\textsubscript{s1s2}~\cite{shu2025earthmind} & 4B & 1.59 & 55.36 & 34.73 & 16.25 \\
        \midrule
        Qwen3-VL~\cite{bai2025qwen3} & 8B & 0.78 & 61.27 & 37.06 & 18.60 \\
        InternVL-3 (1B)~\cite{zhu2025internvl3} & 1B & 0.52 & 51.44 & 27.56 & 5.94 \\
        \midrule
        InternVL$^*$ & 1B & \bfseries 31.06 & \bfseries 80.06 & \bfseries 74.42 & \bfseries {51.35} \\
        \bottomrule
    \end{tabular}
    \label{tab:main_test}
\end{table}

%% file: tables/supplementary/inference_prompts.tex
\begin{table*}[h!t]
    \notsotiny
    \centering
    \caption{Per model per category prompt used during inference. \{Instruction\} denotes the instruction text directly from \bentxt.}
    \label{tab:prompts}
    \setlength{\tabcolsep}{2.5pt}
    \begin{tabular}{c *4{p{0.2\textwidth}}}
    \toprule
    {\thead{Model}} & \thead{Binary \ac{VQA}} & \thead{\ac{MCQ}} & \thead{Ref. Exp. Det.} & \thead{Captioning} \\
    \midrule
    GeoChat & \{Instruction\} Answer with a single word: "yes" or "no". & \{Instruction\} Answer with exactly one of a, b, c, d. & [refer] \{Instruction\} & \{Instruction\} \\
    \midrule
    LHRS-Bot & \{Instruction\} Answer with a single word: "yes" or "no". & \{Instruction\} Answer with exactly one of a, b, c, d. & \{Instruction\} Output ONLY the bounding box as [x1, y1, x2, y2]. Coordinates must be in the 224x224 image space. Do not include any other text. & \{Instruction\} \\
    \midrule
    SkyEyeGPT & [vqa] \{Instruction\} Answer with a single word: "yes" or "no". & [vqa] \{Instruction\} Answer with exactly one of a, b, c, d. & [refer] \{Instruction\} & [caption] \{Instruction\} \\
    \midrule
    ED\textsubscript{RGB} & [s2\_rgb\_10] \{Instruction\} Answer with a single word: "yes" or "no". & [s2\_rgb\_10] \{Instruction\} & [refer] [s2\_rgb\_10] \{Instruction\} & [caption] [s2\_rgb\_10] \{Instruction\} \\
    \midrule
    ED\textsubscript{MS} & [s2\_ms\_10] \{Instruction\} Answer the question using a single word. & [s2\_ms\_10] \{Instruction\} Answer with exactly one of A, B, C, or D. & [refer] [s2\_ms\_10] \{Instruction\} & [caption] [s2\_ms\_10] \{Instruction\} \\
    \midrule
    EarthMind & \{Instruction\} Answer with a single word: "yes"or "no". & \{Instruction\} Answer with exactly one of a, b, c, d. & \{Instruction\} & \{Instruction\} \\
    \midrule
    GPT-5.2 & \{Instruction\} Answer with a single word: "yes" or "no". & \{Instruction\} Answer using either a), b), c), or d). & \{Instruction\} Answer by only providing the bounding box in the format of: [x\_min, y\_min, x\_max, y\_max] where (x\_min, y\_min) is the top-left corner and (x\_max, y\_max) is the bottom-right corner and all values are between 0 and 1. The origin of the image is at the top-left corner. & \{Instruction\} Answer in form of a descriptive, unformatted caption." \\
    \midrule
    Qwen & \{Instruction\} & \{Instruction\} & \{Instruction\} Output ONLY the bounding box as [x1, y1, x2, y2]. Values in the ranges of 0 to 1. & \{Instruction\} \\
    \midrule
    GLM & \{Instruction\} & \{Instruction\} & \{Instruction\} Output ONLY the bounding box as [x1, y1, x2, y2]. Values in the ranges of 0 to 1. & \{Instruction\} \\
    \midrule
    LLaVA & \{Instruction\} & \{Instruction\} & \{Instruction\} Output ONLY the bounding box as [x1, y1, x2, y2]. Values in the ranges of 0 to 1. & \{Instruction\} \\
    \midrule
    \textbf{InternVL$^*$} & \{Instruction\} & \{Instruction\} & \{Instruction\} & \{Instruction\} \\
    \bottomrule
    \end{tabular}\label{tab:inferenceprompts}
\end{table*}

%% file: figures/supplementary/qualitatives.tex
\definecolor{panelblue}{RGB}{210,236,242}
\definecolor{choicegreen}{RGB}{30,140,70}
\definecolor{wrongred}{RGB}{200,50,50}
\definecolor{lightgray}{RGB}{245,245,245}

\newcommand{\imgplaceholder}[2]{%
  \fcolorbox{gray!60}{lightgray}{%
    \parbox[c][#2][c]{#1}{\centering \footnotesize Image placeholder}%
  }%
}

\newcommand{\modelrow}[3]{%
  \texttt{\footnotesize #1} & #2 & #3 \\
}

\newcommand{\modelrowBB}[2]{%
  \texttt{\footnotesize #1} & #2 \\
}

\newcommand{\qaPanel}[9]{%
\begin{minipage}[b]{0.48\textwidth}
\centering\small
\textbf{#1}\par
\vspace{0.4em}

\begin{minipage}[t]{0.47\linewidth}\centering #2\\S2\end{minipage}\hfill
\begin{minipage}[t]{0.47\linewidth}\centering #3\\S1\end{minipage}\par

\vspace{0.8em}

\ifx&#4&\else
\begin{minipage}[t]{0.95\linewidth}\centering #4\end{minipage}\par
\vspace{0.3em}
\fi

\vspace{0.6em}
\setlength{\tabcolsep}{3pt}
\renewcommand{\arraystretch}{1.15}

\begin{tabularx}{\linewidth}
{>{\raggedright\arraybackslash\footnotesize\ttfamily}p{4em}
 >{\RaggedRight\arraybackslash}X
 >{\centering\arraybackslash}m{3.2em}}

\rowcolor{lightgray} \modelrow{Model}{Answer}{Correct}

#7

\end{tabularx}

\vspace{0.6em}

\noindent
\begin{tabular}[t]{@{}p{0.44\linewidth}@{\hfill}p{0.55\linewidth}@{}}

\adjustbox{valign=t}{%
\begin{minipage}{0.99\linewidth}\centering
#8\\Reference map
\end{minipage}}

&
\adjustbox{valign=t}{%
\begin{minipage}{0.99\linewidth}\tiny\raggedright
#9
\end{minipage}}

\end{tabular}\par

\end{minipage}
}

\newcommand{\qaPanelBB}[9]{%
\begin{minipage}[b]{0.48\textwidth}
\centering\small
\textbf{#1}\par
\vspace{0.4em}

\begin{minipage}[t]{0.47\linewidth}\centering #2\\S2\end{minipage}\hfill
\begin{minipage}[t]{0.47\linewidth}\centering #3\\S1\end{minipage}\par

\vspace{0.8em}

\ifx&#4&\else
\begin{minipage}[t]{0.95\linewidth}\centering #4\end{minipage}\par
\vspace{0.3em}
\fi

\vspace{0.6em}
\setlength{\tabcolsep}{3pt}
\renewcommand{\arraystretch}{1.15}

\begin{tabularx}{\linewidth}
{>{\raggedright\arraybackslash\footnotesize\ttfamily}p{4em}
 >{\RaggedRight\arraybackslash}X}

\rowcolor{lightgray} \modelrowBB{Model}{Answer}

#7

\end{tabularx}

\vspace{0.6em}

\noindent
\begin{tabular}[t]{@{}p{0.44\linewidth}@{\hfill}p{0.55\linewidth}@{}}

\adjustbox{valign=t}{%
\begin{minipage}{0.99\linewidth}\centering
#8\\Reference map
\end{minipage}}

&
\adjustbox{valign=t}{%
\begin{minipage}{0.99\linewidth}\tiny\raggedright
#9
\end{minipage}}

\end{tabular}\par

\end{minipage}
}

\setlength{\fboxsep}{0pt}
\begin{figure}[htbp]
\centering
\qaPanel
  {Do regions of any coniferous forest directly adjoin a mixed forest?}
  {\includegraphics[width=\BenImgWidth]{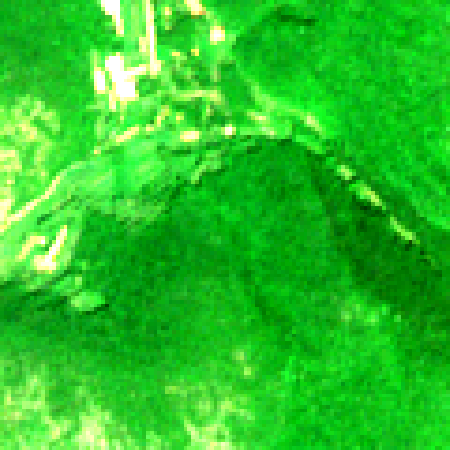}}
  {\includegraphics[width=\BenImgWidth]{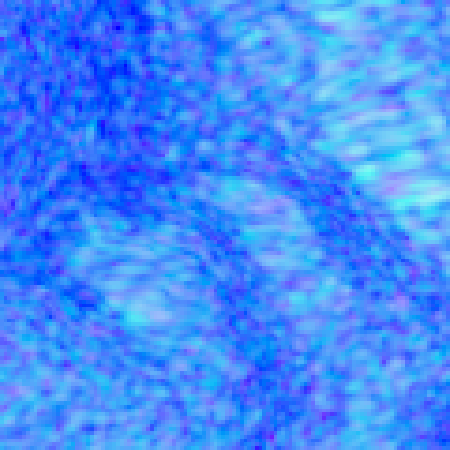}}
  {}
  {Interchange}{\cmark}
  {%
    \rowcolor{white}      \modelrow{\geochatABBRns}{yes}{\rxmark}
    \rowcolor{panelblue}  \modelrow{\lhrsbotABBRns}{yes}{\rxmark}
    \rowcolor{white}      \modelrow{\skyeyegptABBRns}{yes}{\rxmark}
    \rowcolor{panelblue}  \modelrow{\earthdialrgbABBRns}{no}{\gcmark}
    \rowcolor{white}      \modelrow{\earthdialmsABBRns}{yes}{\rxmark}
    \rowcolor{panelblue}  \modelrow{\earthmindrgbABBRns}{yes}{\rxmark}
    \rowcolor{white}      \modelrow{\earthmindsarmsABBRns}{yes}{\rxmark}
    \rowcolor{panelblue}  \modelrow{\gptABBRns}{yes}{\rxmark}
    \rowcolor{white}      \modelrow{\qwenABBRns}{yes}{\rxmark}
    \rowcolor{panelblue}  \modelrow{\glmABBRns}{yes}{\rxmark}
    \rowcolor{white}      \modelrow{\llavaABBRns}{yes}{\rxmark}
    \rowcolor{panelblue}  \modelrow{\internvlABBRns}{no}{\gcmark}
    \rowcolor{white}      \modelrow{\internvlrsABBRns}{no}{\gcmark}
  }
  {\includegraphics[width=\BenImgWidth]{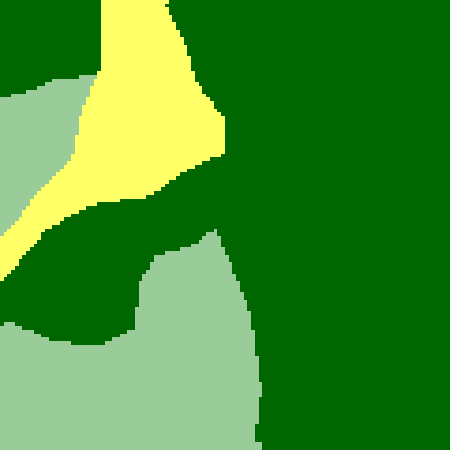}}
  {
    \scriptsize 
    \legendbox{ArableLand211}{Arable land}\\
    \legendbox{BroadLeavedForest311}{Broad-leaved forest}\\
    \legendbox{TransitionalWoodlandShrub324}{Transitional woodland, shrub}
  }
\hfill
\qaPanel
  {Is there only a single continuous region of urban fabrics visible in the image?}
  {\includegraphics[width=\BenImgWidth]{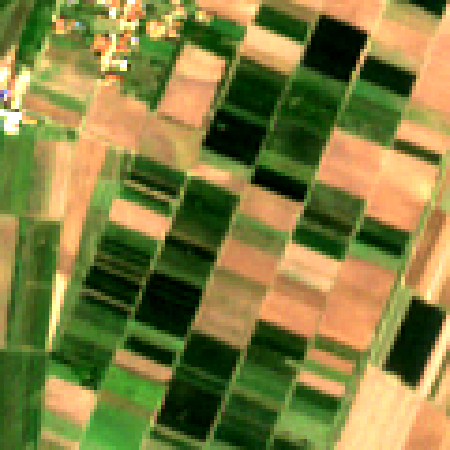}}
  {\includegraphics[width=\BenImgWidth]{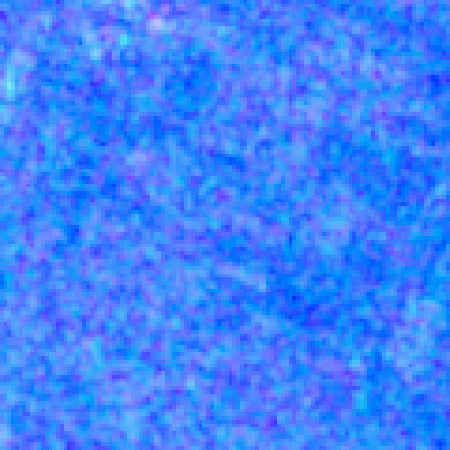}}
  {}
  {Interchange}{\cmark}
  {%
    \rowcolor{white}      \modelrow{\geochatABBRns}{no}{\rxmark}
    \rowcolor{panelblue}  \modelrow{\lhrsbotABBRns}{yes}{\gcmark}
    \rowcolor{white}      \modelrow{\skyeyegptABBRns}{yes}{\gcmark}
    \rowcolor{panelblue}  \modelrow{\earthdialrgbABBRns}{yes}{\gcmark}
    \rowcolor{white}      \modelrow{\earthdialmsABBRns}{no}{\rxmark}
    \rowcolor{panelblue}  \modelrow{\earthmindrgbABBRns}{no}{\rxmark}
    \rowcolor{white}      \modelrow{\earthmindsarmsABBRns}{no}{\rxmark}
    \rowcolor{panelblue}  \modelrow{\gptABBRns}{no}{\rxmark}
    \rowcolor{white}      \modelrow{\qwenABBRns}{no}{\rxmark}
    \rowcolor{panelblue}  \modelrow{\glmABBRns}{yes}{\gcmark}
    \rowcolor{white}      \modelrow{\llavaABBRns}{no}{\rxmark}
    \rowcolor{panelblue}  \modelrow{\internvlABBRns}{no}{\rxmark}
    \rowcolor{white}      \modelrow{\internvlrsABBRns}{yes}{\gcmark}
  }
  {\includegraphics[width=\BenImgWidth]{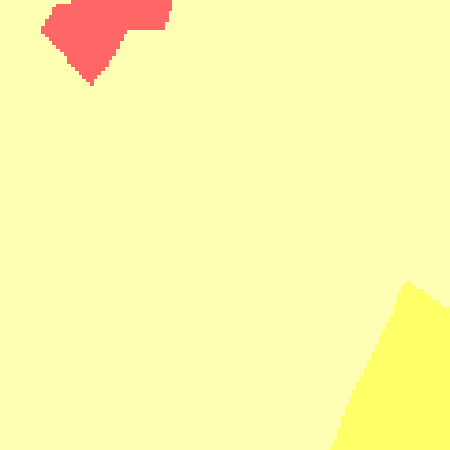}}
  {
    \scriptsize
    \legendbox{ArableLand211}{Arable land}\\
    \legendbox{ComplexCultivationPatterns242}{Complex cultivation patterns}\\
    \legendbox{UrbanFabric112}{Urban fabric}
  }
\caption{Qualitative examples for binary \ac{VQA} capabilites of considered models.
The S2 input is visualized as the true color RGB representation of the S2 image and 
the S1 input is visualized as (RGB: VV, VH, VV/VH). 
We provide the reference maps as aid to the reader only. Reference maps are not provided as input to the model.}
\label{fig:supp_qualitatives1}
\end{figure}

\begin{figure}[htbp]
\centering
\qaPanel
  {Identify the country depicted in the satellite image:}
  {\includegraphics[width=\BenImgWidth]{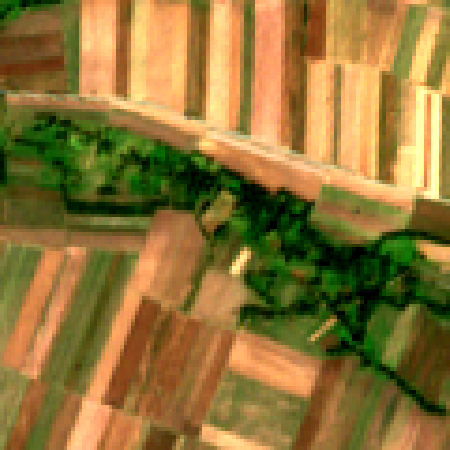}}
  {\includegraphics[width=\BenImgWidth]{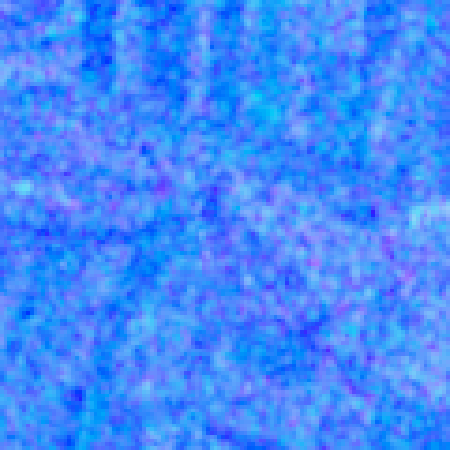}}
  {a)\ Austria\\ b)\ Switzerland\\ c) Luxembourg\\ 
   {\color{choicegreen}d) Serbia}}
  {Stadium}{\cmark}
  {%
    \rowcolor{white}      \modelrow{\geochatABBRns}{a}{\rxmark}
    \rowcolor{panelblue}  \modelrow{\lhrsbotABBRns}{d}{\gcmark}
    \rowcolor{white}      \modelrow{\skyeyegptABBRns}{c}{\rxmark}
    \rowcolor{panelblue}  \modelrow{\earthdialrgbABBRns}{a}{\rxmark}
    \rowcolor{white}      \modelrow{\earthdialmsABBRns}{a}{\rxmark}
    \rowcolor{panelblue}  \modelrow{\earthmindrgbABBRns}{c}{\rxmark}
    \rowcolor{white}      \modelrow{\earthmindsarmsABBRns}{d}{\gcmark}
    \rowcolor{panelblue}  \modelrow{\gptABBRns}{c}{\rxmark}
    \rowcolor{white}      \modelrow{\qwenABBRns}{a}{\rxmark}
    \rowcolor{panelblue}  \modelrow{\glmABBRns}{c}{\rxmark}
    \rowcolor{white}      \modelrow{\llavaABBRns}{b}{\rxmark}
    \rowcolor{panelblue}  \modelrow{\internvlABBRns}{b}{\rxmark}
    \rowcolor{white}      \modelrow{\internvlrsABBRns}{d}{\gcmark}
  }
  {\includegraphics[width=\BenImgWidth]{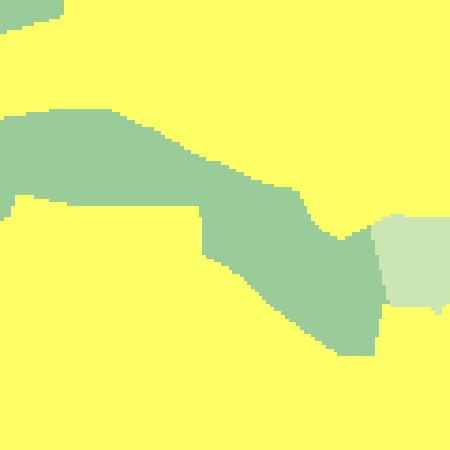}}
  {
    \scriptsize
    \legendbox{ArableLand211}{Arable land}\\
    \legendbox{LandNaturalVegetation243}{Land principally occupied by agriculture, with significant areas of natural vegetation}\\
    \legendbox{TransitionalWoodlandShrub324}{Transitional woodland, shrub}
  }
\hfill
\qaPanel
  {From the following classes, select the one that is present:}
  {\includegraphics[width=\BenImgWidth]{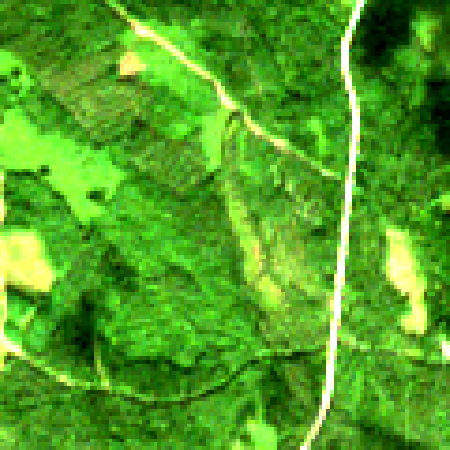}}
  {\includegraphics[width=\BenImgWidth]{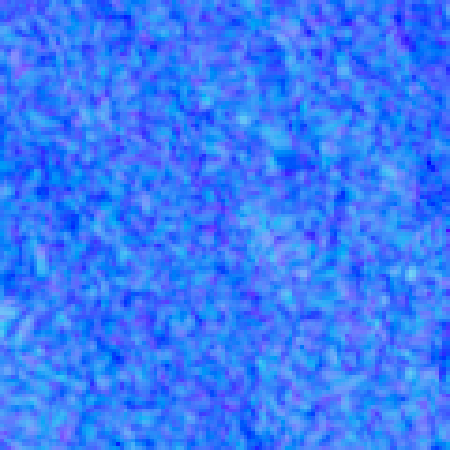}}
  {a)\ Broad-leaved forest\\ {\color{choicegreen} b)\ Coniferous forest}\\ c)\ Coastal wetlands\\ d)\ Inland wetlands}
  {Mixed cereal}{\cmark}
  {%
    \rowcolor{white}      \modelrow{\geochatABBRns}{a}{\rxmark}
    \rowcolor{panelblue}  \modelrow{\lhrsbotABBRns}{a}{\rxmark}
    \rowcolor{white}      \modelrow{\skyeyegptABBRns}{a}{\rxmark} 
    \rowcolor{panelblue}  \modelrow{\earthdialrgbABBRns}{--}{\rxmark}
    \rowcolor{white}      \modelrow{\earthdialmsABBRns}{d}{\rxmark}
    \rowcolor{panelblue}  \modelrow{\earthmindrgbABBRns}{--}{\rxmark}
    \rowcolor{white}      \modelrow{\earthmindsarmsABBRns}{a}{\rxmark}
    \rowcolor{panelblue}  \modelrow{\gptABBRns}{a}{\rxmark}
    \rowcolor{white}      \modelrow{\qwenABBRns}{a}{\rxmark}
    \rowcolor{panelblue}  \modelrow{\glmABBRns}{a}{\rxmark}
    \rowcolor{white}      \modelrow{\llavaABBRns}{a}{\rxmark}
    \rowcolor{panelblue}  \modelrow{\internvlABBRns}{a}{\rxmark}
    \rowcolor{white}      \modelrow{\internvlrsABBRns}{b}{\gcmark}
  }
  {\includegraphics[width=\BenImgWidth]{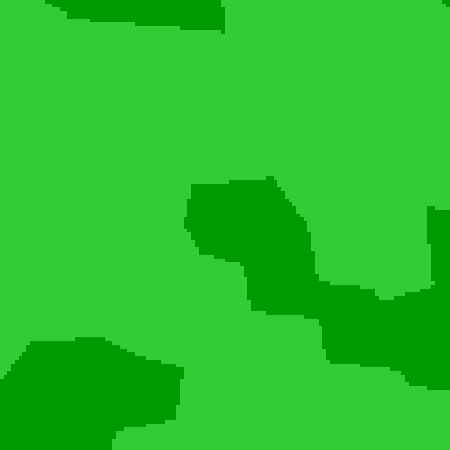}}
  {
    \scriptsize
    \legendbox{ConiferousForest312}{Coniferous forest}\\
    \legendbox{MixedForest313}{Mixed forest}
  }
\caption{Qualitative examples for \ac{MCQ} capabilities of considered models.
The S2 input is visualized as the true color RGB representation of the S2 image and 
the S1 input is visualized as (RGB: VV, VH, VV/VH).
We provide the reference maps as aid to the reader only. Reference maps are not provided as input to the model. \enquote{--} indicates that the answer could not be extracted.}
\label{fig:supp_qualitatives2}
\end{figure}
\begin{figure}[htbp]
\centering
\qaPanelBB
  {Point out a continuous area of broad-leaved forest in the scene.}
  {\includegraphics[width=\BenImgWidth]{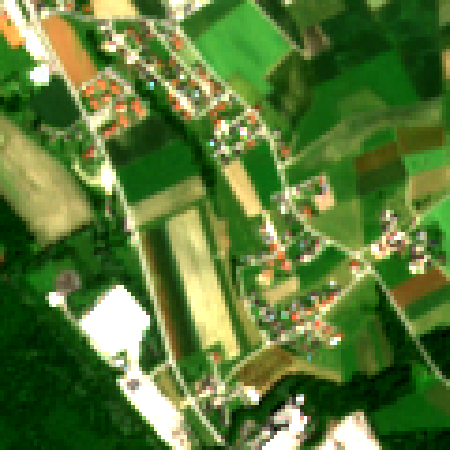}}
  {\includegraphics[width=\BenImgWidth]{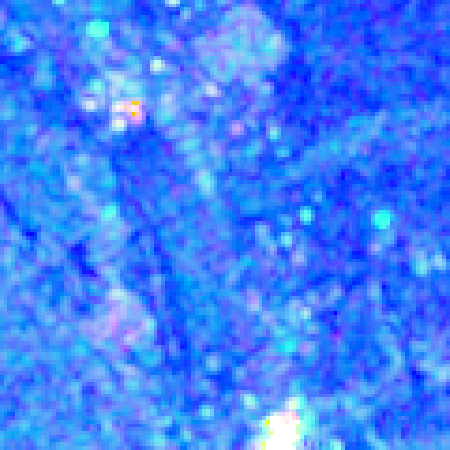}}
  {\color{choicegreen} [0.0, 0.44, 0.35, 1.0]}
  {Stadium}{\cmark}
  {%
    \rowcolor{white}      \modelrowBB{\geochatABBRns}{[0.0, 0.57, 0.24, 0.89]}
    \rowcolor{panelblue}  \modelrowBB{\lhrsbotABBRns}{[0.45, 0.45, 0.89, 0.89]}
    \rowcolor{white}      \modelrowBB{\skyeyegptABBRns}{[0.65, 0.19, 0.83, 0.4]}
    \rowcolor{panelblue}  \modelrowBB{\earthdialrgbABBRns}{[0.0, 0.79, 0.21, 0.91]}
    \rowcolor{white}      \modelrowBB{\earthdialmsABBRns}{--}
    \rowcolor{panelblue}  \modelrowBB{\earthmindrgbABBRns}{[0.08, 0.05, 0.52, 0.97]}
    \rowcolor{white}      \modelrowBB{\earthmindsarmsABBRns}{[0.74, 0.25, 0.98, 0.63]}
    \rowcolor{panelblue}  \modelrowBB{\gptABBRns}{[0.02, 0.29, 0.34, 0.97]}
    \rowcolor{white}      \modelrowBB{\qwenABBRns}{[0.0, 0.0, 0.17, 0.25]}
    \rowcolor{panelblue}  \modelrowBB{\glmABBRns}{[0.0, 0.0, 1.0, 1.0]}
    \rowcolor{white}      \modelrowBB{\llavaABBRns}{[0.0, 0.67, 0.21, 0.99]}
    \rowcolor{panelblue}  \modelrowBB{\internvlABBRns}{--}
    \rowcolor{white}      \modelrowBB{\internvlrsABBRns}{[0.0, 0.45, 0.35, 1.0]}
  }
  {
  \includegraphics[width=\BenImgWidth]{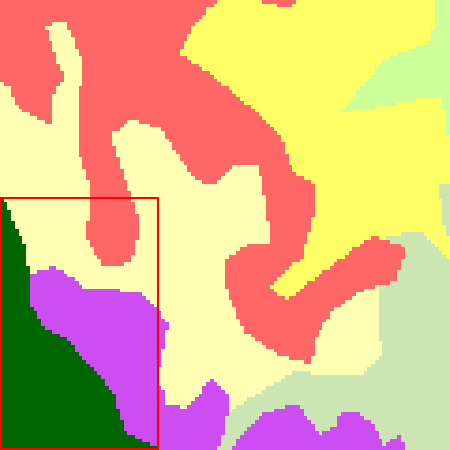}
  }
  {
    \scriptsize
    \legendbox{UrbanFabric112}{Urban fabric}\\
    \legendbox{IndustrialCommercialUnits121}{Industrial or commercial units}\\
    \legendbox{ArableLand211}{Arable land}\\
    \legendbox{Pastures231}{Pastures}\\
    \legendbox{ComplexCultivationPatterns242}{Complex cultivation patterns}\\
    \legendbox{LandNaturalVegetation243}{Land principally occupied by agriculture, with significant areas of natural vegetation}\\
    \legendbox{BroadLeavedForest311}{Broad-leaved forest}
  }
\hfill
\qaPanelBB
  {Output a bounding box surrounding the land cover class instance positioned at (0.92, 0.05) in the image.}
  {\includegraphics[width=\BenImgWidth]{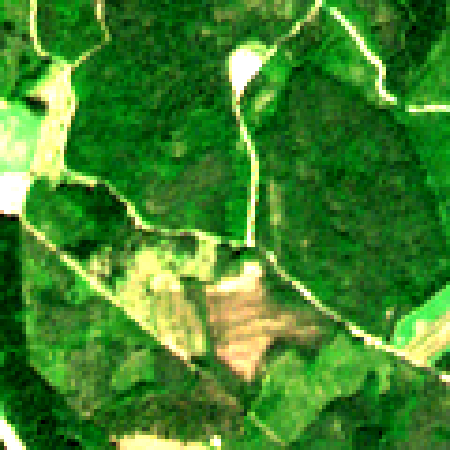}}
  {\includegraphics[width=\BenImgWidth]{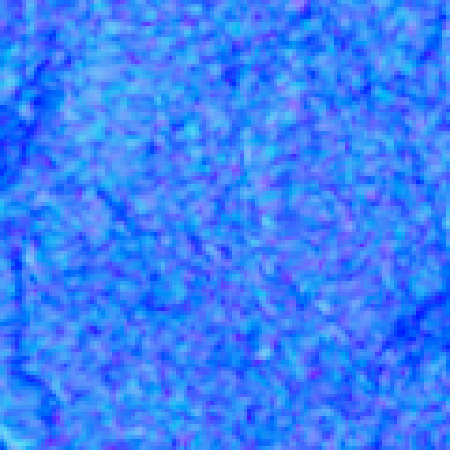}}
  {\color{choicegreen} [0.81, 0.0, 1.0, 0.12]}
  {Stadium}{\cmark}
  {%
    \rowcolor{white}      \modelrowBB{\geochatABBRns}{[0.84, 0.0, 0.96, 0.2]}
    \rowcolor{panelblue}  \modelrowBB{\lhrsbotABBRns}{[0.92, 0.05, 0.92, 0.05]}
    \rowcolor{white}      \modelrowBB{\skyeyegptABBRns}{--}
    \rowcolor{panelblue}  \modelrowBB{\earthdialrgbABBRns}{[0.06, 0.49, 0.1, 0.55, 0.0]}
    \rowcolor{white}      \modelrowBB{\earthdialmsABBRns}{--}
    \rowcolor{panelblue}  \modelrowBB{\earthmindrgbABBRns}{[0.0, 0.23, 0.08, 0.92]}
    \rowcolor{white}      \modelrowBB{\earthmindsarmsABBRns}{[0.86, 0.22, 0.86, 0.22]}
    \rowcolor{panelblue}  \modelrowBB{\gptABBRns}{[0.0, 0.0, 1.0, 1.0]}
    \rowcolor{white}      \modelrowBB{\qwenABBRns}{[0.91, 0.04, 1.0, 0.08]}
    \rowcolor{panelblue}  \modelrowBB{\glmABBRns}{[0.85, 0.0, 1.0, 0.26]}
    \rowcolor{white}      \modelrowBB{\llavaABBRns}{[0.786, 0.054, 0.998, 0.308]}
    \rowcolor{panelblue}  \modelrowBB{\internvlABBRns}{[0.93, 0.29, 0.94, 0.61]}
    \rowcolor{white}      \modelrowBB{\internvlrsABBRns}{[0.8, 0.0, 1.0, 0.13]}
  }
  {\includegraphics[width=\BenImgWidth]{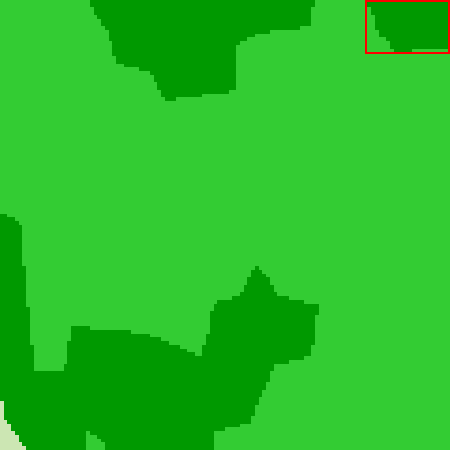}}
  {
    \scriptsize
    \legendbox{LandNaturalVegetation243}{Land principally occupied by agriculture, with significant areas of natural vegetation}\\
    \legendbox{ConiferousForest312}{Coniferous forest}\\
    \legendbox{MixedForest313}{Mixed forest}
  }
\caption{Qualitative examples for \genDet{} capabilities of considered models. 
The S2 input is visualized as the true color RGB representation and 
the S1 input is visualized as (RGB: VV, VH, VV/VH).
Reference maps are not provided as input to the model. The instance that is to be detected is marked by a red rectangle. \enquote{--} indicates that the answer could not be extracted.}
\label{fig:supp_qualitatives3}
\end{figure}

%% file: figures/supplementary/examples_caption.tex
\begin{figure*}[h!t]
\centering
\begin{tikzpicture}[x=1cm,y=1cm]
\node[
    anchor=north west,
    text width=0.99\linewidth,
    style={
        draw=boxblue,
        rounded corners=3pt,
        line width=0.8pt,
        fill=lightbg,
        inner sep=2pt,
        align=left
    }
] (gt) at (0.0, 10.0){
\begin{minipage}{\linewidth}
\centering
\begin{minipage}{0.3\textwidth}
\iftrue
    \setlength{\tabcolsep}{1pt}
    \begin{tabular}{cc}
    \includegraphics[width=\BenImgWidth]{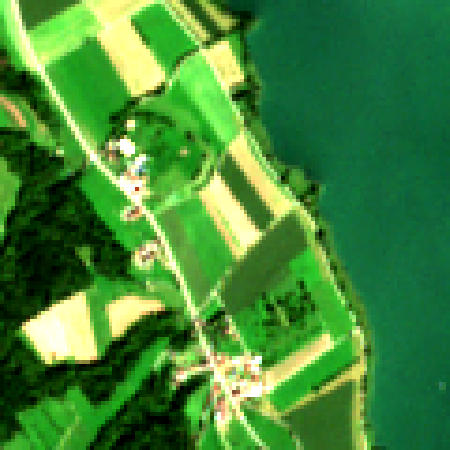} & \includegraphics[width=\BenImgWidth]{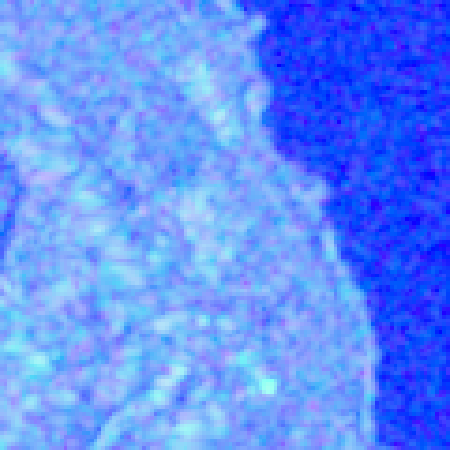} \\ 
    S2 & S1
    \end{tabular}
\fi
\end{minipage}
\begin{minipage}{0.69\textwidth}
\scriptsize
\textbf{Instruction:} Explain what this satellite image shows, mentioning the location, season, and landscape features.\newline\newline
\textbf{Reference:} This satellite image, captured during the summer season in Switzerland, showcases a diverse landscape within the \enquote{temperate, no dry season, warm summer} climate zone. The dominant feature is arable land (770'000 sqm), which spans two areas, one large ($\sim$761'000 sqm) and one marginal, covering a small fraction of the total area. The arable land borders a substantial body of inland waters ($\sim$440'000 sqm) and a smaller patch of mixed forest ($\sim$230'000 sqm). The landscape is characterized by a mix of agricultural areas, water bodies, and forest and seminatural areas, highlighting the varied ecosystems present in the region.
\end{minipage}
\end{minipage}
};

\node[
    anchor=north west,
    row1captionbox
] (earthmindRGB) 
at ($ (gt.west |- gt.south) + (0.0cm,-0.15cm) $)
{
\begin{minipage}{\linewidth}
\footnotesize
\textbf{EarthMind (RGB)}\hfill\scriptsize{\textcolor{gray}{RGB}}
\smallskip
\hrule
\smallskip
\scriptsize
This satellite image from GoogleEarth shows a rural landscape with a mix of agricultural fields and forested areas. The fields are primarily located in the central and bottom left sections of the image, with a notable concentration of greenery. The forested areas are more prominent in the top left and bottom right sections, with a dense cluster of trees. A water body is visible in the top right corner, adding to the rural character of the scene. The image is taken in a season where the vegetation is in full, indicating a time of growth or harvest. The overall scene is a blend of agricultural land and natural forest, with a water feature adding to the rural setting.
\end{minipage}
};

\node[
    anchor=north west,
    row1captionbox
] (earthdialRGB) 
at ($ (earthmindRGB.west |- earthmindRGB.south) + (0.0cm,-0.15cm) $)
{
\begin{minipage}{\linewidth}
\footnotesize
\textbf{EarthDial (RGB)}\hfill\scriptsize \textcolor{gray}{RGB}
\smallskip
\hrule
\smallskip
\scriptsize
The satellite image shows a large area of farmland and a dense forest near the coast. The farmland is located in the center of the image, while the forest is located at the bottom left. The coast is visible at the top right of the image. The image likely represents a coastal region with a mix of agricultural and natural landscapes.
\end{minipage}
};

\node[
    anchor=north west,
    row1captionbox
] (geochat)
at ($ (earthdialRGB.west |- earthdialRGB.south) + (0.0cm,-0.15cm) $)
{
\begin{minipage}{\linewidth}
\footnotesize
\textbf{GeoChat}\hfill\scriptsize \textcolor{gray}{RGB}
\smallskip
\hrule
\smallskip
\scriptsize
The satellite image shows a diverse landscape with a mix of urban and natural features. The center of the image displays a roundabout, which is likely a busy intersection in a city or town. Surrounding the roundabout are various buildings, indicating a dense urban area. In the background, there are trees and a roundabout, suggesting a well-balanced blend of urban and green spaces. The presence of a roundabout and trees together indicates that the location is likely a well-developed and well-maintained area.
\end{minipage}
};

\node[
    anchor=north west,
    row2captionbox
] (skyeye) 
at ($ (earthmindRGB.east |- gt.south) + (0.1cm,-0.15cm) $)
{
\begin{minipage}{\linewidth}
\footnotesize
\textbf{SkyEyeGPT}\hfill\scriptsize \textcolor{gray}{RGB}
\smallskip
\hrule
\smallskip
\scriptsize
many farmlands are near a body of water
\end{minipage}
};

\node[
    anchor=north west,
    row2captionbox
] (earthmindS1S2) 
at ($ (skyeye.west |- skyeye.south) + (0.cm,-0.15cm) $)
{
\begin{minipage}{\linewidth}
\footnotesize
\textbf{EarthMind (S1S2)}\hfill\scriptsize \textcolor{gray}{S1,S2}
\smallskip
\hrule
\smallskip
\scriptsize
This image shows a landscape with a mix of agricultural fields and forested areas. The fields are primarily located in the central and lower parts of the image, with a notable concentration in the bottom right. The forested areas are more prominent in the upper left and bottom left sections. The image is likely taken in a rural setting, possibly in a region with a combination of agricultural and natural land use. The presence of a water body on the right side suggests proximity to a river or lake, which could be a significant feature for the surrounding land use.
\end{minipage}
};

\node[
    anchor=north west,
    row2captionbox
] (earthdialS2) 
at ($ (earthmindS1S2.west |- earthmindS1S2.south) + (0.0cm,-0.15cm) $)
{
\begin{minipage}{\linewidth}
\footnotesize
\textbf{EarthDial (S2)}\hfill\scriptsize \textcolor{gray}{S2}
\smallskip
\hrule
\smallskip
\scriptsize
The satellite image shows a land mixed in the left of the image, water in the right of the image.
\end{minipage}
};

\path let \p1 = (earthdialS2.south), \p2 = (geochat.south) in
node[
    anchor=north west,
    minimum height={\y1 - \y2 - 0.15cm -0.8pt},
    row2captionbox
] (rsIntern) 
at ($ (earthdialS2.west |- earthdialS2.south) + (0.0cm,-0.15cm) $)
{
\begin{minipage}{\linewidth}
\footnotesize
\textbf{InternVL$^*$}\hfill\scriptsize \textcolor{gray}{RGB,S1,S2}
\smallskip
\hrule
\smallskip
\scriptsize
The image, captured during the summer season in Switzerland, showcases a diverse landscape within the \enquote{temperate, no dry season, warm summer} climate zone. The dominant feature is arable land, spanning approximately 787,000 square meters, which is distributed across two areas, one large ($\sim$786,000 sqm) and one marginal. The arable land borders with inland waters ($\sim$350,000 sqm), mixed forest ($\sim$209,000 sqm), and urban fabric ($\sim$72,000 sqm). The inland waters are adjacent to the mixed forest, which is spread over two areas, one smaller ($\sim$199,000 sqm) and one marginal. The landscape presents a mix of agricultural areas, water bodies, and artificial surfaces, highlighting the varied ecosystems within this region.
\end{minipage}
};

\path let \p1 = (earthmindRGB.north), \p2 = (geochat.south) in
node[
    anchor=north west,
    minimum height={\y1-\y2-1pt},
    lhrsbox
] (lhrsBot) 
at ($ (skyeye.east |- gt.south) + (0.1cm,-0.15cm) $)
{
\begin{minipage}{\linewidth}
\footnotesize
\textbf{LHRS-Bot}\hfill\scriptsize \textcolor{gray}{RGB}
\smallskip
\hrule
\smallskip
\scriptsize
Based on the provided image, the location appears to be a rural area with a mix of agricultural lands, forests, and water bodies. The image covers an area of approximately 144,000 square meters.\\
Location:The image appears to be located in a temperate region, with a mix of flat and hilly terrain. The landscape features rolling hills, valleys, and a large lake in the center of the image.\\
Season:The image appears to be captured during the summer season, with lush green vegetation and full crops. The trees are full of leaves, and the fields are filled with ripe crops, indicating a warm and sunny weather.\\
Land Cover Types: 1. Agricultural lands: The image shows a mix of arable lands, pastures, and orchards. The arable lands are cultivated with various crops, including corn, soybeans, and wheat. The pastures are home to livestock, such as cows and sheep. The orchards are filled with fruit trees, such as apple and pear trees. 2. Forests: The image features a mix of deciduous and evergreen forests. The deciduous forests are dominated by trees such as oak, maple, and beech, which shed their leaves in the fall. The evergreen forests are home to coniferous trees such as pine, spruce, and fir. Water bodies: The image features a large lake in the center, surrounded by smaller ponds and streams. The lake is home to a variety of aquatic plants and animals, including fish, birds, and aquatic mammals. 4. Residential areas: The image shows a few small residential areas, likely consisting of single-family homes and farms. 5. Roads and infrastructure: The image features a network of roads, including a main road that runs through the center of the image. Other infrastructure, such as bridges, can be seen in the distance. 6. Recreational areas: The image shows a few small parks and recreational areas, likely used for activities such as hiking, fishing, and boating. Overall, the image depicts a rural landscape with a mix of agricultural lands, forests, water bodies, and residential areas. The season is likely summer, with lush green vegetation and full crops.
\end{minipage}
};

\end{tikzpicture}

\caption{A qualitative example of the captioning capabilities of considered \ac{RS} models. The S2 input is visualized as the true color RGB representation of the S2 image and 
the S1 input is visualized as (RGB: VV, VH, VV/VH). 
The input modality for each model is noted in the top-right corner of the respective model answer.}
\label{fig:qual_caption_skeleton}
\end{figure*}

%% file: figures/supplementary/examples_captions_cv.tex
\begin{figure*}[h!t]
\centering
\begin{tikzpicture}[x=1cm,y=1cm]

\node[
    anchor=north west,
    text width=0.99\linewidth,
    style={
        draw=boxblue,
        rounded corners=3pt,
        line width=0.8pt,
        fill=lightbg,
        inner sep=2pt,
        align=left
    }
] (gt) at (0.0, 10.0){
\begin{minipage}{\linewidth}
\centering
\begin{minipage}{0.3\textwidth}
\iftrue
    \setlength{\tabcolsep}{1pt}
    \begin{tabular}{cc}
    \includegraphics[width=\BenImgWidth]{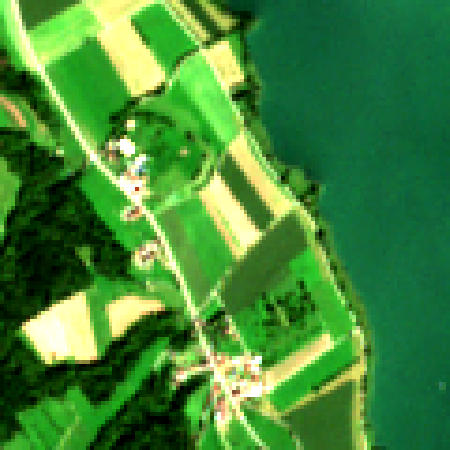} & \includegraphics[width=\BenImgWidth]{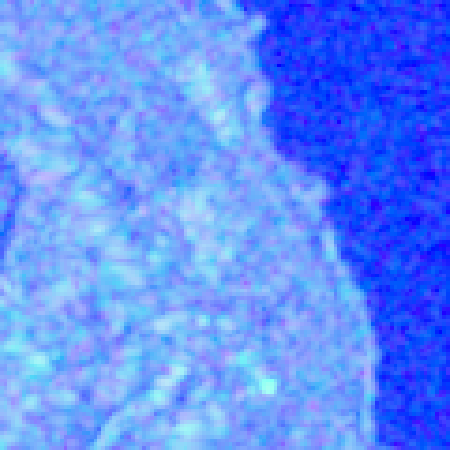} \\ 
    S2 & S1
    \end{tabular}
\fi
\end{minipage}
\begin{minipage}{0.69\textwidth}
\scriptsize
\textbf{Instruction:} Explain what this satellite image shows, mentioning the location, season, and landscape features.\newline\newline
\textbf{Reference:} This satellite image, captured during the summer season in Switzerland, showcases a diverse landscape within the \enquote{temperate, no dry season, warm summer} climate zone. The dominant feature is arable land (770'000 sqm), which spans two areas, one large ($\sim$761'000 sqm) and one marginal, covering a small fraction of the total area. The arable land borders a substantial body of inland waters ($\sim$440'000 sqm) and a smaller patch of mixed forest ($\sim$230'000 sqm). The landscape is characterized by a mix of agricultural areas, water bodies, and forest and seminatural areas, highlighting the varied ecosystems present in the region.
\end{minipage}
\end{minipage}
};
\node[
    anchor=north west,
    captionbox,
    text width=0.31\linewidth
] (intern)
at ($ (gt.west |- gt.south) + (0.0cm,-0.15cm) $)
{
\begin{minipage}{\linewidth}
\footnotesize
\textbf{InternVL3-1B}\hfill\scriptsize \textcolor{gray}{RGB}
\smallskip
\hrule
\smallskip
\scriptsize
The satellite image shows a rural landscape with distinct agricultural fields, indicating a countryside setting. Here's an explanation of what the image shows: 1. Location and Season: - The image area spans 144,000 square meters. - Given that it's a remote or aerial shot, it may have been taken in a time of year when the region is often in a different state of cultivation or growth, possibly indicating spring or summer. The vegetation seems lush, typical of a period when crops are actively harvested. 2. Landscape Features: - The landscape includes multiple fields of varying shades of green, with some sections showing more dense vegetation compared to others, possibly indicating different crops or stages of harvest. - There is a noticeable path or road cutting through the fields, running diagonally from the top left to the bottom right. This seems to be a main road, likely connecting different agricultural zones. - The fields are organized in strips and patterns, suggesting structured farming practices. The scattered structures among the fields could be farm buildings, barns, or storage facilities. - The surrounding area has dense greenery, indicating dense forests or possibly a wooded region. This helps in distinguishing the agricultural areas from the forested zones around them. - The water body to the right is likely a large lake, river, or inlet, typical of a riverine or estuarine setting. Its expansive size and darker color contrast with the green fields to the left, suggesting a significant body of water. Overall, the image captures a serene rural landscape with organized agricultural fields, a path for transportation, and natural surroundings suggesting a transitional period between different seasons or stages in agriculture.
\end{minipage}
};
\node[
    anchor=north west,
    captionbox,
    text width=0.31\linewidth
] (gpt) 
at ($ (intern.west |- intern.south) + (0.0cm,-0.15cm) $)
{
\begin{minipage}{\linewidth}
\footnotesize
\textbf{GPT-5.2}\hfill\scriptsize \textcolor{gray}{RGB}
\smallskip
\hrule
\smallskip
\scriptsize
A small rural area of patchwork farmland bordered by dense green woodland, likely in a temperate region during late spring or summer, showing rectangular crop fields in varying shades of green and tan, a few scattered farm buildings and houses connected by narrow roads, and sharp field boundaries and hedgerows across the roughly \SI{1.44}{\kilo\meter\squared} scene.
\end{minipage}
};

\node[
    anchor=north west,
    captionbox,
    text width=0.33\linewidth
] (llava) 
at ($ (intern.east|- intern.north) + (0.1cm,0.0cm) $)
{
\begin{minipage}{\linewidth}
\footnotesize
\textbf{LLaVA-OneVision}\hfill\scriptsize \textcolor{gray}{RGB}
\smallskip
\hrule
\smallskip
\scriptsize
The image shows a rural landscape with a variety of green fields, indicating a lush and fertile area. There is a body of water on the right side, which could be a river or a lake. The fields are divided into different sections, suggesting a mix of agricultural activities such as crop rotation or different types of farming. There are also some structures visible, which could be farmhouses or small buildings, indicating human habitation and activity in the area. The presence of trees and the overall greenery suggest that the season could be spring or summer. The image does not provide enough detail to determine the exact location or the specific types of crops being grown.
\end{minipage}
};

\node[
    anchor=north west,
    captionbox,
    text width=0.33\linewidth
] (glm) 
at ($ (llava.west |- llava.south) + (0.0cm,-0.15cm) $)
{
\begin{minipage}{\linewidth}
\footnotesize
\textbf{GLM-4.6-Flash}\hfill\scriptsize \textcolor{gray}{RGB}
\smallskip
\hrule
\smallskip
\scriptsize
The satellite image depicts a rural landscape with a mix of agricultural fields, residential structures, and a body of water (likely a lake or river). The location appears to be a countryside area, possibly in a temperate region given the green vegetation. The season is likely spring or summer, as the fields are lush and green, indicating active vegetation growth. Key landscape features include rectangular agricultural plots (suggesting organized farming), a cluster of buildings (residential or small-scale rural development), and a large water body to the right, which may serve as a natural boundary or resource for the area. The overall scene reflects a blend of human settlement and agricultural land use in a natural setting.
\end{minipage}
};

\path let \p1 = (glm.south), \p2 = (gpt.south) in
node[
    anchor=north west,
    captionbox,
    text width=0.33\linewidth,
    minimum height = {\y1 - \y2 - 0.18cm}
] (rsIntern) 
at ($ (glm.west |- glm.south) + (0.0cm,-0.15cm) $)
{
\begin{minipage}{\linewidth}
\footnotesize
\textbf{InternVL$^*$}\hfill\scriptsize \textcolor{gray}{RGB,S1,S2}
\smallskip
\hrule
\smallskip
\scriptsize
The image, captured during the summer season in Switzerland, showcases a diverse landscape within the \enquote{temperate, no dry season, warm summer} climate zone. The dominant feature is arable land, spanning approximately 787,000 square meters, which is distributed across two areas, one large ($\sim$786,000 sqm) and one marginal. The arable land borders with inland waters ($\sim$350,000 sqm), mixed forest ($\sim$209,000 sqm), and urban fabric ($\sim$72,000 sqm). The inland waters are adjacent to the mixed forest, which is spread over two areas, one smaller ($\sim$199,000 sqm) and one marginal. The landscape presents a mix of agricultural areas, water bodies, and artificial surfaces, highlighting the varied ecosystems within this region.
\end{minipage}
};

\path let \p1 = (intern.north), \p2 = (gpt.south) in
node[
    anchor=north west,
    captionbox,
    text width=0.32\linewidth,
    minimum height = {\y1 - \y2 - 1pt}
] (qwen) 
at ($ (llava.east |- intern.north) + (0.1cm,0.0cm) $)
{
\begin{minipage}{\linewidth}
\footnotesize
\textbf{Qwen3-VL}\hfill\scriptsize \textcolor{gray}{RGB}
\smallskip
\hrule
\smallskip
\scriptsize
Based on the satellite image and the provided area, here is an explanation of what the image shows: This satellite image displays a rural landscape located in the Czech Republic, specifically in the South Bohemian Region, near the town of Křešice (or possibly the nearby village of Křešice u České Lípy). The image captures a patchwork of agricultural fields adjacent to a large body of water, which is Lake Černé jezírko (Black Lake) or a similar local reservoir. Season: The image was likely taken during late spring or early summer. This is indicated by the vibrant green coloration of the fields, which suggests active growth of crops like cereals or vegetables. The absence of snow or bare soil rules out winter, and the lushness is more characteristic of spring/summer than autumn (when fields would be golden or harvested) or early winter. Landscape Features: - Agricultural Fields: The dominant feature is the grid-like pattern of cultivated land, divided into rectangular plots of varying shades of green and brown. These represent different crops or stages of growth, with some fields appearing darker green (possibly corn or lush grass) and others lighter or brownish (possibly wheat, barley, or fallow land).- Water Body: A large, dark green body of water occupies the right side of the image, indicating a lake or reservoir. Its smooth surface suggests calm conditions.- Roads and Settlements: A network of roads connects the fields and a small cluster of buildings visible near the center-left. These likely represent a rural village or hamlet.- Vegetation: Darker green patches, particularly along the edges of the fields and near the water, indicate forests or wooded areas, which are common in this part of the Czech countryside.- Topography: The terrain appears relatively flat, typical of the Bohemian lowlands, with slight variations in elevation suggested by the contours of the fields and the shoreline. In summary, the image shows a typical Czech rural landscape in late spring or early summer, featuring agricultural fields, a lake, and a small settlement, all within the South Bohemian Region. The 1,440,000 square meter area encompasses this entire scene, highlighting the interplay between agriculture, water, and human habitation.
\end{minipage}
};

\end{tikzpicture}

\caption{A qualitative example of the captioning task for considered \ac{CV} models. The provided input is the same as in \cref{fig:qual_caption_skeleton}. The input modality for each model is noted in the top-right corner of the respective model answer.
}
\label{fig:qual_caption_skeleton_cv}
\end{figure*}

%% file: figures/model_v1.tex
\begin{figure*}[h!bt]
    \centering

    \begin{tikzpicture}        
        \node (rgbIN) at (0,0)
            {\includegraphics[width=\BenImgWidth]{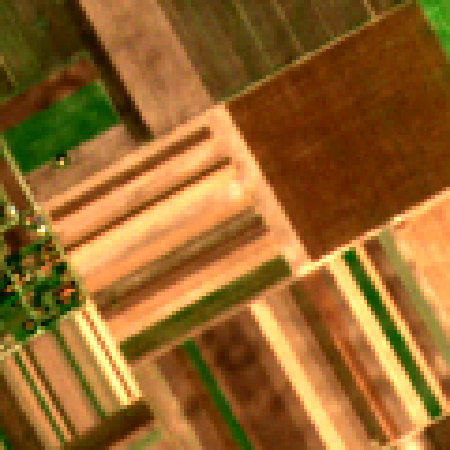}};
        \node[anchor=south] (rgbCap) at ($(rgbIN.north) + (0,-.2)$) {\small\textcolor{gray}{RGB bands of S2}};
        \node[anchor=east] (s1IN) at ($(rgbIN.west) + (-.2, 0)$)
            {\includegraphics[width=\BenImgWidth]{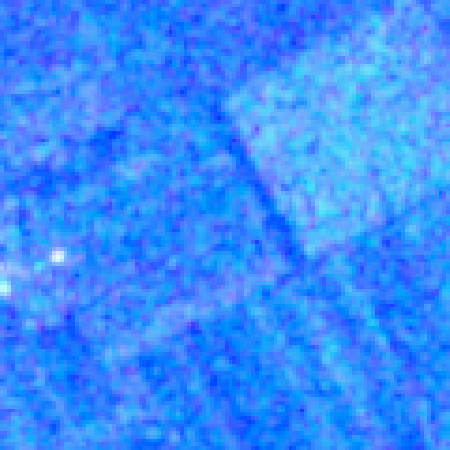}};
        \node[anchor=south] (s1Cap) at ($(s1IN.north) + (0,-.2)$) {\small\textcolor{gray}{S1}};
        \node[anchor=east] (s2IN) at ($(s1IN.west) + (-.2, 0)$)
            {\includegraphics[width=\BenImgWidth]{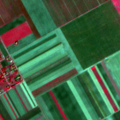}};
        \node[anchor=south] (s2Cap) at ($(s2IN.north) + (0,-.2)$) {\small\textcolor{gray}{S2}};
        \node[align=center, anchor=east, text width=\BenImgWidth, minimum width=\BenImgWidth, minimum height=\BenImgWidth, fill=gray!30, rounded corners=2mm] (Prompt) at ($(s2IN.west) + (-.2,0)$) {\small \textit{Describe the image.}};

        \node[align=center, anchor=north, text width=\BenImgWidth, minimum width=\BenImgWidth, minimum height=0.8\BenImgWidth, fill=green!20, rounded corners=2mm] (rgbViT) at ($(rgbIN.south) + (0,-.4)$) {\small \textbf{InternViT}};
        \node[align=center, anchor=north, text width=\BenImgWidth, minimum width=\BenImgWidth, minimum height=0.8\BenImgWidth, fill=orange!20, rounded corners=2mm] (s1ViT) at ($(s1IN.south) + (0,-.4)$) {\small \textbf{S1 ViT}};
        \node[align=center, anchor=north, text width=\BenImgWidth, minimum width=\BenImgWidth, minimum height=0.8\BenImgWidth, fill=orange!20, rounded corners=2mm] (s2ViT) at ($(s2IN.south) + (0,-.4)$) {\small \textbf{S2 ViT}};

        \node[align=center, anchor=north, text width=\BenImgWidth, minimum width=\BenImgWidth, minimum height=0.6\BenImgWidth, fill=green!20, rounded corners=2mm] (rgbProj) at ($(rgbViT.south) + (0,-.4)$) {\small \textbf{Projection}};
        \node[align=center, anchor=north, text width=\BenImgWidth, minimum width=\BenImgWidth, minimum height=0.6\BenImgWidth, fill=orange!20, rounded corners=2mm] (s1Proj) at ($(s1ViT.south) + (0,-.4)$) {\small \textbf{Projection}};
        \node[align=center, anchor=north, text width=\BenImgWidth, minimum width=\BenImgWidth, minimum height=0.6\BenImgWidth, fill=orange!20, rounded corners=2mm] (s2Proj) at ($(s2ViT.south) + (0,-.4)$) {\small \textbf{Projection}};
        \node[align=center, anchor=center, text width=\BenImgWidth, minimum width=\BenImgWidth, minimum height=0.6\BenImgWidth, fill=green!20, rounded corners=2mm] (tokenizer) at ($(Prompt |- s2Proj) + (0,0)$) {\small \textbf{Tokenizer}};

        \node[align=center, anchor=north, text width=\BenImgWidth,text width=108mm,inner sep=0pt, minimum height=0.6\BenImgWidth, fill=green!20, rounded corners=2mm] (InternLM) at ($(s1Proj.south)!0.5!(s2Proj.south) + (0,-.4)$) {\small \textbf{InternLM}};
        \node [trapezium, trapezium angle=-60, minimum height=0.225\BenImgWidth, fill=orange!20, draw, thick](LoRAA) at ($(InternLM.west) + (2, 0.3)$) {\small A};
        \node [anchor=north, trapezium, trapezium angle=60, minimum height=0.225\BenImgWidth, fill=orange!20, draw, thick](LoRAB) at ($(LoRAA.south) + (0, -.05)$) {\small B};
        \draw[decoration={brace,raise=15pt},decorate]  ([yshift=-1pt, xshift=-5pt]LoRAB.south) -- node[left=18pt] {\small LoRA} ([yshift=1pt, xshift=-5pt]LoRAA.north);

        \node[align=justify, anchor=north, text width=95mm, inner sep=4pt, fill=gray!10, rounded corners=2mm] (FinalCaption)
            at ($(InternLM.south) + (0,-0.4)$)
            {\small \enquote{This satellite image, captured during the summer season in Switzerland, showcases dominantly arable land ...}};

        \draw[thick, ->] (rgbIN) -- (rgbViT);
        \draw[thick, ->] (s1IN) -- (s1ViT);
        \draw[thick, ->] (s2IN) -- (s2ViT);

        \draw[thick, ->] (rgbViT) -- (rgbProj);
        \draw[thick, ->] (s1ViT) -- (s1Proj);
        \draw[thick, ->] (s2ViT) -- (s2Proj);
        \draw[thick, ->] (Prompt) -- (tokenizer);
        
        \draw[thick, ->] (rgbProj) -- (rgbProj |- InternLM.north);
        \draw[thick, ->] (s1Proj) -- (s1Proj |- InternLM.north);
        \draw[thick, ->] (s2Proj) -- (s2Proj |- InternLM.north);
        \draw[thick, ->] (tokenizer) -- (tokenizer |- InternLM.north);
        
        \draw[thick, ->] (InternLM) -- (FinalCaption);

        \node at ($(rgbViT.north east) + (-.3, -.3)$) {\includegraphics[height=0.2\BenImgWidth]{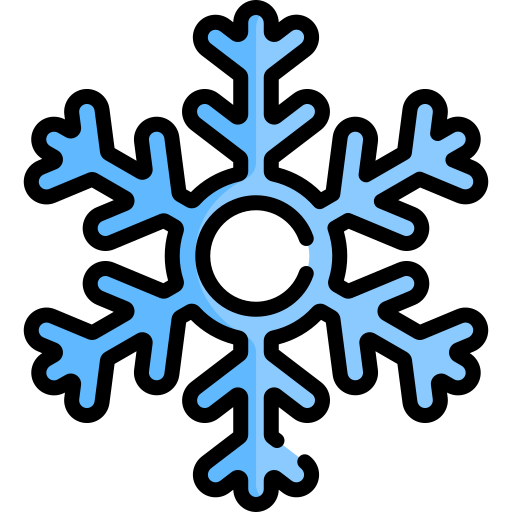}};
        \node at ($(s1ViT.north east) + (-.3, -.3)$) {\includegraphics[height=0.2\BenImgWidth]{figures/frozen.png}};
        \node at ($(s2ViT.north east) + (-.3, -.3)$) {\includegraphics[height=0.2\BenImgWidth]{figures/frozen.png}};
        
        \node at ($(rgbProj.north east) + (-.3, -.3)$) {\includegraphics[height=0.2\BenImgWidth]{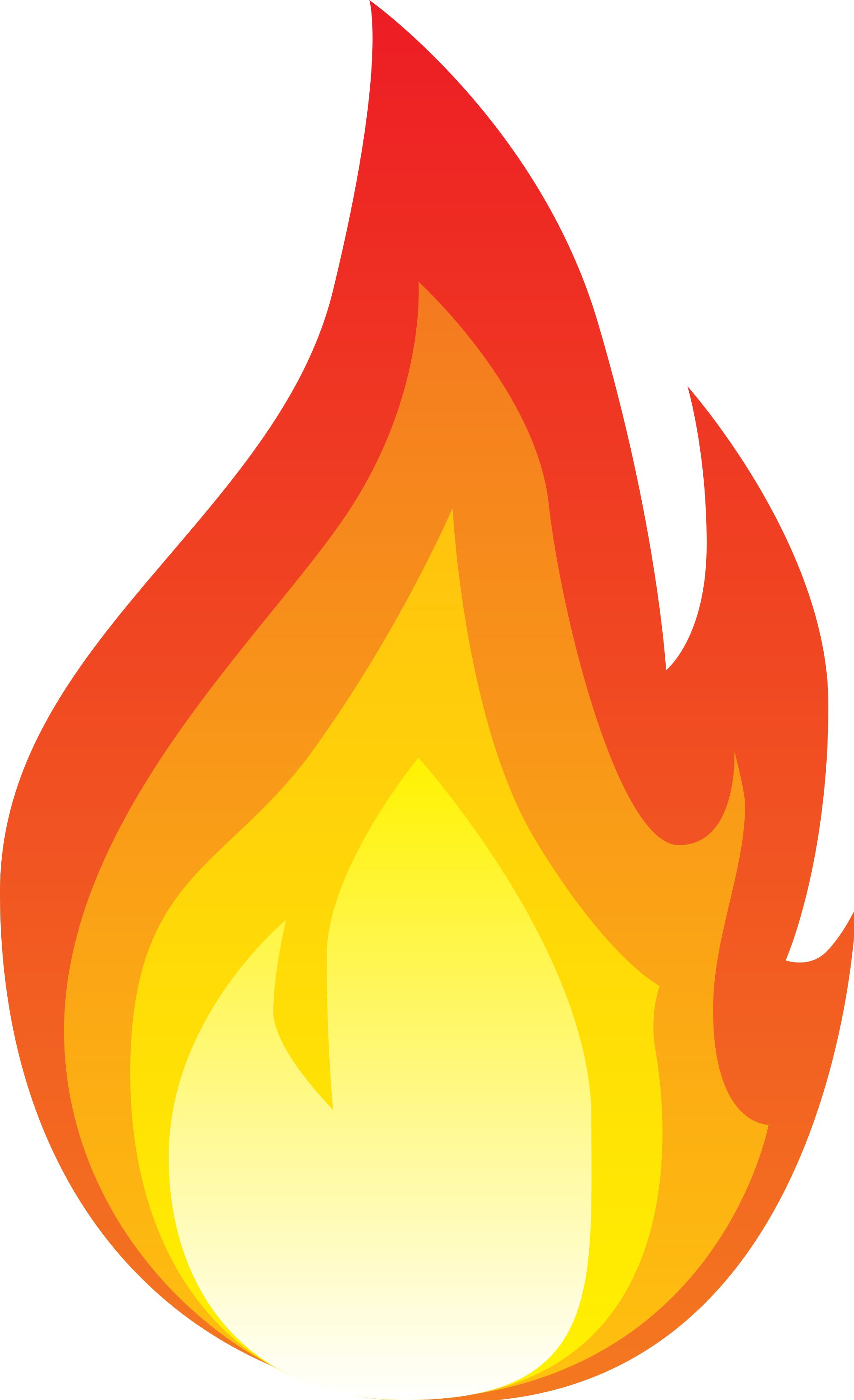}};
        \node at ($(s1Proj.north east) + (-.3, -.3)$) {\includegraphics[height=0.2\BenImgWidth]{figures/tuned.png}};
        \node at ($(s2Proj.north east) + (-.3, -.3)$) {\includegraphics[height=0.2\BenImgWidth]{figures/tuned.png}};
        \node at ($(tokenizer.north east) + (-.3, -.3)$) {\includegraphics[height=0.2\BenImgWidth]{figures/frozen.png}};
        
        \node at ($(LoRAA.north east) + (-.06, -.17)$) {\includegraphics[height=0.1\BenImgWidth]{figures/tuned.png}};
        \node at ($(LoRAB.north east) + (-.06, -.17)$) {\includegraphics[height=0.1\BenImgWidth]{figures/tuned.png}};
        \node at ($(InternLM.north east) + (-.3, -.3)$) {\includegraphics[height=0.2\BenImgWidth]{figures/frozen.png}};

    \end{tikzpicture} 

    \caption{Architectural overview of adapted InternVL to multi-sensor input. InternVL parts are marked in green while additions are orange.
    The \ac{S2} input is visualized as false color composite of an \ac{S2} image (\ac{RGB}: Band 8 – NIR, Band 4 – Red, Band 3 – Green).
    The \ac{S1} input is visualized as 
    (RGB: VV, VH, VV/VH).
    }
    \label{fig:model_architecture}
\end{figure*}

%% file: tables/results/binary_ours.tex
\begin{table}[tb]
    \centering
    \renewcommand{\arraystretch}{\arrayStretchFactor}
    \caption{Results on the binary \ac{VQA} tasks for \ac{VLM}s in \ac{RS} and \ac{CV}: Presence, Area, Counting, Adjacency, and Overall, denoted as \presenceABBR, \areaABBR, \countABBR, \adjacencyABBR, and \overallABBR, in terms of accuracy from the fine-tuned InternVL$^*$ and best results from \acp{VLM} in \ac{RS} and \ac{CV}.
    All results are in percent (\%).}
    \begin{tabular}{l *5{S[table-format=2.2]}}
        \toprule
        {\thead{Model}}
        & {\presenceABBR}
        & {\areaABBR}
        & {\countABBR}
        & {\adjacencyABBR}
        & {\overallABBR}
        \\\midrule
        \ac{SOTA} \ac{RS} & 69.34 & 56.20 & 51.99 & 60.62 & 57.90 \\
        \ac{SOTA} \ac{CV} & 64.33 & 67.38 & 62.33 & 58.94 & 61.96 \\\midrule
        \textbf{InternVL$^*$} & \best{69.68} & \best{82.39} & \best{78.02} & \best{68.81} & \best{73.29} \\
        \bottomrule
        \end{tabular}
    \label{tab:binary_results_ours}
\end{table}

%% file: tables/results/mcq_ours.tex
\begin{table}[h!bt]
    \centering
    \renewcommand{\arraystretch}{\arrayStretchFactor}
    \caption{Results on the \ac{MCQ} tasks for \ac{VLM}s in \ac{RS} and \ac{CV}: Presence, Area, Counting, Adjacency, Relative Position, Country, Season, Climate Zone, and Overall, denoted as \presenceABBR, \areaABBR, \countABBR, \adjacencyABBR, \relativepositionABBR, \countryABBR, \seasonABBR, \climateABBR, and \overallABBR, respectively in terms of accuracy for the fine-tuned InternVL$^*$ and best results from \acp{VLM} in \ac{RS} and \ac{CV}.
    All results are in percent (\%).}
    \begin{tabular}{l *9{S[table-format=2.2]}}
    \toprule
    Model
    & {\presenceABBR}
    & {\areaABBR}
    & {\countABBR}
    & {\adjacencyABBR}
    & {\relativepositionABBR}
    & {\countryABBR}
    & {\seasonABBR}
    & {\climateABBR}
    & {\overallABBR}\\
    \midrule
        \ac{SOTA} \ac{RS} & 46.40 & 27.41 & 36.43 & 44.70 & 31.60 & 54.17 & 25.71 & 37.04 & 35.26 \\
        \ac{SOTA} \ac{CV} & 38.08 & \best{44.42} & 34.01 & 43.33 & 34.82 & 47.76 & 27.43 & 45.93 & 37.55 \\\midrule
        \textbf{InternVL$^*$} & \best{64.88} & 42.11 & \best{55.20} & \best{56.90} & \best{57.98} & \best{88.78} & \best{89.14} & \best{88.44} & \best{65.84} \\
    \bottomrule
    \end{tabular}
    \label{tab:mcq_results_ours}
\end{table}

%% file: tables/results/bbox_reference_ours.tex
\begin{table}[h!bt]
    \centering
    \renewcommand{\arraystretch}{\arrayStretchFactor}
    \caption{Results on \refDet{} for \ac{VLM}s in \ac{RS} and \ac{CV} in terms of \ac{mIoU}, and accuracy@$k$ (where a prediction is counted as correct if the overlap with the reference is greater than $k\%$). All results are in percent (\%).}
    \begin{tabular}{l *5{S[table-format=2.2]}}
        \toprule
        {Model} 
        & {mIoU}
        & {Acc@25} 
        & {Acc@50} 
        & {Acc@75} 
        & {Acc@90}\\
        \midrule
        \ac{SOTA} \ac{RS} & 22.65 & 38.41 & 17.88 & 5.67 & 1.39 \\
        \ac{SOTA} \ac{CV} & 25.16 & 40.81 & 19.52 & 4.28 & 1.39 \\\midrule
        \textbf{InternVL$^*$} & \best{33.16} & \best{47.73} & \best{34.38} & \best{18.77} & \best{6.55} \\
        \bottomrule
    \end{tabular}
    \label{tab:bb_reference_results_ours}
\end{table}

%% file: tables/results/bbox_point_ours.tex
\begin{table}[h!bt]
    \centering
    \renewcommand{\arraystretch}{\arrayStretchFactor}
    \caption{Results on \pointDet{} for \ac{VLM}s in \ac{RS} and \ac{CV} in terms of \ac{mIoU}, and accuracy@$k$ (where a prediction is counted as correct if the overlap with the reference is greater than $k\%$). All results are in percent (\%).} 
    \begin{tabular}{l *5{S[table-format=2.2]}}
        \toprule
        {Model} 
        & {mIoU}
        & {Acc@25} 
        & {Acc@50} 
        & {Acc@75} 
        & {Acc@90}\\
        \midrule
        \ac{SOTA} \ac{RS} & 12.65 & 17.51 & 5.08 & 1.27 & 0.13 \\
        \ac{SOTA} \ac{CV} & 38.95 & 64.21 & 35.53 & 8.63 & 2.79 \\\midrule
        \textbf{InternVL$^*$} & \best{71.81} & \best{97.21} & \best{85.66} & \best{54.19} & \best{15.74} \\
        \bottomrule
    \end{tabular}
    \label{tab:bb_point_results_ours}
\end{table}

%% file: tables/results/captioning_ours.tex
\begin{table}[h!bt]
    \centering
    \renewcommand{\arraystretch}{\arrayStretchFactor}
    \caption{Results on the captioning task for \ac{VLM}s in \ac{RS} and \ac{CV} in terms of n-gram-based (BLEU-4, ROUGE, METEOR, CIDEr), sentence-embedding-based (BERTScore, SBert-Cosine) and \ac{LLM}-based (CLAIR) metrics. All results are in percent (\%).}
    \begin{tabular}{l *7{S[table-format=2.2]}}
        \toprule
         Model
         & {BLEU-4}
         & {ROUGE}
         & {METEOR}
         & {CIDEr}
         & {BERTScore}
         & {\makecell{SBERT-\\Cosine}}
         & {CLAIR} \\
         \midrule
         \ac{SOTA} \ac{RS} & 1.66 &16.54 & 14.93 & 0.96 & 84.34 & 51.97 & 59.70 \\
         \ac{SOTA} \ac{CV} & 0.96 & 14.94 & 17.83 & 0.81 & 83.89 & 55.90 & 58.91 \\\midrule
         \textbf{InternVL$^*$} & \best{34.04} & \best{50.63} & \best{50.62} & \best{78.35} & \best{91.87} & \best{83.29} & \best{70.90} \\
         \bottomrule
    \end{tabular}
    \label{tab:captioning_ours}
\end{table}

%% file: tables/results/country_ablation.tex
\begin{table}[!tb]
    \centering
    \renewcommand{\arraystretch}{\arrayStretchFactor}
    \caption{Results on the main tasks of the \bentxt{} benchmark split for the fine-tuned adapted InternVL$^*$ model. 
    $\dagger$: InternVL$^*$ fine-tuned on all countries except Serbia and Finland, and evaluated on Serbia and Finland.
    Reported metric for captioning is BLEU-4, while that for binary \ac{VQA}, and \ac{MCQ} is accuracy, and that for \genDet{} is mIoU. All results are given in percent (\%).}
    \begin{tabular}{l *6{S[table-format=2.2]}}
        \toprule
        {Model}
        & {\makecell{Captioning}}
        & {\makecell{Binary VQA}}
        & \multicolumn{2}{c}{MCQ}
        & {\makecell{Ref. Exp. Detection}}
        \\
        \cmidrule(lr){4-5}
        & & & {all} & {w/o location} & \\
        \midrule
        \textbf{InternVL$^{*\dagger}$} & 24.04 & 62.62 & 48.31 & 51.58 & 40.96  \\ \midrule
        \textbf{InternVL$^*$} & \best{34.04} & \best{73.29} & \best{65.84} & \best{64.47} & \best{52.41}  \\
        \bottomrule
    \end{tabular}
    \label{tab:supp_countryabl}
\end{table}

%% file: tables/results/rgb_ablation.tex
\begin{table}[!tb]
    \centering
    \renewcommand{\arraystretch}{\arrayStretchFactor}
    \caption{Results on the main tasks of the \bentxt{} benchmark split for the fine-tuned adapted InternVL$^*$ model as well as best results from \acp{VLM} in \ac{RS} and \ac{CV}, which are \earthmindrgbABBR{} and \llavaABBR{} for captioning, \earthdialrgbABBR{} and \qwenABBR{} for binary \ac{VQA}, \earthmindsarmsABBR{} and \qwenABBR{} for \ac{MCQ}, and \earthmindsarmsABBR{} and \gptABBR{} for \genDet. 
    InternVL$_{rgb}^*$ fine-tuned and evaluated using only RGB bands.
    Reported metric for captioning is BLEU-4, while that for binary \ac{VQA}, and \ac{MCQ} is accuracy, and that for \genDet{} is mIoU. All results are given in percent (\%).}
    \begin{tabular}{l *5{S[table-format=2.2]}}
        \toprule
        {Model}
        & {\makecell{Captioning}}
        & {\makecell{Binary VQA}}
        & {\makecell{MCQ}} 
        & {\makecell{Ref. Exp. Detection}}
        \\
        \midrule
        \ac{SOTA} \ac{RS} & 1.66 & 58.38 & 35.26 & 16.18 \\
        \ac{SOTA} \ac{CV} & 0.96 & 61.96 & 37.55 & 31.73  \\
        \midrule
        \textbf{InternVL$_{rgb}^*$} & 30.88 & 65.08 & 55.73 & 47.78  \\
        \midrule
        \textbf{InternVL$^*$} & \best{34.04} & \best{73.29} & \best{65.84} & \best{52.41}  \\
        \bottomrule
    \end{tabular}
    \label{tab:supp_rgbabl}
\end{table}